\documentclass{article} 
\usepackage[utf8]{inputenc}
\usepackage{graphicx}
\usepackage{float}
\usepackage{natbib}

\usepackage{amssymb}
\usepackage[margin=1in]{geometry}
\usepackage{ifthen}
\usepackage[table]{xcolor}
\usepackage{minitoc}
\usepackage{array}
\usepackage{makecell}
\usepackage{pdfpages}
\usepackage{mathtools}
\usepackage{tikz}
\usepackage{multirow}
\usepackage{booktabs}

\usepackage{arydshln} 
\usepackage{xcolor}         
\definecolor{DarkGreen}{rgb}{0.1,0.5,0.1}
\definecolor{DarkRed}{rgb}{0.5,0.1,0.1}
\definecolor{DarkBlue}{rgb}{0.1,0.1,0.5}
\definecolor{DarkYellow}{rgb}{.79,.79,0}
\definecolor{unitednationsblue}{rgb}{0.36, 0.57, 0.9}
\definecolor{blue_ppt}{rgb}{0,0.6,0.93}

\definecolor{darkblue_ppt}{rgb}{0.05,0.4,0.8}

\definecolor{orange_ppt}{rgb}{0.82,0.5,0}

\usepackage{color} 
\definecolor{yc}{RGB}{225,0,0}

\newcommand{\KL}[2]{\mathsf{KL}\left(#1\|#2\right)}

\newcommand{\var}{\mathsf{Var}}

\usepackage{hyperref}
\hypersetup{colorlinks,citecolor=blue,filecolor=blue,linkcolor=blue,urlcolor=blue}
\usepackage{cleveref}
\usepackage{tablefootnote} 

\newcolumntype{?}{!{\vrule width 1pt}}
\usepackage{amsmath}
\usepackage{amsthm}
\usepackage{algorithm,algorithmic}

\usepackage{graphicx} 
\usepackage{subfigure}
\usepackage{bbm}

\usepackage{extarrows}
\usepackage{bbm}
\theoremstyle{plain}
\newtheorem{lm}{Lemma} 

\newtheorem{thm}{Theorem}

\newtheorem{asmp}{Assumption}

\allowdisplaybreaks
\usepackage{xspace}

\usepackage{amsmath,amsfonts,bm}
\usepackage{algorithm,algorithmic}

\def\1{\bm{1}}

\def\vpi{{\bm{\pi}}}

\def\va{{\bm{a}}}

\DeclareMathAlphabet{\mathsfit}{\encodingdefault}{\sfdefault}{m}{sl}
\SetMathAlphabet{\mathsfit}{bold}{\encodingdefault}{\sfdefault}{bx}{n}

\def\gA{{\mathcal{A}}}

\def\gD{{\mathcal{D}}}
\def\gE{{\mathcal{E}}}
\def\gF{{\mathcal{F}}}

\def\gL{{\mathcal{L}}}
\def\gM{{\mathcal{M}}}
\def\gN{{\mathcal{N}}}
\def\gO{{\mathcal{O}}}

\def\gS{{\mathcal{S}}}

\def\gX{{\mathcal{X}}}

\newcommand{\E}{\mathbb{E}}

\newcommand{\R}{\mathbb{R}}

\DeclareMathOperator*{\argmax}{arg\,max}
\DeclareMathOperator*{\argmin}{arg\,min}

\usepackage{amssymb}

\usepackage{xcolor}
\definecolor{mydarkblue}{rgb}{0,0.08,0.45}
\definecolor{mygreen}{rgb}{0.032, 0.6392, 0.2039}
\definecolor{mypurple}{HTML}{B266FF}

\def\NN{{\mathbb N}}

\def\PP{{\mathbb P}}

\def\NN{{\mathbb N}}

\newcommand{\norm}[1]{\left\| #1 \right\|}

\newcommand{\piref}{\pi_{\mathsf{ref}}}
\newcommand{\allpiref}{\vpi_{\mathsf{ref}}}
\newcommand{\muref}{\mu_{\mathsf{ref}}}
\newcommand{\nuref}{\nu_{\mathsf{ref}}}

\newcommand{\hellinger}[2]{D_{\mathsf{H}}^2\left(#1\|#2\right)}

\newcommand{\gap}[1]{\mathsf{Gap}\left(#1\right)}
\newcommand{\dualgap}[1]{\mathsf{DualGap}\left(#1\right)}

\definecolor{blue_ppt}{rgb}{0,0.47,0.97}

\newcommand{\regret}{\mathsf{Regret}}
\newcommand{\gamesolver}{\mathsf{Equilibrium}}
\newcommand{\sample}{\mathsf{Sampler}}

\definecolor{bluegray}{RGB}{77,117,154}
\newcommand{\trianglecomment}[1]{\hfill\textcolor{bluegray}{$\triangleright$ #1}}
\newcommand{\name}{\textsf{VMG}\xspace}

\title{Incentivize without Bonus: Provably Efficient Model-based \\ Online Multi-agent RL for Markov Games}
\author{Tong Yang\thanks{Carnegie Mellon University; Emails:
 	 	\texttt{\{tongyang,yuejiec\}@andrew.cmu.edu}.}  \\
CMU 
\and Bo Dai\thanks{Georgia Institute of Technology; Email: \texttt{bodai@cc.gatech.edu}.} \\
Georgia Tech \\ 
\and Lin Xiao\thanks{Fundamental AI Research, Meta; Email: \texttt{\{linx,ychi\}@meta.com}.} \\
Meta \\
\and Yuejie Chi \\
Meta \& CMU}
\date{January 2025}

\begin{document}
\maketitle

\begin{abstract}

Multi-agent reinforcement learning (MARL) lies at the heart of a plethora of applications involving the interaction of a group of agents in a shared unknown environment. A prominent framework for studying MARL is Markov games, with the goal of finding various notions of equilibria in a sample-efficient manner, such as the Nash equilibrium (NE) and the coarse correlated equilibrium (CCE). 
However, existing sample-efficient approaches either require  tailored uncertainty estimation  under function approximation, or  careful coordination of the players. In this paper, we propose a novel model-based algorithm, called \name, that incentivizes exploration via biasing the empirical estimate of the model parameters towards those with a higher collective  best-response values of all the players when fixing the other players' policies, thus encouraging the policy to deviate from its current equilibrium  for more exploration. \name is oblivious to different forms of function approximation, and permits simultaneous and uncoupled policy updates of all players. Theoretically, we also establish that \name achieves a near-optimal regret for finding both the NEs of two-player zero-sum Markov games and  CCEs of multi-player general-sum Markov games under linear function approximation in an online  environment,  which nearly match their counterparts with sophisticated uncertainty quantification.

\end{abstract}


\setcounter{tocdepth}{2}
\tableofcontents

\section{Introduction}

Multi-agent reinforcement learning (MARL) is emerging as a crucial paradigm for solving complex decision-making problems in various domains, including robotics, game theory, and machine learning~\citep{busoniu2008comprehensive}. While single-agent reinforcement learning (RL) has been extensively studied and theoretically analyzed, MARL is still in its infancy, and many fundamental questions remain unanswered. Due to the interplay of multiple agents in an unknown environment, one of the key challenges is the design of efficient strategies for exploration that can be seamlessly implemented in the presence of a large number of agents\footnote{In this paper, we use the term agent and player interchangeably.} without the need of complicated coordination among the agents. In addition, due to the large dimensionality of the state and action spaces, which grows exponentially with respect to the number of agents in MARL, it necessitate the adoption of function approximation to enable tractable planning in modern RL regimes.

A de facto approach in exploration in RL is the principle of optimism in the face of uncertainty \citep{lai1987adaptive}, which argues the importance of quantifying the uncertainty, known as the {\em bonus} term, in the pertinent objects, e.g., the value functions, and using their upper confidence bound (UCB) to guide action selection. This principle has been embraced in the MARL literature, leading a flurry of algorithmic developments \citep{liu2021sharp,bai2021sample,song2021can,jin2021v,li2022minimax,ni2022representation,cui2023breaking,wang2023breaking,dai2024refined} that claim provable efficiency in solving Markov games \citep{littman1994markov}, a standard model for MARL. However, a major downside of this approach is that constructing the uncertainty sets quickly becomes intractable as the complexity of function approximation increases, which often requiring a tailored approach. For example, near-optimal techniques for constructing the bonus function in the tabular setting cannot be applied for general function approximation using neural networks.

Therefore, it is of great interest to explore alternative exploration strategies without resorting to explicit uncertainty quantification, and can be adopted even for general function approximation. Our work is inspired by the pioneering work of \citet{kumar1982new}, which identified the need to regularize the maximum-likelihood estimator of the model parameters using its optimal value function to incentivize exploration, and has been successfully applied to bandits and  single-agent RL problems ~\citep{liu2020exploration,hung2021reward,mete2021reward,liu2024maximize} with matching performance of their UCB counterparts. However, this strategy of {\em value-incentivized} exploration has not yet been fully realized in the Markov game setting; a recent attempt \citep{liu2024maximize} addressed    two-player zero-sum Markov games, however, it requires  asymmetric updates and solving bilevel optimization problems with the lower level problem being a Markov game itself. 
These limitations motivate the development of more efficient algorithms for the general multi-agent setting while enabling symmetric and independent updates of the players. 
 We address the following question: 
\begin{center}
    \emph{Can we develop provably efficient algorithms for online multi-player general-sum Markov games with function approximation using value-incentivized exploration?}
\end{center}

\subsection{Contribution}

In this paper, we propose a provably-efficient model-based framework, named \name (\emph{\textbf{V}alue-incentivized \textbf{M}arkov \textbf{G}ame solver}), for solving online multi-player general-sum Markov games with function approximation. 
\name  
incentivizes exploration via biasing the empirical estimate of the model parameters towards those with a higher collective {\em best-response} values of all the players when fixing the other players' policies, thus encouraging the policy to deviate from the equilibrium of the current model estimate for more exploration. This approach is oblivious to different forms of function approximation, bypassing the need of designing tailored bonus functions to quantity the uncertainty in standard approaches. \name also permits simultaneous and uncoupled policy updates of all players, making it more suitable when the number of players scales. Theoretically, we also establish that \name achieves a near-optimal regret for a number of game-theoretic settings under linear function approximation, which are on par to their counterparts requiring explicit uncertainty quantification. Specifically, our main results are as follows.
\begin{itemize} 
    \item For two-player zero-sum matrix games, \name  achieves a near-optimal regret on the order of $\widetilde{O}(d\sqrt{T})$,\footnote{The notation $\widetilde{O}(\cdot)$ hides logarithmic factors in the standard order-wise notation.} where $d$ is the dimension of the feature space and $T$ is the number of iterations for model updates.  This translates to a sample complexity of $\widetilde{O}(d^2/\varepsilon^2)$ for finding an $\varepsilon$-optimal NE in terms of the duality gap.
    \item For finite-horizon multi-player general-sum Markov games, under the linear mixture model of the transition kernel, \name achieves a near-optimal regret on the order of $\widetilde{\gO}(d\sqrt{H^3T})$, where $H$ is the horizon length, and $T$ is the number of iterations for model updates. This translates to a near-optimal  --- up to a factor of $H$ ---  complexity of $\widetilde{O}(Nd^2H^4/\varepsilon^2)$ samples or $\widetilde{O}(Nd^2H^3/\varepsilon^2)$ trajectories for finding an $\varepsilon$-optimal CCE in terms of the optimality gap, which is also applicable to  finding $\varepsilon$-optimal NE for two-player zero-sum Markov games. We also extend \name to the infinite-horizon setting, which achieves a sample complexity of $\widetilde{\gO}(Nd^2/ ((1-\gamma)^4\varepsilon^2 ))$ to achieve $\varepsilon$-optimality.

     \item The unified framework of \name allows its reduction to important special cases such as symmetric matrix games, linear bandits and single-agent RL, which not only recovers the existing reward-biased MLE framework  but also discovers new formulation that might be of independent interest. 

\end{itemize}

\subsection{Related work}
\label{sec:related}

We  discuss a few threads of related work, focusing on those with theoretical guarantees.

\paragraph{Two-player matrix games.} Finding the equilibrium of two-player zero-sum matrix games has been studied extensively in the literature, e.g., \citet{mertikopoulos2018cycles,shapley1953stochastic,daskalakis2018last,wei2020linear}, where faster last-iterate linear convergence is achieved in the presence of KL regularization \citep{cen2021fast,zhan2023policy}. Many of the proposed algorithms focus on the tabular setting with full information, where the expected returns in each iteration can be computed exactly when the payoff matrix is given. More pertinent to our work, \citet{o2021matrix} considered matrix games with bandit feedback under the tabular setting, where only a noisy payoff from the players' actions is observed at each round, and proposed to estimate the payoff matrix using the upper confidence bounds (UCB) in an entry-wise manner~\citep{lai1987adaptive,bouneffouf2016finite}, as well as K-learning~\citep{o2021variational} that is akin to Thompson sampling \citep{russo2018tutorial}. Our work goes beyond the tabular setting, and proposes an alternative to UCB-based exploration that work seamlessly with different forms of function approximation.

\paragraph{Multi-player general-sum Markov games.} General-sum Markov games are an important class of multi-agent RL (MARL) problems \citep{littman1994markov}, and a line of recent works~\citep{liu2021sharp,bai2021sample,mao2023provably,song2021can,jin2021v,li2022minimax,sessa2022efficient} studied the non-asymptotic sample complexity for learning various equilibria in general-sum Markov games for the tabular setting under different data generation mechanisms. These works again rely heavily on carefully constructing confidence bounds of  the value estimates to guide data collection and obtain tight sample complexity bounds. In addition, policy optimization algorithms have also been developed assuming full information of the underlying Markov games, e.g., \citet{erez2023regret,zhang2022policy,cen2023faster}.

\paragraph{MARL with linear function approximation.} Modern MARL problems often involve large state and action spaces, and thus require function approximation to generalize from limited data. Most theoretical results focus on linear function approximation, where the transition kernel, reward or value functions are assumed to be linear functions of some known feature maps. The linear mixture model of the transition kernel considered herein follows a line of existing works in both single-agent and multi-agent settings, e.g., \citet{ayoub2020model,chen2022almost,modi2020sample,jia2020model,chen2022almost,liu2024maximize}, which is subtly different from another popular linear model \citep{jin2020provably,wang2019optimism,yang2019sample,xie2020learning}, and these two models are not mutually exclusive in general~\citep{chen2022almost}. Moreover, \citet{ni2022representation,huang2022towards} considered general function approximation and \citet{cui2023breaking,wang2023breaking,dai2024refined} considered independent function approximation to allow more expressive function classes that lead to stronger statistical guarantees, which usually require solving complicated constrained optimization problems to construct the bonus functions. 

\paragraph{Uncertainty estimation in online RL.} Uncertainty estimation is crucial for efficient exploration in online RL. Common approaches are constructing the confidence set of the model parameters based on the observed data, which have been demonstrated to be provably near-optimal in the tabular and linear function approximation settings~\citep{jin2018q,agarwal2023vo} but have  limited success in the presence of function approximation in practice~\citep{gawlikowski2023survey}. Thompson sampling provides an alternative approach to exploration by maintaining a posterior distribution over model parameters and sampling from this distribution to make decisions, which however becomes generally intractable under complex function approximation schemes~\citep{russo2018tutorial}. Our approach draws inspiration from the reward-biased maximum likelihood estimation framework, originally proposed by \citet{kumar1982new}, which has been recently adopted in the context of bandits \citep{liu2020exploration,hung2021reward,cen2024value} and single-agent RL \citep{mete2021reward,liu2024maximize}. However, to the best of our knowledge, this work is the first to generalize this idea to the multi-player game-theoretic setting, which not only recovers but leads to new formulations for the single-agent setting.

\subsection{Paper organization and notation}

The rest of this paper is organized as follows. Section~\ref{sec:matrix_game} studies two-player zero-sum matrix games,  Section~\ref{sec:markov} focuses on episodic multi-player general-sum Markov games, and we conclude in Section~\ref{sec:conclusion}. The proofs as well as the extension to the infinite-horizon setting are deferred to the appendix.

\paragraph{Notation.}
We let $[n]$ denote the index set $\{1, \dots, n\}$. Let $I_n$ denote the $n\times n$ identity matrix, and inner product in Euclidean space \(\mathbb{R}^n\) by $\langle\cdot,\cdot\rangle$. We let $\Delta^n$ denote the $n$-dimensional simplex, i.e., $\Delta^n=\{x\in\mathbb{R}^n: x\geq 0,\sum_{i=1}^n x_i=1\}$. For any $x\in\mathbb{R}^n$, we let $\|x\|_p$ denote the $\ell_p$ norm of $x$, $\forall p\in[1,\infty]$. We let $\mathbb{B}^d_2(R)$ denote the $d$-dimensional $\ell_2$ ball of radius $R$. The Kullback-Leibler (KL) divergence between two distributions $P$ and $Q$ is denoted as $\KL{P}{Q}\coloneqq \sum_{x}P(x)\log\frac{P(x)}{Q(x)}$.

\section{Two-Player Zero-Sum Matrix Games}
\label{sec:matrix_game}

In this section, we start with a simple setting of two-player zero-sum matrix games, to develop our algorithmic framework.

\subsection{Problem setting}\label{sec:matrix_setting}

\paragraph{Two-player zero-sum matrix game.} 
We consider the (possibly KL-regularized) two-player zero-sum matrix games with the following objective:
\begin{align}\label{eq:game_obj}
    \max_{\mu\in\Delta^m} \min_{\nu\in\Delta^{n}} \; f^{\mu,\nu}(A)\coloneqq\mu^\top A \nu - \beta\KL{\mu}{\muref} + \beta\KL{\nu}{\nuref},
\end{align}
where $A\in\mathbb{R}^{m\times n}$ is the payoff matrix, $\mu\in\Delta^m$ and $\muref\in\Delta^m$ (resp. $\nu\in\Delta^n$ and $\nuref\in\Delta^n$) are the policy and reference policy for the max (resp. min) player, and $\beta\geq 0$ is the regularization parameter.\footnote{For simplicity, we set the same regularization parameter for both players; our analysis continues to hold with different regularization parameters $\beta_1$ and $\beta_2$ for each player.} Here, the reference policies can be used to incorporate prior knowledge or preference of the game; when the reference policies are uniform distributions, the KL regularization becomes entropy regularization, which are studied in, e.g., \citet{cen2021fast}.

\paragraph{Nash equilibrium.} The policy pair $(\mu^\star,\nu^\star)$ corresponding to the solution to the saddle-point problem \eqref{eq:game_obj} represents a desirable state of the game, where both players perform their (regularized) best-response strategies against  the other player, so that no players will unitarily deviate from its current policy. Specifically,  the policy pair $(\mu^\star,\nu^\star)$ satisfies
$$\forall (\mu,\nu)\in\Delta^m\times\Delta^n:\quad f^{\mu,\nu^\star}(A)\leq f^{\mu^\star,\nu^\star}(A)\leq f^{\mu^\star,\nu}(A), $$
and is called the {\em Nash equilibrium} (NE) of the matrix game \citep{nash1950non}.\footnote{We note that under entropy regularization, the equilibrium is also known as the {\em quantal response equilibrium} (QRE) \citep{mckelvey1995quantal} when $\beta>0$.}

\paragraph{Noisy bandit feedback.}
We are interested in learning the NE when the payoff matrix $A$ is unknown and can only be accessed through a stochastic oracle. Specifically, for any $i\in[m]$ and $j\in[n]$, 
we can query the entry $A(i,j)$, and receive a noisy feedback $\widehat{A}(i,j)$ of $A(i,j)$ from an oracle, i.e.,
\begin{align}\label{eq:noisy_feedback}
    \widehat{A}(i,j) = A(i,j) + \xi,
\end{align}
where the noise $\xi$ is an i.i.d. zero-mean random variable across different queries. Each of the collected data tuple is thus in the form of $(i,j, \widehat{A}(i,j) )$.

\paragraph{Goal: regret minimization.} Our goal is to design an easy-to-implement framework that can find the approximate NE of the matrix game \eqref{eq:game_obj} with as few queries as possible to the stochastic oracle in a sequential manner. To begin, we define the following 
\begin{align}\label{eq:f_notation}
    f^{\star,\nu}(A)\coloneqq \max_{\mu\in\Delta^m} f^{\mu,\nu}(A),\quad f^{\mu,\star}(A)\coloneqq \min_{\nu\in\Delta^n} f^{\mu,\nu}(A),\quad\text{and}\quad f^\star(A)\coloneqq \max_{\mu\in\Delta^m}\min_{\nu\in\Delta^n} f^{\mu,\nu}(A)
\end{align}
for any payoff matrix $A$. The duality gap of the matrix game \eqref{eq:game_obj} at a policy pair $(\mu,\nu)$ is defined as
\begin{align}\label{eq:dualgap}
    \dualgap{\mu,\nu}\coloneqq    f^{\star,\nu}(A) - f^{\mu,\star}(A) ,
\end{align}
where it is evident that $ \dualgap{\mu^{\star},\nu^{\star}} = 0$. A policy pair $(\mu,\nu)$ is called an $\varepsilon$-approximate NE (abbreviated as $\epsilon$-NE) of the matrix game \eqref{eq:game_obj} if $\dualgap{\mu,\nu}\leq \varepsilon$.

In an online setting, given a sequence of policy updates $\{(\mu_t, \nu_t)\}_{t=1,\ldots, T}$ over $T$ rounds, a common performance metric is the cumulative regret, defined as 
\begin{align}\label{eq:regret}
    \regret(T)&\coloneqq \sum_{t=1}^T\dualgap{\mu_t,\nu_t}  = \underbrace{ \sum_{t=1}^T \left(  f^{\star,\nu_t}(A)-f^\star(A) \right) }_{\text{regret for min-player}}  + \underbrace{  \sum_{t=1}^T \left( f^\star(A) - f^{\mu_t,\star}(A) \right) }_{\text{regret for max-player}},
\end{align}
which encapsulates the regret from both players. Our goal is to achieve a sublinear, and ideally near-optimal, regret with respect to the number of rounds $T$, by carefully balancing the trade-off between exploration and exploitation, even under function approximation of the model class.

\subsection{Algorithm development}\label{sec:matrix_alg}

We propose a model-based approach, called \name, that enables provably efficient exploration-exploitation trade-off via resorting to a carefully-regularized model (i.e., the payoff matrix) estimator without constructing uncertainty intervals. To enable function approximation, we parameterize the payoff matrix by $A_{\omega} \in \mathbb{R}^{m\times n}$,  where $\omega\in\Omega\subset\mathbb{R}^d$ is some vector in the parameter space $\Omega$. 

The proposed approach, on a high level, alternates between updating the payoff matrix based on all the samples collected so far, and collecting new samples using the updated  policies. Let's elaborate a bit further. At each round $t$, let  the current payoff matrix estimate be $A_{\omega_{t-1}}$, and its corresponding NE be $(\mu_t, \nu_t)$. 
\begin{itemize}
\item {\em Value-incentivized model updates.} Given all the collected data tuples $\gD_{t-1}$ and the policy pair $(\mu_t, \nu_t)$, \name updates the model parameter $\omega_t$  via solving a regularized least-squares estimation problem as \eqref{eq:model_update}, favoring models that {\em minimizes} the squared loss between the model and the noisy feedback stored in $\gD_{t-1}$, and {\em maximizes} the value of each player when the other player's strategy is fixed. In other words, the regularization term aims to maximize the duality gap at $(\mu_t, \nu_t)$, which tries to pull the model away from its current estimate $A_{t-1}$, whose duality gap is $0$ at $(\mu_t, \nu_t)$. The regularized estimator thus strikes a balance of exploitation (via least-squares on $\gD_{t-1}$) and exploration (via regularization against the current model $A_{\omega_{t-1}}$).

\item {\em Data collection from best-response policy updates.} Using the updated payoff matrix $A_{\omega_t}$, \name updates the best-response policy of each player while fixing the policy of the other player via \eqref{eq:policy_update_2}, resulting in policy pairs $(\widetilde{\mu}_t,\nu_t)$ and $(\mu_t,\widetilde{\nu}_t)$. Finally, \name collects one new sample from each of the policy pairs respectively following the oracle \eqref{eq:noisy_feedback}, and add them to the dataset $\gD_{t-1}$ to form $\gD_t$. 

\end{itemize}
The complete procedure of \name is summarized in Algorithm~\ref{alg:matrix_game}. \name invokes the mechanism of regularization as a means for incentivizing exploration, rendering it more amenable to implement in the presence of function approximation. In contrast, prior approach \citep{o2021matrix} heavily relies on 
explicitly adding an exploration bonus to the estimate of the payoff matrix using confidence intervals, which is challenging to construct under general function approximation. In addition, \name allows parallel and independent policy execution from both players.

\begin{algorithm}[th]
    \caption{Value-incentivized Online Matrix Game (\name)}
    \label{alg:matrix_game}
    \begin{algorithmic}[1]
    \STATE \textbf{Input:}  initial parameter $\omega_0$, regularization coefficient $\alpha>0$, iteration number $T$.  
    \STATE \textbf{Initialization:} dataset $\gD_0\coloneqq \emptyset$.
    \FOR{$t = 1,\cdots,T$}
    \STATE 
    Compute the Nash equilibrium $(\mu_t,\nu_t)$ of  the matrix game with the current parameter $\omega_{t-1}$:
    \begin{align}\label{eq:policy_update}
        \mu_t&=\argmax_{\mu\in\Delta^m}\min_{\nu\in\Delta^n}\; f^{\mu,\nu}(A_{\omega_{t-1}}),\qquad \nu_t=\argmin_{\nu\in\Delta^n}\max_{\mu\in\Delta^m}\; f^{\mu,\nu}(A_{\omega_{t-1}}).
    \end{align} 
    \STATE Model update: Update the parameter $\omega_t$ by minimizing the following objective:
    \begin{align}\label{eq:model_update}
        \omega_t=\argmin_{\omega\in\Omega}\sum_{(i,j,\widehat{A}(i,j))\in\gD_{t-1}}\left(A_\omega(i,j)-\widehat{A}(i,j)\right)^2  \underbrace{ -\alpha f^{\star, \nu_t}(A_{\omega})+\alpha f^{\mu_t,\star}(A_{\omega})}_{\textrm{value-incentivized~reg.}}.
    \end{align}
    \STATE Compute $\widetilde{\mu}_t$ and $\widetilde{\nu}_t$ by solving the following optimization problems:
    \begin{align}\label{eq:policy_update_2}
        \widetilde{\mu}_t& = \argmax_{\mu\in\Delta^m}\; f^{\mu,\nu_t}(A_{\omega_t}),\qquad \widetilde{\nu}_t = \argmin_{\nu\in\Delta^n}\; f^{\mu_t,\nu}(A_{\omega_t}).
    \end{align}
    \STATE Data collection: Sample $(i_t,j_t)\sim (\widetilde{\mu}_t,\nu_t)$ and $(i'_t,j'_t)\sim (\mu_t,\widetilde{\nu}_t)$ and get the noisy feedback $\widehat{A}(i_t,j_t)$ and $\widehat{A}(i'_t,j'_t)$ following the oracle \eqref{eq:noisy_feedback}. Update the dataset ${\gD}_t=\gD_{t-1}\cup\left\{ (i_t,j_t,\widehat{A}(i_t,j_t) ), \, (i'_t,j'_t,\widehat{A}(i'_t,j'_t) )\right\}$.
    \ENDFOR 
    \end{algorithmic}
\end{algorithm}

\paragraph{The benefit of regularization.} While \name is agnostic to the power of KL regularization in \eqref{eq:game_obj}, the major benefit of regularization comes in terms of computational efficiency. When the KL regularization parameter $\beta>0$, common first-order game solvers such as mirror descent ascent~\citep{sokota2022unified} or policy extragradient~\citep{cen2021fast} achieve a last-iterate  linear convergence rate when solving the matrix game \eqref{eq:policy_update}. Turning to the model update, when $\beta>0$, the regularization term in \eqref{eq:model_update} can be computed in closed form:
\begin{align}\label{eq:regularization_closed_form}
    & -  f^{\star,\nu_t}(A_\omega)+   f^{\mu_t,\star}(A_\omega) \notag\\
    &= - \beta\left[\log\left(\sum_{i=1}^n \mu_{\mathsf{ref},i} \exp\left(\frac{A_{\omega}(i,:)\nu_t}{\beta}\right)\right) +\log\left(\sum_{j=1}^m \nu_{\mathsf{ref},j} \exp\left(-\frac{\mu_t^\top A_{\omega}(:,j)}{\beta}\right)\right)\right]+C,
\end{align}
where $\mu_{\mathsf{ref},i}$ (resp. $\nu_{\mathsf{ref},j}$) is the $i$-th (resp. $j$-th) entry of $\mu_{\mathsf{ref}}$ (resp. $\nu_{\mathsf{ref}}$), $A_{\omega}(i,:)$ (resp. $A_{\omega}(:,j)$) is the $i$-th row (resp. $j$-th column) of $A_{\omega}$, and $C$ is a constant that does not depend on $A_\omega$. Leveraging the closed-form expression, one can bypass solving a bi-level optimization problem \eqref{eq:model_update} on its surface, but resorts to more efficient first-order methods. Last but not least, the policies $\widetilde{\mu}_t$ and $\widetilde{\nu}_t$ in \eqref{eq:policy_update_2} can be computed in closed form as well:
\begin{align}\label{eq:tilde_pi_closed_form}
    \widetilde{\mu}_{t,i} \propto \mu_{\mathsf{ref},i} \exp\left(\frac{A_{\omega_t}(i,:)\nu_t}{\beta}\right) ,\qquad \widetilde{\nu}_{t,j} \propto  \nu_{\mathsf{ref},j} \exp\left(-\frac{\mu_t^\top A_{\omega_t}(:,j)}{\beta}\right) ,\quad \forall i\in[m], j\in[n].
\end{align}

\paragraph{The case of symmetric payoff.} One important special class of matrix games is the symmetric matrix game~\citep{cheng2004notes}, with $A=-A^\top$, $\muref=\nuref$, and $m=n$. Many well-known games are symmetric, from classic games like rock-paper-scissors to the recent example of LLM alignment~\citep{munos2023nash,swamy2024minimaximalist,yang2024faster}. For a symmetric matrix game, it admits a symmetric Nash $(\mu^\star,\mu^\star)$, and Algorithm~\ref{alg:matrix_game}  reduces to a single-player algorithm by only tracking a single policy $\mu_t$, recognizing $\mu_t=\nu_t$ and $\widetilde{\mu}_t= \widetilde{\nu}_t$ due to $f^{\mu,\nu}(A)=-f^{\nu,\mu}(A)$. In addition,  \name only needs to collect one sample from the policy pair $(\widetilde{\mu}_t,\mu_t)$ in each iteration. This is particularly desirable when the policy is expensive to store and update, such as large-scale neural networks or LLMs. 

\paragraph{Reduction to the bandit case.} By setting the action space of the min player to $n=1$, \name seamlessly reduces to the bandit setting, where the payoff matrix becomes a reward vector $A \in \mathbb{R}^m$. Here, we let $f^{\mu}(A) = \mu^{\top} A - \beta\KL{\mu}{\muref}$ and 
     $f^{\star}(A)\coloneqq \max_{\mu\in\Delta^m} f^{\mu}(A)$. Interestingly, to encourage exploration, the regularization term favors a reward estimate that maximizes its regret $  f^{\star}(A_{\omega}) -  f^{\mu_t}(A_{\omega})$ on the current policy $\mu_t$, which is {\em different from} the reward-biasing framework that only regularizes against $  f^{\star}(A_{\omega})$ \citep{cen2024value,liu2020exploration}.   

\subsection{Theoretical guarantee}\label{sec:matrix_analysis}

We demonstrate that \name achieves near-optimal regret, assuming linear function approximation of the payoff matrix. Specifically, we have the following assumption.

\begin{asmp}[Linear function approximation]\label{asmp:bounded_payoff}
The payoff matrix is parameterized as  
\begin{align}\label{eq:A_param}
    A_\omega(i,j)\coloneqq \phi(i,j)^\top \omega, \quad\forall i\in[m], j\in[n],
\end{align}
where $\omega\in\Omega\subset\mathbb{R}^d$ is the parameter vector and $\phi(i,j)\in\mathbb{R}^d$ is the feature vector for the $(i,j)$-th entry. Here, the feature vectors are known and fixed, and satisfy $\norm{\phi(i,j)}_2\leq 1$ for all $i\in[m]$ and $j\in[n]$. For all $ \omega\in\Omega$, we suppose $\norm{\omega}_2\leq \sqrt{d}$ and $\norm{A_\omega}_{\infty}\leq B_l$ for some $B_l>0$. 
\end{asmp}

We also assume that the linear function class is expressive enough to describe the true payoff matrix $A$.
\begin{asmp}[realizability]\label{asmp:expressive}
    There exists $\omega^\star\in\Omega$ such that $A_{\omega^\star}=A$.
\end{asmp}

Next, we impose the noise follows standard sub-Gaussian distribution.
\begin{asmp}[i.i.d. sub-Gaussian noise]\label{asmp:noise}
    The noise $\xi$ in \eqref{eq:noisy_feedback} are i.i.d. mean-zero sub-Gaussian random variables with sub-Gaussian parameter $\sigma>0$.
\end{asmp}

\paragraph{Regret guarantee.} The following theorem states the regret bound of \name under appropriate choice of the regularization parameter.  
\begin{thm}\label{thm:regret}
    Suppose Assumptions~\ref{asmp:bounded_payoff}, \ref{asmp:expressive} and \ref{asmp:noise} hold. Let $\delta\in(0,1)$, setting the 
regularization coefficient $\alpha$ as
\begin{align}\label{eq:alpha}
\alpha=\sqrt{\frac{T}{d\log\left(1+(T/d)^{3/2}\right)} \left(\log(4T/\delta)+d\log (dT)\right)},
\end{align}
then for any $\beta\geq 0$, with any initial parameter $\omega_0$ and reference policies $\muref$ and $\nuref$, we have with probability at least $1-\delta$,
    \begin{align}\label{eq:regret_bound}
  \regret(T)=\gO\left(B_l\left(B_l+\sigma\sqrt{2\log(8T/\delta)}\right)d\sqrt{T  \log (dT )}\right)
    \end{align}
    for all $T\in \NN_+$.
\end{thm}

The proof of Theorem~\ref{thm:regret} is deferred to Appendix~\ref{app:proof_regret}. Theorem~\ref{thm:regret} establishes that by setting $\alpha$ on the order of $\widetilde{O}\big(\sqrt{T}\big)$, with high probability, the regret of \name is no larger than an order of
$$\widetilde{\gO}\left(d\sqrt{T}\right),$$
assuming the payoff matrix and the noise $\sigma$ are well-bounded.  In particular, when reduced to the linear bandit setting, this matches with the lower bound $\Omega(d\sqrt{T})$ established in  \citet{dani2008stochastic}, suggesting the near-optimality of our result. 
In addition, since $ \min_{t\in[T]}\dualgap{\mu_t,\nu_t}\leq \regret(T) /T$, \name is guaranteed to find an $\varepsilon$-NE of the matrix game \eqref{eq:game_obj} for any $\varepsilon>0$ within $\widetilde{\gO}\left( d^2 /\varepsilon^2 \right)$ samples.

\section{Multi-player General-sum Markov Games}\label{sec:markov}
 
We now turn to the more challenging setting of online multi-player general-sum Markov games, which includes the two-player zero-sum Markov game as a special case. 

\subsection{Problem setting}\label{sec:markov_setting}

\paragraph{Multi-player general-sum Markov game.} We consider an $N$-player general-sum episodic Markov game with a finite horizon denoted as $\gM_\PP\coloneqq(\gS,\gA,\PP, r, H)$, where $\gS$ is the state space, $\gA\coloneqq\gA_1\times\cdots\times\gA_N\coloneqq\prod_{n=1}^{N} \gA_n$ is the joint action space for all players, with $\gA_n$  the action space of player $n$, and
$H\in\NN_+$ is the horizon length. Let $\Delta(\gS)$ and $\Delta(\gA)$ denote the set of probability distributions over $\gS$ and $\gA$, respectively.  
$\PP=\{\PP_h\}_{h\in[H]}$ with $\PP_h:\gS\times\gA\rightarrow\Delta(\gS)$ is the inhomogeneous transition kernel: at step $h$, the probability of transitioning from state $s$ to state $s'$ by the action $\va=(a^1,\cdots,a^n)$ is $\PP_h(s'|s,\va)$. $r =\{r_h^n\}_{h\in[H],n\in[N]}$ stands for the reward function with
$r_h^n:\gS\times\gA\rightarrow[0,1]$ the reward of the $n$-th player at step $h$.

\paragraph{Markov policies.} In this paper, we focus on the class of Markov policies, where the policy of each player depends only on the current state, without dependence on the history. We let $\pi^n:\gS\times[H]\rightarrow \Delta(\gA)$ denote the policy of player $n$, and $\pi^n_h(\cdot|s)\in\Delta(\gA_n)$ denotes the probability distribution of the action of player $n$ at step $h$ given any state $s$. 
We let $\vpi=(\pi^1,\cdots,\pi^N):\gS\times[H]\rightarrow\Delta(\gA)$ denote the joint Markov policy (we assume all policies appear in this paper are Markovian, and we let \textit{joint policy} stands for joint Markov policy), where $\vpi_h(\cdot|s)\coloneqq (\pi_h^1,\cdots,\pi_h^N)(\cdot|s)\in\Delta(\gA)$ for all $s\in\gS$ and $h\in[H]$.  
 For any joint policy $\vpi$, we let $\vpi^{-n}$ denote the joint policy excluding player $n$. With a slight abuse of notation, we write $\vpi=(\pi^n,\vpi^{-n})$. In addition, a joint policy $\vpi$ is called a \textit{product policy} if $\pi^1,\cdots,\pi^N$ are executed independently, i.e., under policy $\vpi$, each player takes actions independently. We denote $\vpi = \pi^1\times \cdots \times \pi^N$ for a product policy.

 \paragraph{KL-regularized value function and Q-function.} 
 Given a joint policy $\vpi$, the KL-regularized state-value function (\textit{value function}) $V^{\vpi}_{h,n}: \gS\rightarrow\R$ and the KL-regularized state-action value function (\textit{Q-function}) $Q^{\vpi}_{h,n}:\gS\times\gA\rightarrow\R$ of the $n$-th player under $\vpi$ --- with regularization parameter $\beta\geq 0$ --- are respectively defined as
 \begin{subequations}
 \begin{align}
    \forall s\in\gS, h\in[H]:\quad V_{h,n}^{\vpi}(s) &\coloneqq \E_{\PP,\vpi}\left[\sum_{i=h}^H r_i^n(s_i,\va_i)-\beta\log\frac{\pi^n(a_i^n|s_i)}{\piref^n(a_i^n|s_i)}\bigg|s_h=s\right],\label{eq:V}\\
    \forall (s,\va)\in\gS\times\gA, h\in[H]:\quad Q_{h,n}^{\vpi}(s,\va) &\coloneqq r_h^n(s,\va) +  \E_{s'\sim\PP_h(\cdot|s,\va)}\left[V_{h+1,n}^{\vpi}(s')\right],\label{eq:Q}
 \end{align}
 \end{subequations}
 where $s_i$ and $\va_i$ are the state and action at step $i$, respectively, and $\allpiref:\gS\times[H]\rightarrow\Delta(\gA)$ is the reference policy. When the reference policy is a uniform distribution over the joint action space, the regularization becomes the entropy regularization. 
  In \eqref{eq:V}, $\pi^n(\cdot|s)$ (resp.~$\piref^n(\cdot|s)$) should be understood as the marginal distribution of player $n$ under joint distribution $\vpi(\cdot|s)$ (resp.~$\allpiref(\cdot|s)$),
 and we define $V_{H+1,n}^{\vpi}(s)=0$ for all $s\in\gS$ and $\beta\geq 0$. To simplify the notation, we define $V_n^{\vpi}\coloneqq V_{1,n}^{\vpi}$ and $Q_n^{\vpi}\coloneqq Q_{1,n}^{\vpi}$ for all $n\in[N]$.
We assume $\rho\in\Delta(\gS)$ is the initial state distribution, i.e., $s_1\sim\rho$. Furthermore, we define $V_n^\vpi(\rho)\coloneqq \E_{s\sim\rho}[V_n^\vpi(s)]$.

We let $\vpi=\pi^n\times\vpi^{-n}$ denote the policy profile where all players but the $n$-th player execute policy $\vpi^{-n}$, and the $n$-th player executes policy $\pi^n$ independent of the other players. For all $n\in[N]$, we define the best-response value function
\begin{align}\label{eq:V_dagger}
    \forall s\in\gS,h\in[H],n\in[N]:\quad V_{h,n}^{\star,\vpi^{-n}}(s)\coloneqq \max_{\pi^n:\gS\times[H]\rightarrow\Delta(\gA_n)} V_{h,n}^{\pi^n\times\vpi^{-n}}(s),
\end{align}
which is the optimal value of player $n$ when the policies of other agents are fixed by $\vpi^{-n}$.
Importantly, there exists at least one policy $\pi^{n,\star}(\vpi^{-n})$ that achieves the maximum in \eqref{eq:V_dagger} for all $s\in\gS$, and this policy is referred to the \textit{best-response policy} of player $n$ under joint policy $\vpi^{-n}$~\citep{shapley1953stochastic}.
 We also define 
 $$V_n^{\star,\vpi^{-n}}(\rho)\coloneqq \max_{\pi^n:\gS\times[H]\rightarrow\Delta(\gA_n)} V_n^{\pi^n\times\vpi^{-n}}(\rho).$$
One important thing to notice is that the best-response policy $\pi^{n,\star}(\vpi^{-n})$ does not depend on the initial state distribution $\rho$~\citep{mei2020global}.

\paragraph{Equilibria of Markov games.} In a multi-player general-sum Markov game, each agent aims to maximize its own value function, where the Nash equilibrium (NE)~\citep{nash1950non} and the coarse correlated equilibrium (CCE)~\citep{aumann1987correlated} are two widely studied solution concepts, whose definitions are as follows.
\begin{itemize}
    \item \textit{Nash equilibrium (NE):} a product policy $\vpi=\pi^1\times\cdots\times\pi^N$ is a Nash equilibrium of $\gM_\PP$ if
    \begin{align}\label{eq:def_NE}
            \forall s\in\gS, n\in[N]:\quad V_n^{\star,\vpi^{-n}}(s)=V_n^{\pi^n,\vpi^{-n}}(s).
    \end{align}
    \item \textit{Coarse correlated equilibrium (CCE):} a joint policy $\vpi$ is a CCE of $\gM_\PP$ if
    \begin{align}\label{eq:def_CCE}
            \forall s\in\gS, n\in[N]:\quad V_n^{\star,\vpi^{-n}}(s)\leq V_n^{\pi^n,\vpi^{-n}}(s).
    \end{align}
\end{itemize}
It is obvious that every NE is a CCE, but the converse is not true in general. In general, computing the NE in general-sum Markov games is intractable \citep{daskalakis2009complexity}, except for two-player zero-sum Markov games.


\paragraph{Goal: regret minimization.} 
To measure the proximity of a policy $\vpi$ to the equilibrium, we define the (average) sub-optimality gap of policy $\vpi$ w.r.t. the initial distribution $\rho$ as
\begin{align}\label{eq:gap}
    \gap{\vpi}\coloneqq \frac{1}{N}\sum_{n=1}^N \left(V_n^{\star,\vpi^{-n}}(\rho) - V_n^{\vpi}(\rho)\right).
\end{align}
A \textit{product} policy $\vpi$ is said to be an $\varepsilon$-approximate NE (abbreviated as $\varepsilon$-NE) if $\gap{\vpi}\leq \varepsilon$, and a \textit{joint} policy $\vpi$ is said to be an $\varepsilon$-approximate CCE (abbreviated as $\varepsilon$-CCE) if $\gap{\vpi}\leq \varepsilon$.

We aim to design a model-based framework that find the approximate NE or CCE of the Markov game $\gM_\PP$ in a provably efficient manner. Similar to the matrix game setting, we consider the following regret measure:
\begin{align}\label{eq:regret_MG}
 \regret(T)\coloneqq\sum_{t=1}^T\gap{\vpi_t}=\sum_{t=1}^T\frac{1}{N}\sum_{n=1}^N \left(V_n^{\star,\vpi_t^{-n}}(\rho) - V_n^{\vpi_t}(\rho)\right),
\end{align}
where $\vpi_t$ is the policy profile at time $t$. Our goal is to achieve a sublinear regret with respect to the number of rounds $T$, by carefully balancing the trade-off between exploration and exploitation, even under function approximation of the model class.

\subsection{Algorithm development}\label{sec:markov_alg}

For simplicity, we will focus on the function approximation over the transition kernel of the Markov game assuming the reward function is fixed and deterministic, while it is straightforwardly to also incorporate the reward function approximation. We let $\gF$ denote the function class of the estimators of the transition kernel of the Markov game, and we denote the parameterized transition kernel as
$$\PP_f=(\PP_{f,1},\cdots,\PP_{f,H})\in\gF=\gF_1\times\cdots\gF_H,$$ 
where $\gF$ is the function class and $f$ is its parameterization.  We define the value function $V_{f,h,n}^{\vpi}$ under the transition kernel $\PP_f$ as
\begin{align}\label{eq:V_f}
    \forall s\in\gS, h\in[H]:\quad V_{f,h,n}^{\vpi}(s)\coloneqq \E_{\PP_f,\vpi}\left[\sum_{i=h}^H \left(r_i^n(s_i,\va_i)-\beta\log\frac{\pi^n(a_i^n|s_i)}{\piref^n(a_i^n|s_i)}\right)\bigg|s_h=s\right],
\end{align}
and the Q-function  $Q_{f,h,n}^{\vpi}$ is defined likewise.

Akin to the matrix game case, \name alternates between updating the model updates based on all the transitions observed so far, and collecting new trajectories using the updated policies.  Suppose that at the $t$-th iteration, the  current estimate of the transition kernel is $\PP_{f_{t-1}}$, and its corresponding NE or CCE is $\vpi_t$. \name alternates between the following two steps.
\begin{itemize}
\item {\em Value-incentivized model updates.} Given all the collected transitions $\gD_{t-1,h}$ at each step $h$ and the equilibrium policy $\vpi_t$, \name updates the model parameter $f_t$  via solving a regularized maximum likelihood estimation (MLE) problem as \eqref{eq:model_update_mg}, favoring models that {\em minimizes} the negative log-likelihood $\gL_t(f)$ of the model, i.e.
\begin{align}\label{eq:loss_mg}
    \gL_t(f)\coloneqq \sum_{h=1}^H\sum_{(s_h,\va_h,s_{h+1})\in\gD_{t-1,h}}-\log \PP_{f,h}(s_{h+1}|s_h,\va_h),
\end{align}
and {\em maximizes} the sum of the {\em best-response} values of each player when the other player's strategy is fixed at $\vpi_t^{-n}$. In words, the regularizer tries to encourage models that incentive the players to deviate from their current policy, resulting in better exploration. 

\item {\em Trajectory collection from best-response policy updates.} Using the updated model  $\PP_{f_{t}}$, \name updates the best-response policy $\widetilde{\pi}_t^n$ of each player while fixing the policy $\vpi_t^{-n}$ of the other player via \eqref{eq:policy_update_2_mg}. \name then collects new trajectories by following policy $\vpi_t$ and $(\widetilde{\pi}_t^n,\vpi_t^{-n})$ for all $n\in [N]$, and update the dataset. 
\end{itemize}
The complete procedure of \name is summarized in Algorithm~\ref{alg:markov_game}, where the function $\gamesolver(\gM_f)$ returns the NE or CCE of the Markov game $\gM_f$ by calling off-the-shelf solvers, e.g., \citet{cai2024near,zhang2022policy}. Note that we are primarily interested in finding the NE for two-player zero-sum Markov games, and the CCE for multi-player general-sum Markov games, due to computational tractability.

\paragraph{Comparison with MEX.} \citet[Algorithm~2]{liu2024maximize} proposed the MEX framework, which also considered using value functions as a means to incentive exploration for   two-player zero-sum Markov games. Their algorithm requires asymmetric updates --- and two sets of model parameters as a result --- of the max and min players, where the model update of the max player is regularized by the optimal value $V_f^{\star} = \max_{\pi^1}\min_{\pi^2} V_{1,1}^{\vpi}(\rho)$ of the Markov game, which is an expensive saddle-point optimization problem, and the model update of the min player is regularized by the best-response value function. In contrast, \name only leverages best-response value functions as a regularization, which is much easier to solve. \name also permits simultaneous  updates for all the players, making it amenable to multi-player general-sum Markov games. In contrast, MEX does not apply to this more general setting.

\begin{algorithm}[!thb]
    \caption{Value-incentivized Online Markov Game (\name)}
    \label{alg:markov_game}
    \begin{algorithmic}[1]
    \STATE \textbf{Input:} reference policies $\allpiref$, initial transition kernel estimate $f_0\in\gF$, regularization coefficient $\alpha>0$, iteration number $T$.
    \STATE \textbf{Initialization:} dataset $\gD_{0,h}\coloneqq \emptyset$, $\forall  h\in[H]$.  
    \FOR{$t = 1,\cdots,T$}
    \STATE $\vpi_t \leftarrow \gamesolver(\gM_{f_{t-1}})$. \trianglecomment{$\gamesolver(\gM_f)$ returns a CCE or NE of game $\gM_f$.}
    \STATE Model update: Update the estimator $f_t$ by minimizing the following objective:
    \begin{align}\label{eq:model_update_mg}         
        f_t=\argmin_{f\in\gF}\; \gL_t(f)  - \alpha \sum_{n=1}^N V_{f,n}^{\star,\vpi_t^{-n}}(\rho) .
    \end{align}
    \STATE Compute best-response policies $\{\widetilde{\pi}_t^n\}_{n\in[N]}$:
    \begin{align}\label{eq:policy_update_2_mg}
        \forall n\in[N]:\quad \widetilde{\pi}_t^n& = \argmax_{\pi^n:\gS\times[H]\rightarrow\Delta(\gA_n)} V_{f_t,n}^{\pi^n,\vpi_t^{-n}}(\rho).
    \end{align}
    \STATE Data collection: 
    sample a trajectory with transition tuples $\{(s_{t,h},\va_{t,h}, s_{t,h+1})\}_{h=1}^H$ by executing $\vpi_t$, and sample a trajectory with transition tuples $\{(s_{t,h}^n,\va_{t,h}^n, s_{t,h+1}^n)\}_{h=1}^H$ by executing $(\widetilde{\pi}_t^n,\vpi_t^{-n})$
    for each $n\in[N]$.
    Update the dataset  $\gD_{t,h} = \gD_{t-1,h}  \cup_{n=1}^N \{ (s_{t,h},\va_{t,h},s_{t,h+1}), (s_{t,h}^n,\va_{t,h}^n,s_{t,h+1}^n) \}$, $\forall h\in[H]$.  
    \ENDFOR 
    \end{algorithmic}
\end{algorithm}

\paragraph{Reduction to the single-agent MDP case.} \name can be reduced to the Markov decision process (MDP) setting via either setting the number of players $N=1$  in the multi-player general-sum Markov game, or setting the action space of the min player  to a singleton in the two-player zero-sum Markov game. Interestingly, the former leads to the value regularization $V_{f}^{\star}(\rho)$ studied in MEX \citep{liu2024maximize}, while the latter leads to a new form of regularizer $V_{f}^{\star}(\rho)-V_{f}^{\pi_t}(\rho)$, adding friction from the current policy $\pi_t$.

\subsection{Theoretical guarantee}\label{sec:markov_analysis}

We demonstrate that \name achieves near-optimal regret under the following linear mixture model of the transition kernel for Markov games.

\begin{asmp}[linear mixture model]\label{asmp:function_class}
    The function class $\gF=\gF_1\times\cdots\gF_H$ is
    $$\forall h\in[H]:\quad\gF_h\coloneqq\left\{f_h|f_h(s'|s,\va)=\phi_h(s,\va,s')^\top{\theta_h},\quad \forall (s,\va,s')\in\gS\times\gA\times\gS,\theta_h\in\Theta_h\right\},$$
    where $\phi_h=(\phi_h^1,\cdots,\phi_h^d):\gS\times\gA\times\gS\rightarrow\R^d$ are the known feature maps with $\phi_h^i:\gS\times\gA\rightarrow\Delta(\gS)$ being transition kernels for all $i\in[d]$. $\norm{\phi_h(s,\va,s')}_2\leq 1$ for all $(s,\va,s')$, and $\Theta_h\subseteq\mathbb{B}^d_2(\sqrt{d})$, $\forall h\in[H]$. For each $f_h\in\gF_h$ and $(s,\va)\in\gS\times\gA$, $f_h(\cdot|s,\va)\in\Delta(\gS)$, $\forall h\in[H]$.
\end{asmp}
The linear mixture model is a common assumption in the RL literature, see, for example, \citet{ayoub2020model,modi2020sample,cai2020provably} for  single-agent RL, and \citet{chen2022almost,liu2024maximize} for Markov games.
We also assume the function class $\gF$ is expressive enough to describe the true transition kernel of the Markov game.
\begin{asmp}[realizability]\label{asmp:realizable}
    There exists $f^\star\in\gF$ such that $\PP_{f^\star}=\PP$.
\end{asmp}

\paragraph{Regret guarantee.} We now present our main result for the regret of the online Markov game, whose  proof is deferred to Appendix~\ref{app:proof_thm_regret_MP}. 
\begin{thm}\label{thm:regret_MG}
    Under Assumptions~\ref{asmp:function_class} and \ref{asmp:realizable}, if setting the regularization coefficient $\alpha$ as
    $$\alpha = \sqrt{\frac{ T}{Hd\log\left(1+\frac{T^{3/2}H^2}{\sqrt{d}}\right)} \left( \log\left(\frac{HN}{\delta}\right) + d\log\left(d|\gS|T\right) \right) },$$
    then for any $\beta\geq 0$, with any initial state distribution $\rho$, transition kernel estimator $f_0\in\gF$ and reference policy $\allpiref$, the regret of Algorithm~\ref{alg:markov_game} satisfies the following bound with probability at least $1-\delta$ for any $\delta\in(0,1)$:
    \begin{align}\label{eq:regret_MG_bd}
        \forall T\in\NN_+:\quad\regret(T)\leq \widetilde{\gO}\left(d \sqrt{H^3T}\cdot\sqrt{\frac{1}{d}\log\left(\frac{NH}{\delta}\right) + \log\left(d|\gS|T\right) 
        }\right).
    \end{align}
\end{thm}
Theorem~\ref{thm:regret_MG} establishes that by setting $\alpha$ on the order of $\widetilde{O}(\sqrt{T/H})$, with high probability, the regret of \name is no larger than an order of
$$ \widetilde{\gO}\left(d \sqrt{H^3T}\right)$$
for general-sum Markov games. When reducing to  two-player zero-sum Markov games, our  regret bound --- established for both players --- matches that of MEX~\citep{liu2024maximize}, which only covers  the max player. To the best of our knowledge, this is the first result that establishes a near-optimal sublinear regret for general-sum Markov games without explicit uncertainty quantification via constructing bonus functions or uncertainty sets. 

In addition, since
$    \min_{t\in[T]}\gap{\vpi_t}\leq  \regret(T) / T$ and each iteration collects $N+1$ trajectories, \name is guaranteed to find an $ \varepsilon$-NE ($\varepsilon$-CCE) of $\gM_\PP$ for any $\varepsilon>0$ within
$\widetilde{\gO}\left(\frac{ Nd^2H^3}{\varepsilon^2}\right)$ trajectories
or
$\widetilde{\gO}\left(\frac{ N d^2 H^4}{\varepsilon^2}\right)$ samples. Compared to the minimax sample complexity \citep{chen2022almost}, our sample complexity is near-optimal up to a factor of $H$ when the number of players $N$ is fixed.


\section{Conclusion}
\label{sec:conclusion}
In this paper, we introduced \name, a provably-efficient model-based algorithm for online MARL that balances  exploration and exploitation without requiring explicit uncertainty quantification. The key innovation lies in incentivizing the model estimation to maximize the best-response value functions across all players to implicitly drive exploration. In addition, \name is readily compatible with modern deep reinforcement learning architectures using function approximation, and is demonstrated to achieve a near-optimal regret under linear function approximation of the model class. We believe this work takes an important step toward making MARL more practical and scalable for real-world applications.

Several promising directions remain for future work. For example, designing a model-free counterpart of \name that can be used in conjunction with function approximation could be a valuable extension. Additionally, it will be interesting to develop the performance guarantee of \name under alternative assumptions of function approximation, such as general function approximation and independent function approximation across the players to tame the curse of dimensionality and multi-agency. Last but not least, it will be of interest to study the performance of \name under adversarial environments.

\section*{Acknowledgement}

This work is supported in part by the grants NSF DMS-2134080, CCF-2106778, and ONR N00014-19-1-2404.

{
\bibliographystyle{abbrvnat}
\bibliography{bibfileGame}
}

\appendix

\section{Special Cases}

\paragraph{Symmetric matrix game.}  One important special class of matrix games is the symmetric matrix game~\citep{cheng2004notes}, with $A=-A^\top$, $\muref=\nuref$, and $m=n$. In this case, we assume the parameter space $\Omega$ preserves anti-symmetry of $A$, i.e., $A_{\omega}  = -A_{\omega}^{\top}$ for any $\omega \in \Omega$. For a symmetric matrix game, it admits a symmetric Nash $(\mu^\star,\mu^\star)$, and Algorithm~\ref{alg:matrix_game}  reduces to a single-player algorithm by only tracking a single policy $\mu_t$, recognizing $\mu_t=\nu_t$ and $\widetilde{\mu}_t= \widetilde{\nu}_t$ due to $f^{\mu,\nu}(A)=-f^{\nu,\mu}(A)$. In addition,  \name only needs to collect one sample from the policy pair $(\widetilde{\mu}_t,\mu_t)$ in each iteration. Altogether, these lead to a simplified algorithm summarized in Algorithm~\ref{alg:symmetric_matrix_game}.

\begin{algorithm}[th]
    \caption{Value-incentivized Online Symmetric Matrix Game (\name)}
    \label{alg:symmetric_matrix_game}
    \begin{algorithmic}[1]
    \STATE \textbf{Input:}  initial parameter $\omega_0$, regularization coefficient $\alpha>0$, iteration number $T$.  
    \STATE \textbf{Initialization:} dataset $\gD_0\coloneqq \emptyset$.
    \FOR{$t = 1,\cdots,T$}
    \STATE 
    Compute $\mu_t$ by solving the matrix game with the current parameter $\omega_{t-1}$:
    \begin{align}
        \mu_t&=\argmax_{\mu\in\Delta^m}\min_{\nu\in\Delta^n} f^{\mu,\nu}(A_{\omega_{t-1}}) .
    \end{align}
    \STATE Model update: Update the parameter $\omega_t$ by minimizing the following objective:
    \begin{align}
        \omega_t=\argmin_{\omega\in\Omega}\sum_{(i,j,\widehat{A}(i,j))\in\gD_{t-1}}\left(A_\omega(i,j)-\widehat{A}(i,j)\right)^2+\alpha f^{\mu_t,\star}(A_{\omega}).
    \end{align}
    \STATE Compute $\widetilde{\mu}_t$  by solving the following optimization problem:
    \begin{align}
        \widetilde{\mu}_t& = \argmax_{\mu\in\Delta^m} f^{\mu,\mu_t}(A_{\omega_t}) .
    \end{align}
    \STATE Data collection: Sample $(i_t,j_t)\sim (\widetilde{\mu}_t,\mu_t)$  and get the noisy feedback $\widehat{A}(i_t,j_t)$ following the oracle \eqref{eq:noisy_feedback}. Update the dataset ${\gD}_t=\gD_{t-1}\cup\left\{ (i_t,j_t,\widehat{A}(i_t,j_t) ) \right\}$.
    \ENDFOR 
    \end{algorithmic}
\end{algorithm}

\paragraph{Bandit setting.} By setting $n=1$, we can reduce the matrix game to the bandit setting, where the payoff matrix becomes a reward vector $ A \in \mathbb{R}^{m}$, leading to a simplified algorithm in Algorithm~\ref{alg:bandit}. Here, we let $f^{\mu}(A) = \mu^{\top} A - \beta\KL{\mu}{\muref}$ and 
     $f^{\star}(A)\coloneqq \max_{\mu\in\Delta^m} f^{\mu}(A)$. Interestingly, to encourage exploration, the regularization term favors a reward estimate that maximizes its regret $  f^{\star}(A_{\omega}) -  f^{\mu_t}(A_{\omega}).$ on the current policy $\mu_t$, which is {\em different from} the reward-biasing framework that only regularizes against $  f^{\star}(A_{\omega})$ \citep{cen2024value,liu2020exploration}.   
     
\begin{algorithm}[th]
    \caption{Value-incentivized Online Bandit (\name)}
    \label{alg:bandit}
    \begin{algorithmic}[1]
    \STATE \textbf{Input:}  initial parameter $\omega_0$, regularization coefficient $\alpha>0$, iteration number $T$.  
    \STATE \textbf{Initialization:} dataset $\gD_0\coloneqq \emptyset$.
    \FOR{$t = 1,\cdots,T$}
    \STATE 
    Policy update: Compute $\mu_t$ with the current parameter $\omega_{t-1}$:
    \begin{align}
        \mu_t&=\argmax_{\mu\in\Delta^m}  f^{\mu}(A_{\omega_{t-1}})  \propto \muref \exp\left( \frac{A_{\omega_{t-1}} }{ \beta } \right).
    \end{align}
        \STATE Data collection: Sample $i_t \sim \mu_t $  and get the noisy feedback $\widehat{A}(i_t)$  following the oracle \eqref{eq:noisy_feedback}. Update the dataset ${\gD}_t=\gD_{t-1}\cup\left\{ (i_t,\widehat{A}(i_t) ) \right\}$.
   
    \STATE Model update: Update the parameter $\omega_t$ by minimizing the following objective:  
    \begin{align}
        \omega_t=\argmin_{\omega\in\Omega}\sum_{(i,\widehat{A}(i))\in\gD_{t}}\left(A_\omega(i)-\widehat{A}(i)\right)^2 -\alpha f^{\star}(A_{\omega})+\alpha f^{\mu_t}(A_{\omega}).
    \end{align}

    \ENDFOR 
    \end{algorithmic}
\end{algorithm}

\paragraph{MDP setting.}  \name can be reduced to the single-agent setting via either setting the number of players $N=1$  in the multi-player general-sum Markov game, or setting the action space of the min player  to a singleton, i.e., $|\gA_2|=1$, in the two-player zero-sum Markov game. Interestingly, the former (option I) leads to the value regularization $V_{f}^{\star}(\rho)$ studied in MEX (c.f., Algorithm 1 in \citet{liu2024maximize}), while the latter (option II) leads to a new form of regularizer $V_{f}^{\star}(\rho)-V_{f}^{\pi_t}(\rho)$, adding friction from the current policy $\pi_t$. The latter regularizer is also the MDP counterpart of the bandit algorithm in Algorithm~\ref{alg:bandit}. We summarize both variants in Algorithm~\ref{alg:mdp}.

\begin{algorithm}[!ht]
    \caption{Value-incentivized Online Single-agent MDP (\name)}
    \label{alg:mdp}
    \begin{algorithmic}[1]
    \STATE \textbf{Input:}  initial transition kernel estimate $f_0\in\gF$, regularization coefficient $\alpha>0$, iteration number $T$.  
    \STATE \textbf{Initialization:} dataset $\gD_{0,h}\coloneqq \emptyset$ for all $h\in[H]$.
    \FOR{$t = 1,\cdots,T$}
    \STATE 
    Policy update: Compute $\pi_t$ with the current transition kernel estimator $f_{t-1}$:
    \begin{align}
        \pi_t&=\argmax_{\pi\in\Delta(\gA_1)} V^\pi_{f_{t-1}}(\rho).
    \end{align} 
        \STATE Data collection: sample a trajectory with transition tuples $\{(s_{t,h},a_{t,h},s_{t,h+1})\}_{h=1}^H$ following $\pi_t$. Update the dataset $\gD_{t,h}=\gD_{t-1,h}\cup\left\{ (s_{t,h},a_{t,h},s_{t,h+1}) \right\}$ for all $h\in[H]$.

    \STATE Model update: update the estimator $f_t$ by minimizing the following objective
        \begin{align}
    f_t=   \begin{cases} 
    \argmin_{f\in\gF}\sum_{h=1}^H\sum_{(s_h,a_h,s_{h+1})\in\gD_{t,h}}-\log \PP_{f,h}(s_{h+1}|s_h,a_h) -\alpha V_f^\star(\rho)  &  \text{(option I)} \\
    \argmin_{f\in\gF}\sum_{h=1}^H\sum_{(s_h,a_h,s_{h+1})\in\gD_{t,h}}-\log \PP_{f,h}(s_{h+1}|s_h,a_h) -\alpha V_f^\star(\rho)+\alpha V_{f}^{\pi_t}(\rho) & \text{(option II)} 
    \end{cases}
     .
    \end{align}
 
    \ENDFOR 
    \end{algorithmic}
\end{algorithm}

%

\section{Proofs of Main Theorems}\label{app:proofs}

\subsection{Auxiliary lemmas}\label{app:lemmas}

We provide some technical lemmas that will be used in our proofs.

\begin{lm}[Freedman's inequality, Lemma D.2 in \citet{liu2024maximize}]
    \label{lm:Freedman}
    Let $\{X_t\}_{t\leq T}$ be a real-valued martingale difference sequence adapted to filtration $\{\mathcal{F}_t\}_{t\leq T}$. If $|X_t| \leq R$ almost surely, then for any $\eta \in (0,1/R)$ it holds that with probability at least $1-\delta$,
    $$\sum_{t=1}^T X_t \leq \gO\left(\eta\sum_{t=1}^T \mathbb{E}[X_t^2|\mathcal{F}_{t-1}] + \frac{\log(1/\delta)}{\eta}\right).$$
\end{lm}

\begin{lm}[Lemma 11 in \citet{abbasi2011improved}]\label{lm:potential}
    Let $\{x_s\}_{s\in[T]}$ be a sequence of vectors with $x_s \in \mathcal{V}$ for some Hilbert space $\mathcal{V}$. Let $\Lambda_0$ be a positive definite matrix and define $\Lambda_t = \Lambda_0 + \sum_{s=1}^t x_sx_s^\top$. Then it holds that
    \begin{align*}
        \sum_{s=1}^T \min\left\{1,\|x_s\|_{\Lambda_{s-1}^{-1}}\right\} \leq 2\log\left(\frac{\det(\Lambda_{T})}{\det(\Lambda_0)}\right).
    \end{align*}
\end{lm}

\begin{lm}[Lemma F.3 in \citet{du2021bilinear}]\label{lm:information_gain}
    Let $\gX\subset \R^d$ and $\sup_{x\in\gX}\norm{x}_2\leq B_X$. Then for any $n\in\NN_+$, we have
    \begin{align*}
        \forall \lambda>0:\quad\max_{x_1,\cdots,x_n\in\gX}\log \det\left(I_d+\frac{1}{\lambda}\sum_{i=1}^n x_i x_i^\top\right)\leq d\log\left(1+\frac{nB_X^2}{d\lambda}\right).
    \end{align*}
\end{lm}

\begin{lm}[Martingale exponential concentration, Lemma D.1 in \citet{liu2024maximize}]\label{lm:martingale_exp}
Let $\delta \in (0,1)$. For a sequence of real-valued random variables $\{X_t\}_{t\in[T]}$ adapted to filtration $\{\mathcal{F}_t\}_{t\in[T]}$, the following holds with probability at least $1-\delta$:
    \begin{align*}
        \forall t\in[T]: \quad -\sum_{s=1}^t X_s \leq \sum_{s=1}^t\log\E[\exp(-X_s)|\mathcal{F}_{s-1}] + \log(1/\delta).
    \end{align*}
\end{lm}

\begin{lm}[Covering number of $\ell_2$ ball, Lemma D.5 in \citet{jin2020provably}]
    \label{lm:covering}
    For any $\epsilon>0$ and $d\in\NN_+$, the $\epsilon$-covering number of the $\ell_2$ ball of radius $R$ in $\R^d$ is bounded by $(1+2R/\epsilon)^d$.
\end{lm}

\subsection{Proof of Theorem~\ref{thm:regret}}\label{app:proof_regret}
In the proof, for any sequence $\{x_i\}_{i \in \mathbb{Z}}$ and any integers $a,b \in \mathbb{Z}$ where $a > b$, we define $\sum_{i=a}^b x_i \coloneqq 0$.
 
We begin by decomposing the regret as
\begin{align}\label{eq:regret_dec1}
    \regret(T) &= \sum_{t=1}^T f^{\star,\nu_t}(A) - f^{\mu_t,\star}(A)\notag\\
    &= \sum_{t=1}^T \left( f^{\star,\nu_t}(A) - f^{\mu_t,\star}(A) - \left(f^{\star,\nu_t}(A_{\omega_t}) - f^{\mu_t,\star}(A_{\omega_t}) \right)\right) \notag\\
    &\quad + \sum_{t=1}^T \left(f^{\star,\nu_t}(A_{\omega_t})-f^{\widetilde{\mu}_t,\nu_t}(A)\right) + \sum_{t=1}^T \left(f^{\mu_t,\widetilde{\nu}_t}(A)-f^{\mu_t,\star}(A_{\omega_t})\right)\notag\\
    &\quad + \sum_{t=1}^T \left(f^{\widetilde{\mu}_t,\nu_t}(A)-f^{\widetilde{\mu}_t,\nu_t}(A_{\omega_{t-1}})\right) + \sum_{t=1}^T \left(f^{\mu_t,\widetilde{\nu}_t}(A_{\omega_{t-1}})-f^{\mu_t,\widetilde{\nu}_t}(A)\right)\notag\\
    &\quad + \sum_{t=1}^T \left(f^{\widetilde{\mu}_t,\nu_t}(A_{\omega_{t-1}})-f^{\mu_t,\widetilde{\nu}_t}(A_{\omega_{t-1}})\right).
\end{align}

Recall that $(\mu_t,\nu_t)$ is the Nash equilibrium of the matrix game with the pay-off matrix $A_{\omega_{t-1}}$ (see \eqref{eq:policy_update}), we have
\begin{align}\label{eq:last_term_nash}
    \forall t\in[T]:\quad f^{\widetilde{\mu}_t,\nu_t}(A_{\omega_{t-1}}) \leq f^{\mu_t,\nu_t}(A_{\omega_{t-1}})\leq f^{\mu_t,\widetilde{\nu_t}}(A_{\omega_{t-1}}).
\end{align}
This implies the last term in the regret decomposition is non-positive, i.e.,
\begin{align}\label{eq:last_term_nonpos}
    \sum_{t=1}^T \left(f^{\widetilde{\mu}_t,\nu_t}(A_{\omega_{t-1}})-f^{\mu_t,\widetilde{\nu}_t}(A_{\omega_{t-1}})\right)\leq 0.
\end{align}

Moreover, by the definition of $\widetilde{\mu}_t$ and $\widetilde{\nu}_t$ in \eqref{eq:policy_update_2}, we have
\begin{align}\label{eq:dagger_equation}
    f^{\star,\nu_t}(A_{\omega_t})=f^{\widetilde{\mu}_t,\nu_t}(A_{\omega_t})\quad\text{and}\quad f^{\mu_t,\star}(A_{\omega_t})=f^{\mu_t,\widetilde{\nu}_t}(A_{\omega_t}).
\end{align}

Combining \eqref{eq:last_term_nonpos}, \eqref{eq:dagger_equation} with \eqref{eq:regret_dec1}, we have
\begin{align}\label{eq:regret_decomposition}
    \regret(T) &\leq \underbrace{\sum_{t=1}^T \left( f^{\star,\nu_t}(A) - f^{\mu_t,\star}(A) - \left(f^{\star,\nu_t}(A_{\omega_t}) - f^{\mu_t,\star}(A_{\omega_t}) \right)\right) }_{\text{(i)}}\notag\\
    &\quad + \underbrace{\sum_{t=1}^T \left(f^{\widetilde{\mu}_t,\nu_t}(A_{\omega_t})-f^{\widetilde{\mu}_t,\nu_t}(A)\right) + \sum_{t=1}^T \left(f^{\mu_t,\widetilde{\nu}_t}(A)-f^{\mu_t,\widetilde{\nu}_t}(A_{\omega_t})\right)}_{\text{(ii)}}\notag\\
    &\quad + \underbrace{\sum_{t=1}^T \left(f^{\widetilde{\mu}_t,\nu_t}(A)-f^{\widetilde{\mu}_t,\nu_t}(A_{\omega_{t-1}})\right) + \sum_{t=1}^T \left(f^{\mu_t,\widetilde{\nu}_t}(A_{\omega_{t-1}})-f^{\mu_t,\widetilde{\nu}_t}(A)\right)}_{\text{(iii)}}.
\end{align}

We will upper bound the three terms in the right-hand side of \eqref{eq:regret_decomposition} separately.


\paragraph{Step 1: bounding term (i).} Define the squared loss function $L_t(\omega)$ over the dataset $\gD_{t-1}$ as
\begin{align}\label{eq:L}
    L_t(\omega)\coloneqq \sum_{(i,j,\widehat{A}(i,j))\in\gD_{t-1}}\left(A_\omega(i,j)-\widehat{A}(i,j)\right)^2.
\end{align}
Then by the optimality of $\omega_t$ for \eqref{eq:model_update}, we know that
\begin{align*}
    L_t(\omega_t) - \alpha f^{\star,\nu_t}(A_{\omega_t}) + \alpha f^{\mu_t,\star}(A_{\omega_t})\leq L_t(\omega^\star) - \alpha f^{\star,\nu_t}(A) + \alpha f^{\mu_t,\star}(A),
\end{align*}
where we use Assumption~\ref{asmp:expressive}, which implies $A_{\omega^\star}=A$.
Reorganizing the terms, we have
\begin{align}\label{eq:(i)_ub_by_L}
    \text{(i)}\leq \frac{1}{\alpha}\sum_{t=1}^T\left(L_t(\omega^\star)-L_t(\omega_t)\right).
\end{align}
Thus, it is sufficient to bound the term $\sum_{t=1}^T\left(L_t(\omega^\star)-L_t(\omega_t)\right)$. 
For any $t\in[T]$, we denote
$$\widehat{A}(i_t,j_t)=A(i_t,j_t)+\xi_t\quad{and}\quad  \widehat{A}(i'_t,j'_t)=A(i_t,j_t)+\xi'_t.$$
It follows that we can rewrite $L_t(\omega)$ as
\begin{align*}
    L_t(\omega)&=\sum_{s=1}^{t-1} \left(A_\omega(i_s,j_s)-A(i_s,j_s)-\xi_s\right)^2+\sum_{s=1}^{t-1} \left(A_\omega(i'_s,j'_s)-A(i'_s,j'_s)-\xi'_s\right)^2,
\end{align*}
from which we deduce
\begin{align}\label{eq:L_decomposition}
    L_t(\omega^\star)-L_t(\omega) 
    &=-\sum_{s=1}^{t-1} \left[\left(A_\omega(i_s,j_s)-A(i_s,j_s)-\xi_s\right)^2 - \xi_s^2\right]
    -\sum_{s=1}^{t-1} \left[\left(A_\omega(i'_s,j'_s)-A(i'_s,j'_s)-\xi'_s\right)^2 - (\xi'_s)^2\right]\notag\\ 
    &= -\sum_{s=1}^{t-1} \underbrace{\left[\left(A_\omega(i_s,j_s)-A(i_s,j_s)\right)\left(A_\omega(i_s,j_s)-A(i_s,j_s)-2\xi_s\right)\right]}_{\coloneqq X_s^\omega}\notag\\
    &\quad-\sum_{s=1}^{t-1} \underbrace{\left[\left(A_\omega(i'_s,j'_s)-A(i'_s,j'_s)\right)\left(A_\omega(i'_s,j'_s)-A(i'_s,j'_s)-2\xi'_s\right)\right]}_{\coloneqq Y_s^\omega}.
\end{align}

It is then sufficient to bound $-\sum_{s=1}^{t-1} X_s^\omega$ and $-\sum_{s=1}^{t-1} Y_s^\omega$, which is supplied by the following lemma.  
\begin{lm}\label{lm:regret_on_squared_loss} 
When Assumption~\ref{asmp:bounded_payoff}, \ref{asmp:expressive}, \ref{asmp:noise} hold, for any $\delta \in (0,1)$, with probability at least $1-\delta$, it holds  
\begin{align}\label{eq:sum_X_bound}
    \forall t\in[T],\omega\in\Omega:\quad -\sum_{s=1}^{t-1} X_s^\omega &\leq - \frac{1}{2} \sum_{s=1}^{t-1}\E_{i\sim\widetilde{\mu}_s,j\sim\nu_s}\left[\left(A_\omega(i,j)-A(i,j)\right)^2\right]\notag\\
    &\quad+ C B_l\left(B_l+\sigma\sqrt{2\log(8T/\delta)}\right)  \left(\log\left(\frac{4T}{\delta}\right)+d\log\left(1+2T\sqrt{d}\right)\right) 
\end{align}
and
\begin{align}\label{eq:sum_Y_bound}
    \forall t\in[T],\omega\in\Omega:\quad -\sum_{s=1}^{t-1} Y_s^\omega &\leq - \frac{1}{2} \sum_{s=1}^{t-1}\E_{i\sim\mu_s,j\sim\widetilde{\nu}_s}\left[\left(A_\omega(i,j)-A(i,j)\right)^2\right]\notag\\
    &\quad+ C B_l\left(B_l+\sigma\sqrt{2\log(8T/\delta)}\right)  \left(\log\left(\frac{4T}{\delta}\right)+d\log\left(1+2T\sqrt{d}\right)\right)
\end{align}
where $C>0$ is some universal constant.
\end{lm}

Combining \eqref{eq:L_decomposition}, \eqref{eq:(i)_ub_by_L} and Lemma~\ref{lm:regret_on_squared_loss} leads to a bound of term (i):
\begin{align}\label{eq:bound_i}
    \text{(i)}\leq \frac{1}{\alpha}\bigg\{-\frac{1}{2} \sum_{t=1}^T\sum_{s=1}^{t-1}\E_{i\sim\widetilde{\mu}_s,j\sim\nu_s}\left[\left(A_{\omega_t}(i,j)-A(i,j)\right)^2\right]
    -  \frac{1}{2} \sum_{t=1}^T\sum_{s=1}^{t-1}\E_{i\sim\mu_s,j\sim\widetilde{\nu}_s}\left[\left(A_{\omega_t}(i,j)-A(i,j)\right)^2\right]\notag\\
    +2T\cdot C B_l\left(B_l+\sigma\sqrt{2\log(8T/\delta)}\right)  \left(\log\left(\frac{4T}{\delta}\right)+d\log\left(1+2T\sqrt{d}\right)\right)\bigg\}.
\end{align}

\paragraph{Step 2: bounding terms (ii) and (iii).} To bound (ii) and (iii), we first prove the following lemma.
\begin{lm}\label{lm:diff_f}
    For any $\{(\widehat{\mu}_t,\widehat{\nu}_t)\}_{t\in[T]}\subset \Delta^m\times\Delta^n$ and any $\{\widehat{\omega}_t\}_{t\in[T]}\subset\Omega$, we have
    \begin{align}\label{eq:diff_f}
        \sum_{t=1}^T \left| f^{\widehat{\mu}_t,\widehat{\nu}_t}(A_{\widehat{\omega}_t}) - f^{\widehat{\mu}_t,\widehat{\nu}_t}(A)\right| 
    &\leq \frac{d(\lambda)}{2\eta} +\frac{\eta}{2}\sum_{t=1}^T \sum_{s=1}^{t-1} \E_{i\sim\widehat{\mu}_s,j\sim\widehat{\nu}_s} [(A_{\widehat{\omega}_t}(i,j)-A(i,j))^2]  \nonumber\\ 
    & \quad + (\sqrt{d}+2B_l) \min\{d(\lambda),T\} +\sqrt{d} \lambda
    \end{align}
    for any $\lambda,\eta>0$, where $d(\lambda)\coloneqq 2d\log\left(1+\frac{T}{d\lambda}\right)$.
\end{lm}

By letting $\widehat{\mu}_t=\widetilde{\mu}_t$, $\widehat{\nu}_t=\nu_t$ and $\widehat{\omega}_t=\omega_t$ in Lemma~\ref{lm:diff_f}, we have
\begin{align}\label{eq:bound_ii_1}
    \sum_{t=1}^T \left(f^{\widetilde{\mu}_t,\nu_t}(A_{\omega_t})-f^{\widetilde{\mu}_t,\nu_t}(A)\right)
    &\leq \frac{d(\lambda)}{2\eta} +\frac{\eta}{2}\sum_{t=1}^T \sum_{s=1}^{t-1} \E_{i\sim\widetilde{\mu}_s,j\sim\nu_s} [(A_{\omega_t}(i,j)-A(i,j))^2] \notag\\
    & \quad+ (\sqrt{d}+2B_l) \min\{d(\lambda),T\} +\sqrt{d} \lambda T. 
\end{align}

By letting $\widehat{\mu}_t=\mu_t$, $\widehat{\nu}_t=\widetilde{\nu}_t$ and $\widehat{\omega}_t=\omega_t$ in Lemma~\ref{lm:diff_f}, we have 
\begin{align}\label{eq:bound_ii_2}
    \sum_{t=1}^T \left(f^{\mu_t,\widetilde{\nu}_t}(A)-f^{\mu_t,\widetilde{\nu}_t}(A_{\omega_t})\right)     &\leq \frac{d(\lambda)}{2\eta} +\frac{\eta}{2}\sum_{t=1}^T \sum_{s=1}^{t-1} \E_{i\sim\mu_s,j\sim\widetilde{\nu}_s} [(A_{\omega_t}(i,j)-A(i,j))^2] \nonumber \\
    & \quad + (\sqrt{d}+2B_l) \min\{d(\lambda),T\} +\sqrt{d} \lambda T. 
\end{align}

Similarly, we have  
\begin{align}\label{eq:bound_iii_1a}
     \sum_{t=1}^T \left(f^{\widetilde{\mu}_t,\nu_t}(A)-f^{\widetilde{\mu}_t,\nu_t}(A_{\omega_{t-1}})\right)&\leq \frac{d(\lambda)}{2\eta} +\frac{\eta}{2}\sum_{t=1}^T \sum_{s=1}^{t-2} \E_{i\sim\widetilde{\mu}_s,j\sim\nu_s} [(A_{\omega_{t-1}}(i,j)-A(i,j))^2]  \nonumber \\
    &\quad + (\sqrt{d}+2B_l) \min\{d(\lambda),T\} +\sqrt{d} \lambda T + 2B_l^2\eta T,
 \end{align}
 which uses  the fact  
$$\E_{i\sim\widetilde{\mu}_{t-1},j\sim\nu_{t-1}} [(A_{\omega_{t-1}}(i,j)-A(i,j))^2]\leq 4B_l^2.$$
Notice that the second term in \eqref{eq:bound_iii_1a} can be further bounded by
 \begin{align}   
   \sum_{t=1}^T \sum_{s=1}^{t-2} \E_{i\sim\widetilde{\mu}_s,j\sim\nu_s} [(A_{\omega_{t-1}}(i,j)-A(i,j))^2]   &= \sum_{t=0}^{T-1} \sum_{s=1}^{t-1} \E_{i\sim\widetilde{\mu}_s,j\sim\nu_s} [(A_{\omega_{t}}(i,j)-A(i,j))^2]   \notag\\
    &\leq \sum_{t=0}^{T}  \sum_{s=1}^{t-1} \E_{i\sim\widetilde{\mu}_s,j\sim\nu_s} [(A_{\omega_{t}}(i,j)-A(i,j))^2]  \notag\\
    &= \sum_{t=1}^{T} \sum_{s=1}^{t-1} \E_{i\sim\widetilde{\mu}_s,j\sim\nu_s} [(A_{\omega_{t}}(i,j)-A(i,j))^2] ,
\end{align}
where the first line shifts the index of $t$ by 1, and the last equality holds 
because the term is $0$ when $t=0$. Plugging the above inequality back to \eqref{eq:bound_iii_1a} leads to
\begin{align}\label{eq:bound_iii_1}
     \sum_{t=1}^T \left(f^{\widetilde{\mu}_t,\nu_t}(A)-f^{\widetilde{\mu}_t,\nu_t}(A_{\omega_{t-1}})\right) &= \frac{d(\lambda)}{2\eta} +\frac{\eta}{2}\sum_{t=1}^{T} \sum_{s=1}^{t-1} \E_{i\sim\widetilde{\mu}_s,j\sim\nu_s} [(A_{\omega_{t}}(i,j)-A(i,j))^2] \nonumber \\
   &\quad  + (\sqrt{d}+2B_l) \min\{d(\lambda),T\} +\sqrt{d} \lambda T + 2B_l^2\eta T.
\end{align}
Analogously, we have
\begin{align}\label{eq:bound_iii_2}
    \sum_{t=1}^T \left(f^{\mu_t,\widetilde{\nu}_t}(A_{\omega_{t-1}})-f^{\mu_t,\widetilde{\nu}_t}(A)\right) &\leq \frac{d(\lambda)}{2\eta} +\frac{\eta}{2}\sum_{t=1}^{T} \sum_{s=1}^{t-1} \E_{i\sim\mu_s,j\sim\widetilde{\nu}_s} [(A_{\omega_{t}}(i,j)-A(i,j))^2]  \notag\\
    &\quad  + (\sqrt{d}+2B_l) \min\{d(\lambda),T\} +\sqrt{d} \lambda T + 2B_l^2\eta T.
\end{align}

Combining \eqref{eq:bound_ii_1}, \eqref{eq:bound_ii_2}, \eqref{eq:bound_iii_1}, and \eqref{eq:bound_iii_2}, we have \begin{align}\label{eq:bound_ii}
    \text{(ii)}+\text{(iii)}\leq  
    \eta \left\{\sum_{t=1}^T \sum_{s=1}^{t-1} \E_{i\sim\widetilde{\mu}_s,j\sim\nu_s} [(A_{\omega_t}(i,j)-A(i,j))^2] + \sum_{t=1}^T \sum_{s=1}^{t-1} \E_{i\sim\mu_s,j\sim\widetilde{\nu}_s} [(A_{\omega_t}(i,j)-A(i,j))^2]\right\}\notag\\
    +\frac{2d(\lambda)}{\eta} +4(\sqrt{d}+2B_l) \min\{d(\lambda),T\} +4\sqrt{d} \lambda T + 4B_l^2\eta T.
\end{align}


\paragraph{Step 3: combining the bounds.} 
Letting $\eta=\frac{1}{2\alpha}$ in \eqref{eq:bound_ii}, the first line of \eqref{eq:bound_i} could cancel out the first line of \eqref{eq:bound_ii}, which leads to
\begin{align}
    \regret(T) &= \text{(i)}+\text{(ii)}+\text{(iii)}\notag\\
    &= \frac{T}{\alpha}\cdot2C B_l\left(B_l+\sigma\sqrt{2\log(8T/\delta)}\right)  \left(\log\left(\frac{4T}{\delta}\right)+d\log\left(1+2T\sqrt{d}\right)\right)\notag\\
    &\quad + 4\alpha d(\lambda) +4(\sqrt{d}+2B_l) \min\{d(\lambda),T\} +4\sqrt{d} \lambda T + \frac{2B_l^2T}{\alpha}
\end{align}
with probability at least $1-\delta$.

By choosing
$$\alpha=\sqrt{\frac{T\left(\log(4T/\delta)+d\log\left(1+2\sqrt{d}T\right)\right)}{d\log\left(1+(T/d)^{3/2}\right)}}\quad\text{and}\quad \lambda=\sqrt{\frac{d}{T}},$$
we have with probability at least $1-\delta$,
\begin{align}\label{eq:regret_bd_complete}
    &\regret(T)\notag\\ &\leq  
2\left(CB_l\left(B_l+\sigma\sqrt{2\log(8T/\delta)}\right)+1\right)d\sqrt{T}\cdot\sqrt{\left(\frac{1}{d}\log(4T/\delta)+\log\left(1+2\sqrt{d}T\right)\right)\log\left(1+(T/d)^{3/2}\right)}\notag\\
    &\quad+ 2B_l^2\sqrt{T}\sqrt{\frac{d\log\left(1+(T/d)^{3/2}\right)}{\log(4T/\delta)+d\log\left(1+2\sqrt{d}T\right)}}+4(\sqrt{d}+2B_l) d\log\left(1+(T/d)^{3/2}\right)+4d\sqrt{T}
\end{align}
for some absolute constant $C>0$,
and thus the regret could be bounded by \eqref{eq:regret_bound} by simplifying the logarithmic terms.

\subsubsection{Proof of Lemma~\ref{lm:regret_on_squared_loss}}

To begin, by Assumption~\ref{asmp:noise} together with the sub-Gaussian concentration inequality, we  have that with probability at least $1-\frac{\delta}{2}$, for any $s\in[T]$ and $\omega\in\Omega$,
\begin{align*}
    \PP(|\xi_s|\geq a)\leq 2\exp\left(-\frac{a^2}{2\sigma^2}\right)\quad\text{and}\quad \PP(|\xi'_s|\geq a)\leq 2\exp\left(-\frac{a^2}{2\sigma^2}\right)
    \quad\text{for all }a>0,
\end{align*}
which implies that with probability at least $1-\frac{\delta}{2}$,
\begin{align}\label{eq:bound_xi}
    |\xi_s|\leq \sigma\sqrt{2\log(8T/\delta)},\quad |\xi'_s|\leq \sigma\sqrt{2\log(8T/\delta)},\quad\forall s\in[T].
\end{align} 
We let  $\mathcal{E}$ be the event that \eqref{eq:bound_xi} holds for all $s\in[T]$, which satisfies $\PP(\mathcal{E})\geq 1-\frac{\delta}{2}$.

Next, we define filtrations $\gF_t\coloneqq \sigma(\gD_{t})$ for all $t\in[T]$. By Assumption~\ref{asmp:noise}, we have for all $s\in[T]$ and $\omega\in\Omega$,
\begin{align}
    \E[X_s^\omega|  \gF_{s-1}]&=\E_{i\sim\widetilde{\mu}_s,j\sim\nu_s}\left[\left(A_\omega(i,j)-A(i,j)\right)^2\right],\label{eq:conditional_expectation_X}\\
    \E[Y_s^\omega|\gF_{s-1}]&=\E_{i\sim\mu_s,j\sim\widetilde{\nu}_s}\left[\left(A_\omega(i,j)-A(i,j)\right)^2\right]\label{eq:conditional_expectation_Y}.
\end{align}

We also have
\begin{align}\label{eq:conditional_var}
    \var\left[X_s^\omega|\gF_{s-1}\right]&\leq \E\left[(X_s^\omega)^2|\gF_{s-1}\right]\notag\\
    &=\E\left[\left(A_\omega(i_s,j_s)-A(i_s,j_s)\right)^2\left(A_\omega(i_s,j_s)-A(i_s,j_s)-2\xi_s\right)^2|\gF_{s-1}\right]\notag\\
    &\leq 4(B_l^2+\sigma^2)\E\left[\left(A_\omega(i_s,j_s)-A(i_s,j_s)\right)^2|\gF_{s-1}\right]\notag\\
    &=4(B_l^2+\sigma^2)\E_{i\sim\widetilde{\mu}_s,j\sim\nu_s}\left[\left(A_\omega(i,j)-A(i,j)\right)^2\right],
\end{align}
where we use Assumptions~\ref{asmp:bounded_payoff}, \ref{asmp:expressive} and \ref{asmp:noise} in the last inequality: Assumption~\ref{asmp:expressive} guarantees that $A_{\omega^\star}=A$, and Assumption~\ref{asmp:bounded_payoff} indicates that $\norm{A(i,j)}_\infty\coloneqq\max_{i\in[m],j\in[n]}|A(i,j)|\leq B_l$ and $\norm{A_\omega(i,j)}_\infty\leq B_l$ for all $\omega\in\Omega$; moreover, Assumption \ref{asmp:noise} implies $\E\xi_s^2\leq \sigma^2$. 
Conditioned on event $\mathcal{E}$,
we can bound $|X_s^\omega-\E[X_s^\omega|\gF_{s-1}]|$ using \eqref{eq:L_decomposition} and \eqref{eq:conditional_expectation_X} as follows:
\begin{align}\label{eq:bound_abs_X}
    & \left|X_s^\omega-\E[X_s^\omega|\gF_{s-1}]\right|\notag\\
    &=\left|\left(A_\omega(i_s,j_s)-A(i_s,j_s)\right)\left(A_{\omega}(i_s,j_s)-A(i_s,j_s)-2\xi_s\right)-\E_{i\sim\widetilde{\mu}_s,j\sim\nu_s}\left[\left(A_{\omega}(i,j)-A(i,j)\right)^2\right]\right|\notag\\
    &\leq \left|\left(A_\omega(i_s,j_s)-A(i_s,j_s)\right)^2-\E_{i\sim\widetilde{\mu}_s,j\sim\nu_s}\left[\left(A_\omega(i,j)-A(i,j)\right)^2\right]\right|+2|\xi_s|\left|A_\omega(i_s,j_s)-A(i_s,j_s)\right|\notag\\
    &\leq 4B_l\left(B_l+\sigma\sqrt{2\log(8T/\delta)}\right).
\end{align}

In what follows, we apply a standard covering argument together with the Freedman's inequality to prove the desired bound, conditioned on event $\mathcal{E}$. First, for any $\gX\subset\R^d$, let $\gN(\gX,\epsilon,\norm{\cdot})$ be the $\epsilon$-covering number of $\gX$ with respect to the norm $\norm{\cdot}$. 
By Assumption~\ref{asmp:bounded_payoff} we know that $\Omega\subset\mathbb{B}_2^d(\sqrt{d})$. Thus by Lemma~\ref{lm:covering} we have
\begin{align}\label{eq:covering}
    \log\gN(\Omega, \epsilon, \norm{\cdot}_2)\leq \log\gN(\mathbb{B}_2^d(\sqrt{d}), \epsilon, \norm{\cdot}_2)\leq d\log\left(1+\frac{2\sqrt{d}}{\epsilon}\right)
\end{align}
for any $\epsilon>0$.
In other words, for any $\epsilon>0$, there exists an $\epsilon$-net $\Omega_\epsilon\subset\Omega$ such that $\log|\Omega_\epsilon|\lesssim d\log\left(1+\frac{2\sqrt{d}}{\epsilon}\right)$.

Applying Freedman's inequality (c.f. Lemma~\ref{lm:Freedman}) to the martingale difference sequence $\{\E[X_s^\omega|\gF_{s-1}]-X_s^\omega\}_{s\in[T]}$ and making use of \eqref{eq:conditional_expectation_X}, \eqref{eq:conditional_var} and \eqref{eq:bound_abs_X} we have under event $\mathcal{E}$, with probability at least $1-\frac{\delta}{4}$,
\begin{align*}
    \forall t\in[T],\,\omega\in\Omega_\epsilon: \qquad &  \sum_{s=1}^t\left(\E_{i\sim\widetilde{\mu}_s,j\sim\nu_s}\left[\left(A_\omega(i,j)-A(i,j)\right)^2\right] - X_s^\omega\right)\notag\\
    &\leq\frac{1}{2} \sum_{s=1}^t\E_{i\sim\widetilde{\mu}_s,j\sim\nu_s}\left[\left(A_\omega(i,j)-A(i,j)\right)^2\right]\notag\\
    &\quad+ 4C B_l\left(B_l+\sigma\sqrt{2\log(8T/\delta)}\right) \left(\log\left(\frac{4T}{\delta}\right)+d\log\left(1+\frac{2\sqrt{d}}{\epsilon}\right)\right),
\end{align*}
where $C>0$ is an absolute constant.

In addition, conditioned on event $\mathcal{E}$, 
for any $\omega,\omega'\in\Omega$, $\norm{\omega-\omega'}_2\leq \epsilon$, we have
\begin{align*}
& \left|\left(\frac{1}{2}\E_{i\sim\widetilde{\mu}_s,j\sim\nu_s}\left[\left(A_\omega(i,j)-A(i,j)\right)^2\right] - X_s^\omega\right)-\left(\frac{1}{2}\E_{i\sim\widetilde{\mu}_s,j\sim\nu_s}\left[\left(A_{\omega'}(i,j)-A(i,j)\right)^2\right] - X_s^{\omega'}\right)\right|\notag\\
&\leq \frac{1}{2}\E_{i\sim\widetilde{\mu}_s,j\sim\nu_s}\left|\left(A_\omega(i,j)-A(i,j)\right)^2-\left(A_{\omega'}(i,j)-A(i,j)\right)^2\right|+|X_s^\omega-X_s^{\omega'}|\notag\\
&\leq \left(6B_l+2\sigma\sqrt{2\log(8T/\delta)}\right)\epsilon.
\end{align*}
Thus combining the above two expressions and set $\epsilon=\frac{1}{T}$, we have that under event $\mathcal{E}$, with probability at least $1-\frac{\delta}{4}$,
\begin{align}
    \forall t\in[T],\omega\in\Omega:&\quad \sum_{s=1}^t\left(\E_{i\sim\widetilde{\mu}_s,j\sim\nu_s}\left[\left(A_\omega(i,j)-A(i,j)\right)^2\right] - X_s^\omega\right)\notag\\
    &\leq\frac{1}{2} \sum_{s=1}^t\E_{i\sim\widetilde{\mu}_s,j\sim\nu_s}\left[\left(A_\omega(i,j)-A(i,j)\right)^2\right]\notag\\
    &\quad+ 4C B_l\left(B_l+\sigma\sqrt{2\log(8T/\delta)}\right)  \left(\log\left(\frac{4T}{\delta}\right)+d\log\left(1+2T\sqrt{d}\right)\right)  
    \end{align}
for sufficiently large constant $C$.

Rearanging terms, we have with probability at least $1-\frac{\delta}{4}$,
\begin{align}\label{eq:sum_X_bound_proof}
    \forall t\in[T],\omega\in\Omega:\quad -\sum_{s=1}^{t-1} X_s^\omega &\leq - \frac{1}{2} \sum_{s=1}^{t-1}\E_{i\sim\widetilde{\mu}_s,j\sim\nu_s}\left[\left(A_\omega(i,j)-A(i,j)\right)^2\right]\notag\\
    &\quad+ 4C B_l\left(B_l+\sigma\sqrt{2\log(8T/\delta)}\right)  \left(\log\left(\frac{4T}{\delta}\right)+d\log\left(1+2T\sqrt{d}\right)\right).  
\end{align}

Similar to \eqref{eq:sum_X_bound}, conditioned on event $\mathcal{E}$, we could upper bound $-\sum_{s=1}^{t-1} Y_s^\omega$ as follows with probability at least $1-\frac{\delta}{4}$:
\begin{align}\label{eq:sum_Y_bound_proof}
    \forall t\in[T],\omega\in\Omega:\quad -\sum_{s=1}^{t-1} Y_s^\omega &\leq - \frac{1}{2} \sum_{s=1}^{t-1}\E_{i\sim\mu_s,j\sim\widetilde{\nu}_s}\left[\left(A_\omega(i,j)-A(i,j)\right)^2\right]\notag\\
    &\quad+ 4C B_l\left(B_l+\sigma\sqrt{2\log(8T/\delta)}\right)  \left(\log\left(\frac{4T}{\delta}\right)+d\log\left(1+2T\sqrt{d}\right)\right) .  
\end{align}

Applying union bound completes the proof of Lemma~\ref{lm:regret_on_squared_loss}.

\subsubsection{Proof of Lemma~\ref{lm:diff_f}}

For any $\mu,\nu\in\Delta^m\times\Delta^n$ and any $\omega\in\Omega$, notice that
\begin{align}\label{eq:innerprod}
    f^{\mu,\nu}(A_\omega) - f^{\mu,\nu}(A) = \langle\underbrace{\E_{i\sim\mu,j\sim\nu}[\phi(i,j)]}_{=: x(\mu,\nu)},\omega-\omega^\star\rangle,
\end{align}
where we denote $\E_{i\sim\mu,j\sim\nu}[\phi(i,j)]$ as $x(\mu,\nu)$ for simplicity. By Assumption~\ref{asmp:bounded_payoff}, it guarantees that $\norm{x(\mu,\nu)}_\infty\leq 1$ for all $\mu,\nu$.
For each $t\in[T]$, we define $\Lambda_t\in\R^{d\times d}$ as
\begin{align}\label{eq:Lambda}
    \Lambda_t\coloneqq \lambda I_d + \sum_{s=1}^{t-1} x(\widehat{\mu}_s,\widehat{\nu}_s)x(\widehat{\mu}_s,\widehat{\nu}_s)^\top
\end{align}
for any $\lambda>0$. By Lemma~\ref{lm:potential} and Lemma~\ref{lm:information_gain}, we have
\begin{align}\label{eq:|x|_bound}
    \sum_{s=1}^t \min\left\{ \norm{x(\widehat{\mu}_s,\widehat{\nu}_s)}_{\Lambda_{s}^{-1}},1\right\}\leq 2d\log\left(1+\frac{T}{d\lambda}\right)\coloneqq d(\lambda) ,
\end{align}
which will be used repeatedly in the proof.

We decompose $\sum_{t=1}^T \left| f^{\widehat{\mu}_t,\widehat{\nu}_t}(A_{\widehat{\omega}_t}) - f^{\widehat{\mu}_t,\widehat{\nu}_t}(A)\right|$ into two terms:
\begin{align}\label{eq:dec}
    \sum_{t=1}^T \left| f^{\widehat{\mu}_t,\widehat{\nu}_t}(A_{\widehat{\omega}_t}) - f^{\widehat{\mu}_t,\widehat{\nu}_t}(A)\right|&= \underbrace{\sum_{t=1}^T \left| \langle x(\widehat{\mu}_t,\widehat{\nu}_t),\widehat{\omega}_t-\omega^\star\rangle\right|\mathbbm{1}\left\{\norm{x(\widehat{\mu}_t,\widehat{\nu}_t)}_{\Lambda_{t}^{-1}}\leq 1\right\}}_{(a)}\notag\\
    & \quad + \underbrace{\sum_{t=1}^T \left| \langle x(\widehat{\mu}_t,\widehat{\nu}_t),\widehat{\omega}_t-\omega^\star\rangle\right|\mathbbm{1}\left\{\norm{x(\widehat{\mu}_t,\widehat{\nu}_t)}_{\Lambda_{t}^{-1}}> 1\right\}}_{(b)}.
\end{align}
To prove Lemma~\ref{lm:diff_f}, below we bound (a) and (b) separately.

\paragraph{Step 1: bounding term (a).} To bound term (a), it follows that
\begin{align}\label{eq:bound_i_1}
    (a) &= \sum_{t=1}^T \left| \langle x(\widehat{\mu}_t,\widehat{\nu}_t),\widehat{\omega}_t-\omega^\star\rangle\right|\mathbbm{1}\left\{\norm{x(\widehat{\mu}_t,\widehat{\nu}_t)}_{\Lambda_{t}^{-1}}\leq 1\right\}\notag\\
    &\leq \sum_{t=1}^T \norm{\widehat{\omega}_t-\omega^\star}_{\Lambda_t}\norm{x(\widehat{\mu}_t,\widehat{\nu}_t)}_{\Lambda_t^{-1}}\mathbbm{1}\left\{\norm{x(\widehat{\mu}_t,\widehat{\nu}_t)}_{\Lambda_{t}^{-1}}\leq 1\right\}\notag\\
    &\leq \sum_{t=1}^T \norm{\widehat{\omega}_t-\omega^\star}_{\Lambda_t}\min\left\{ \norm{x(\widehat{\mu}_t,\widehat{\nu}_t)}_{\Lambda_{t}^{-1}},1\right\},
\end{align}
where the first inequality uses the Cauchy-Schwarz inequality, and the second inequality uses the fact that
$$\norm{x(\widehat{\mu}_t,\widehat{\nu}_t)}_{\Lambda_t^{-1}}\mathbbm{1}\left\{\norm{x(\widehat{\mu}_t,\widehat{\nu}_t)}_{\Lambda_{t}^{-1}}\leq 1\right\}\leq \min\left\{ \norm{x(\widehat{\mu}_t,\widehat{\nu}_t)}_{\Lambda_{t}^{-1}},1\right\}.$$

Also, by Assumption~\ref{asmp:bounded_payoff} and the definition of $\Lambda_t$ in \eqref{eq:Lambda}, we have
\begin{align}\label{eq:|w|_bound}
    \norm{\widehat{\omega}_t-\omega^\star}_{\Lambda_t}\leq 2\sqrt{\lambda d}+\left(\sum_{s=1}^{t-1} |\langle \widehat{\omega}_t-\omega^\star,x(\widehat{\mu}_s,\widehat{\nu}_s)\rangle|^2\right)^{1/2},
\end{align}
which gives
\begin{align}\label{eq:bound_i_2}
    & \sum_{t=1}^T \norm{\widehat{\omega}_t-\omega^\star}_{\Lambda_t}\min\left\{ \norm{x(\widehat{\mu}_t,\widehat{\nu}_t)}_{\Lambda_{t}^{-1}},1\right\}\notag\\
    &\leq \sum_{t=1}^T\left(2\sqrt{\lambda d} +\left(\sum_{s=1}^{t-1} |\langle \widehat{\omega}_t-\omega^\star,x(\widehat{\mu}_s,\widehat{\nu}_s)\rangle|^2\right)^{1/2}\right)\cdot\min\left\{ \norm{x(\widehat{\mu}_t,\widehat{\nu}_t)}_{\Lambda_{t}^{-1}},1\right\}\notag\\
    &\leq \left(\sum_{t=1}^T 4\lambda d\right)^{1/2}
    \left(\sum_{t=1}^T \min\left\{ \norm{x(\widehat{\mu}_t,\widehat{\nu}_t)}_{\Lambda_{t}^{-1}},1\right\}\right)^{1/2}\notag\\
    &\quad + \left(\sum_{t=1}^T \sum_{s=1}^{t-1} |\langle \widehat{\omega}_t-\omega^\star,x(\widehat{\mu}_s,\widehat{\nu}_s)\rangle|^2\right)^{1/2}\left(\sum_{t=1}^T \min\left\{ \norm{x(\widehat{\mu}_t,\widehat{\nu}_t)}_{\Lambda_{t}^{-1}},1\right\}\right)^{1/2}\notag\\
    &\leq 2\sqrt{\lambda dT \min\{d(\lambda),T\}} + \left(d(\lambda)\sum_{t=1}^T \sum_{s=1}^{t-1} |\langle \widehat{\omega}_t-\omega^\star,x(\widehat{\mu}_s,\widehat{\nu}_s)\rangle|^2\right)^{1/2},
\end{align}
where the first inequality uses \eqref{eq:|w|_bound} and the second inequality uses the Cauchy-Schwarz inequality and the fact that 
$$\min\left\{ \norm{x(\widehat{\mu}_t,\widehat{\nu}_t)}_{\Lambda_{t}^{-1}},1\right\}^2\leq \min\left\{ \norm{x(\widehat{\mu}_t,\widehat{\nu}_t)}_{\Lambda_{t}^{-1}},1\right\},$$
and the last inequality uses \eqref{eq:|x|_bound}.

Plugging \eqref{eq:bound_i_2} into \eqref{eq:bound_i_1}, we have
\begin{align}\label{eq:bound_i_lm}
    (a)\leq 2\sqrt{d}\cdot\sqrt{\lambda T \min\{d(\lambda),T\}} + \left(d(\lambda)\sum_{t=1}^T \sum_{s=1}^{t-1} |\langle \widehat{\omega}_t-\omega^\star,x(\widehat{\mu}_s,\widehat{\nu}_s)\rangle|^2\right)^{1/2}.
\end{align}

\paragraph{Step 2: bounding term (b).} It follows that
\begin{align}\label{eq:bound_ii_lm}
    (b) &= \sum_{t=1}^T \left| \langle x(\widehat{\mu}_t,\widehat{\nu}_t),\widehat{\omega}_t-\omega^\star\rangle\right|\mathbbm{1}\left\{\norm{x(\widehat{\mu}_t,\widehat{\nu}_t)}_{\Lambda_{t}^{-1}}> 1\right\} \notag\\
    &\leq \sum_{t=1}^T \left| \langle x(\widehat{\mu}_t,\widehat{\nu}_t),\widehat{\omega}_t-\omega^\star\rangle\right|\min\left\{ \norm{x(\widehat{\mu}_t,\widehat{\nu}_t)}_{\Lambda_{t}^{-1}},1\right\}  \leq 2B_l \min\{T, d(\lambda)\},
\end{align}
where the first inequality uses the fact that
$$\mathbbm{1}\left\{\norm{x(\widehat{\mu}_t,\widehat{\nu}_t)}_{\Lambda_{t}^{-1}}> 1\right\}\leq \min\left\{ \norm{x(\widehat{\mu}_t,\widehat{\nu}_t)}_{\Lambda_{t}^{-1}},1\right\},$$
and the last inequality uses Assumption~\ref{asmp:bounded_payoff} and \eqref{eq:|x|_bound}.

\paragraph{Step 3: combining (a) and (b).} Plugging \eqref{eq:bound_i_lm} and \eqref{eq:bound_ii_lm} into \eqref{eq:dec}, we have
\begin{align}
    & \sum_{t=1}^T \left| f^{\widehat{\mu}_t,\widehat{\nu}_t}(A_{\widehat{\omega}_t}) - f^{\widehat{\mu}_t,\widehat{\nu}_t}(A)\right|\notag\\
    &\leq 2\sqrt{d}\cdot\sqrt{\lambda T \min\{d(\lambda),T\}} + \left(d(\lambda)\sum_{t=1}^T \sum_{s=1}^{t-1} |\langle \widehat{\omega}_t-\omega^\star,x(\widehat{\mu}_s,\widehat{\nu}_s)\rangle|^2\right)^{1/2}+2B_l \min\{T, d(\lambda)\}\notag\\
    &\leq \left(\frac{d(\lambda)}{\eta}\cdot\eta\sum_{t=1}^T \sum_{s=1}^{t-1}  |\langle \widehat{\omega}_t-\omega^\star,x(\widehat{\mu}_s,\widehat{\nu}_s)\rangle|^2\right)^{1/2} + (\sqrt{d}+2B_l) \min\{d(\lambda),T\} +\sqrt{d} \lambda T\notag\\
    &\leq \frac{d(\lambda)}{2\eta} +\frac{\eta}{2}\sum_{t=1}^T \sum_{s=1}^{t-1} |\langle \widehat{\omega}_t-\omega^\star,x(\widehat{\mu}_s,\widehat{\nu}_s)\rangle|^2 + (\sqrt{d}+2B_l) \min\{d(\lambda),T\} +\sqrt{d} \lambda T 
\end{align}
for any $\eta>0$, where the second and third inequalities both use the fact that $\sqrt{ab}\leq \frac{a+b}{2}$ for any $a,b\geq 0$. The proof is completed by plugging in the following fact into the above relation: for any $\mu,\nu\in\Delta^m\times\Delta^n$ and any $\omega\in\Omega$, we have
\begin{align}\label{eq:square_bound}
    |\langle x(\mu,\nu),\omega-\omega^\star\rangle|^2= |\E_{i\sim\mu,j\sim\nu} [A_\omega(i,j)-A(i,j)]|^2\leq \E_{i\sim\mu,j\sim\nu} [(A_\omega(i,j)-A(i,j))^2] .
\end{align}

\subsection{Proof of Theorem~\ref{thm:regret_MG}}\label{app:proof_thm_regret_MP}

For notation simplicity, we define
    \begin{align}\label{eq:pi_t_n}
        \widetilde{\vpi}_{t,n}\coloneqq (\widetilde{\pi}_t^n, \vpi_t^{-n}),\quad \forall n\in[N].
    \end{align}

Analogous to \eqref{eq:regret_dec1}, here we decompose the regret as
\begin{align}\label{eq:regret_dec1_mg}
    \regret(T) &= \sum_{t=1}^T\frac{1}{N}\sum_{n=1}^N \left(V_{n}^{\star,\vpi_t^{-n}}(\rho) - V_{n}^{\vpi_t}(\rho)\right),\notag\\
    &= \sum_{t=1}^T \frac{1}{N}\sum_{n=1}^N \left(V_{n}^{\star,\vpi_t^{-n}}(\rho) - V_{f_t,n}^{\star,\vpi_t^{-n}}(\rho)\right) + \sum_{t=1}^T \frac{1}{N}\sum_{n=1}^N \left(V_{f_t,n}^{\star,\vpi_t^{-n}}(\rho) - V_{n}^{\widetilde{\pi}_t^n,\vpi_t^{-n}}(\rho)\right)\notag\\
    &\quad + \sum_{t=1}^T \frac{1}{N}\sum_{n=1}^N \left(V_{n}^{\widetilde{\pi}_t^n,\vpi_t^{-n}}(\rho) - V_{f_{t-1},n}^{\widetilde{\pi}_t^n,\vpi_t^{-n}}(\rho)\right)
    + \sum_{t=1}^T \frac{1}{N}\sum_{n=1}^N \left(V_{f_{t-1},n}^{\widetilde{\pi}_t^n,\vpi_t^{-n}}(\rho) - V_{f_{t-1},n}^{\vpi_t}(\rho)\right)\notag\\
    &\quad + \sum_{t=1}^T \frac{1}{N}\sum_{n=1}^N \left(V_{f_{t-1},n}^{\vpi_t}(\rho) - V_{n}^{\vpi_t}(\rho)\right).
\end{align}

By line 4 in Algorithm~\ref{alg:markov_game} we know that the second term in the third line of \eqref{eq:regret_dec1_mg} is non-positive. 
Besides, \eqref{eq:policy_update_2_mg} indicates
\begin{align}\label{eq:dagger_equation_mg}
    \forall n\in[N]:\quad V_{f_t,n}^{\star,\vpi_t^{-n}}(\rho) =V_{f_t,n}^{\widetilde{\pi}_t^n,\vpi_t^{-n}}(\rho).
\end{align}
Combining these two facts, we have 
\begin{align}\label{eq:regret_dec2_mg}
    \regret(T) &\leq \underbrace{\sum_{t=1}^T \frac{1}{N}\sum_{n=1}^N \left(V_{n}^{\star,\vpi_t^{-n}}(\rho) - V_{f_t,n}^{\star,\vpi_t^{-n}}(\rho)\right)}_{\text{(i)}} 
    + \underbrace{\sum_{t=1}^T \frac{1}{N}\sum_{n=1}^N \left(V_{f_t,n}^{\widetilde{\pi}_t^n,\vpi_t^{-n}}(\rho) - V_{n}^{\widetilde{\pi}_t^n,\vpi_t^{-n}}(\rho)\right)}_{\text{(ii)}}\notag\\
    &\quad + \underbrace{\sum_{t=1}^T \frac{1}{N}\sum_{n=1}^N \left(V_{n}^{\widetilde{\pi}_t^n,\vpi_t^{-n}}(\rho) - V_{f_{t-1},n}^{\widetilde{\pi}_t^n,\vpi_t^{-n}}(\rho)\right)}_{\text{(iii)}}
    + \underbrace{\sum_{t=1}^T \frac{1}{N}\sum_{n=1}^N \left(V_{f_{t-1},n}^{\vpi_t}(\rho) - V_{n}^{\vpi_t}(\rho)\right)}_{\text{(iv)}}.
\end{align}

In the following we upper bound each term in \eqref{eq:regret_dec2_mg} separately.

\paragraph{Step 1: bounding term (i).} 
By Assumption~\ref{asmp:realizable} we know that there exists $f^\star\in\gF$ such that $f^\star\coloneqq \PP = \PP_{f^\star}$.
By the model update rule~\eqref{eq:model_update_mg} in Algorithm~\ref{alg:markov_game} and the definition of the loss function~\eqref{eq:loss_mg}, we have
\begin{align*}
    \gL_t(f_t)-\alpha\sum_{n=1}^N V_{f_t,n}^{\star,\vpi_t^{-n}}(\rho) \leq \gL_t(f^\star)-\alpha\sum_{n=1}^N V_{n}^{\star,\vpi_t^{-n}}(\rho)
\end{align*}
from which we deduce
\begin{align}\label{eq:loss_diff_mg}
   \mathrm{(i)} \leq \frac{1}{N\alpha}\sum_{t=1}^T\left(\gL_t(f^\star)-\gL_t(f_t)\right).
\end{align}
It then boils down to bounding the right-hand side of \eqref{eq:loss_diff_mg}. 

We first define random variables $X_{t,h}^f$ and $Y_{t,h,n}^f$ as
\begin{align}\label{eq:X_t_mg}
    X_{t,h}^f\coloneqq \log \left(\frac{\PP_h(s_{t,h+1}|s_{t,h},\va_{t,h})}{\PP_{f,h}(s_{t,h+1}|s_{t,h},\va_{t,h})}\right)\quad\text{and}\quad Y_{t,h,n}^f\coloneqq \log \left(\frac{\PP_h(s_{t,h+1}^n|s_{t,h}^n,\va_{t,h}^n)}{\PP_{f,h}(s_{t,h+1}^n|s_{t,h}^n,\va_{t,h}^n)}\right),\quad \forall n\in[N].
\end{align}
By the definition of the loss function~\eqref{eq:loss_mg}, we have
\begin{align}\label{eq:loss_diff_decompose_mg}
    \gL_t(f^\star)-\gL_t(f) = -\sum_{i=1}^{t-1}\sum_{h=1}^H\sum_{n=1}^N \left(X_{i,h}^f+Y_{i,h,n}^f\right).
\end{align}

Let $\hellinger{\cdot}{\cdot}$ denote the Hellinger divergence defined as:
\begin{align}\label{eq:Hellinger}
    \hellinger{P}{Q}\coloneqq \frac{1}{2}\int_\gX\left(\sqrt{P(x)}-\sqrt{Q(x)}\right)^2 dx 
\end{align}
for any probability measures $P$ and $Q$ on $\gX$, and define
\begin{align}\label{eq:Hellinger_loss}
    \ell(f_h,s,\va)\coloneqq \hellinger{\PP_{f,h}(\cdot|s,\va)}{\PP_h(\cdot|s,\va)}.
\end{align}
In the following lemma we provide a concentration result for the random variables $X_{t,h}^f$ and $Y_{t,h,n}^f$ in \eqref{eq:loss_diff_decompose_mg} (recall we define $\widetilde{\vpi}_{t,n}\coloneqq (\widetilde{\pi}_t^n, \vpi_t^{-n})$ in \eqref{eq:pi_t_n}), where  the state-action visitation distribution $d_h^{\vpi}(\rho)\in\Delta(\gS\times\gA)$ at step $h$ under the policy $\vpi$ and the initial state distribution $\rho$ is defined as
\begin{align}\label{eq:state_action_visitation_finite}
    d_{h}^{\vpi}(s,a ; \rho)\coloneqq \E_{s\sim\rho}\PP^{\vpi}(s_h=s,\va_h=\va|s_1=s).
\end{align}

\begin{lm}\label{lm:bound_X_Y_mg}
    When Assumptions~\ref{asmp:function_class} and \ref{asmp:realizable} hold, for any $\delta \in (0,1)$, with probability at least $1-\delta$, we have  for all $t\in[T]$, $ f\in\gF$ and $n\in[N]$:
    \begin{align}\label{eq:sum_X_mg}
        -\sum_{i=1}^{t-1}\sum_{h=1}^H X_{i,h}^f& \leq -2\sum_{i=1}^{t-1}\sum_{h=1}^H \E_{(s_{i,h},\va_{i,h})\sim d_h^{\vpi_i}(\rho)}\left[\ell(f_h,s_{i,h},\va_{i,h})\right] \notag\\
        & \quad + 2\sqrt{2}H + 2H\log\left(\frac{(N+1)H}{\delta}\right) + 2dH\log\left(1+2\sqrt{d}|\gS|^2 T^2\right) .\\ 
     -\sum_{i=1}^{t-1}\sum_{h=1}^H Y_{i,h,n}^f  & \leq -2\sum_{i=1}^{t-1}\sum_{h=1}^H \E_{(s_{i,h}^n,\va_{i,h}^n)\sim d_h^{\widetilde{\vpi}_{i,n}}(\rho)}\left[\ell(f_h,s_{i,h}^n,\va_{i,h}^n)\right] \nonumber \\
        & \quad + 2\sqrt{2}H + 2H\log\left(\frac{(N+1)H}{\delta}\right) + 2dH\log\left(1+2\sqrt{d}|\gS|^2 T^2\right). \label{eq:sum_Y_mg}
    \end{align}
\end{lm}

Combining \eqref{eq:loss_diff_mg}, \eqref{eq:loss_diff_decompose_mg}, \eqref{eq:sum_X_mg}, \eqref{eq:sum_Y_mg}, we have with probability at least $1-\delta$:
\begin{align}\label{eq:a_bound}
    \text{(i)} & \leq -\frac{2}{N\alpha}\sum_{n=1}^N\bigg\{\sum_{t=1}^T\sum_{i=1}^{t-1}\sum_{h=1}^H 
    \E_{(s_{i,h},\va_{i,h})\sim d_h^{\vpi_i}(\rho)}\left[\ell(f_{t,h},s_{i,h},\va_{i,h})\right]\notag\\
  &  \quad +\sum_{t=1}^T\sum_{i=1}^{t-1}\sum_{h=1}^H 
    \E_{(s_{i,h}^n,\va_{i,h}^n)\sim d_h^{\widetilde{\vpi}_{i,n}}(\rho)}\left[\ell(f_{t,h},s_{i,h}^n,\va_{i,h}^n)\right]\bigg\}\notag\\
   & \quad + \frac{4HT}{\alpha}\left(\sqrt{2} + \log\left(\frac{(N+1)H}{\delta}\right) + d\log\left(1+2\sqrt{d}|\gS|^2T^2\right)\right). 
\end{align}

\paragraph{Step 2: bounding terms (ii), (iii) and (iv).} To bound (ii), (iii) and (iv), we introduce the following lemma.  
\begin{lm}\label{lm:V_diff_bound}
    Under Assumptions~\ref{asmp:function_class} and \ref{asmp:realizable}, for any $n\in[N]$, $\beta\geq 0$, $\{\widehat{\vpi}_{t}: \gS\times[H]\rightarrow\Delta(\gA)\}_{t\in[T]}$ and $\{\widehat{f}_{t}\}_{t\in[T]}\subset \gF$,  we have
    \begin{align}\label{eq:V_diff_bound}
       \sum_{t=1}^T \left|V_{\widehat{f}_{t},n}^{\widehat{\vpi}_{t}}(\rho)-V_{n}^{\widehat{\vpi}_{t}}(\rho)\right|&\leq \frac{\eta}{2}\sum_{t=1}^T \sum_{i=1}^{t-1}\sum_{h=1}^H \E_{(s,\va)\sim d_h^{\widehat{\vpi}_i}(\rho)}\ell(\widehat{f}_{t,h},s,\va) \notag\\
       &\quad+H\left(\frac{4d_H(\lambda)H}{\eta}
       + \left(\sqrt{d}+H\right) \min\{d_H(\lambda),T\} +\sqrt{d} \lambda T\right)
    \end{align}
    for any $\eta>0$ and $\lambda>0$, where $d_H(\lambda)$ is defined as
    $$d_H(\lambda)\coloneqq 2d\log\left(1+\frac{H^2T}{\lambda}\right).$$
\end{lm}

Now we are ready to bound (ii), (iii) and (iv). To bound (ii), letting $\widehat{f}_{t}=f_t$ and
$\widehat{\vpi}_{t}=\widetilde{\vpi}_{t,n}$ for each $n \in [N]$ in Lemma~\ref{lm:V_diff_bound} (recall we define $\widetilde{\vpi}_{t,n}\coloneqq (\widetilde{\pi}_t^n,\vpi_t^{-n})$ in \eqref{eq:pi_t_n}), we have for any $\eta>0$:
\begin{align}\label{eq:b_bound} 
    \text{(ii)}&\leq \frac{\eta}{2N}\sum_{h=1}^H\sum_{n=1}^N\sum_{t=1}^T \sum_{i=1}^{t-1} \E_{(s,\va)\sim d_h^{\widetilde{\vpi}_{i,n}}(\rho)}\ell(f_{t,h},s,\va) \notag\\
    &\quad+H\left(\frac{4d_H(\lambda)H}{\eta}
    + \left(\sqrt{d}+H\right) \min\{d_H(\lambda),T\} +\sqrt{d} \lambda T\right).
\end{align}

Letting $\widehat{f}_{t,h}=f_{t-1}$ and
$\widehat{\vpi}_{t,h}=\widetilde{\vpi}_{t,n}$ for each $n \in [N]$ in Lemma~\ref{lm:V_diff_bound}, we can bound (iii) as follows:
\begin{align}\label{eq:c_bound_temp} 
    \text{(iii)}&\leq \frac{\eta}{2N}\sum_{h=1}^H\sum_{n=1}^N\sum_{t=1}^T \sum_{i=1}^{t-1} \E_{(s,\va)\sim d_h^{\widetilde{\vpi}_{i,n}}(\rho)}\ell(f_{t-1,h},s,\va) \notag\\
    &\quad+H\left(\frac{4d_H(\lambda)H}{\eta}
    + \left(\sqrt{d}+H\right) \min\{d_H(\lambda),T\} +\sqrt{d} \lambda T\right).
\end{align}
To continue to bound the first term, note that
\begin{align}
\sum_{h=1}^H \sum_{t=1}^T \sum_{i=1}^{t-1} \E_{(s,\va)\sim d_h^{\widetilde{\vpi}_{i,n}}(\rho)}\ell(f_{t-1,h},s,\va)   & \leq  \sum_{h=1}^H \sum_{t=1}^T \sum_{i=1}^{t-2} \E_{(s,\va)\sim d_h^{\widetilde{\vpi}_{i,n}}(\rho)}\ell(f_{t-1,h},s,\va) + HT  \nonumber \\
    &= \sum_{h=1}^H \sum_{t=0}^{T-1} \sum_{i=1}^{t-1} \E_{(s,\va)\sim d_h^{\widetilde{\vpi}_{i,n}} 
    (\rho)}\ell(f_{t,h},s,\va) +HT \notag\\
    &\leq  \sum_{h=1}^H \sum_{t=1}^{T} \sum_{i=1}^{t-1} \E_{(s,\va)\sim d_h^{\widetilde{\vpi}_{i,n}}(\rho)}\ell(f_{t,h},s,\va) + HT, \nonumber
\end{align}
where the first inequality uses the fact that
\begin{align}
    \ell(f_h,s,\va)=\hellinger{\PP_{f,h}(\cdot|s,\va)}{\PP_h(\cdot|s,\va)}\leq 1,
\end{align}
the second line shifts the index of $t$ by 1, and the last line follows by noticing the first summand is $0$ at $t=0$. Plugging the above relation back to \eqref{eq:c_bound_temp} leads to
\begin{align}\label{eq:c_bound} 
    \text{(iii)}&\leq \frac{\eta}{2N}\sum_{h=1}^H \sum_{n=1}^N \sum_{t=1}^{T} \sum_{i=1}^{t-1} \E_{(s,\va)\sim d_h^{\widetilde{\vpi}_{i,n}}(\rho)}\ell(f_{t,h},s,\va)  \notag\\
    &\quad+H\left(\frac{4d_H(\lambda)H}{\eta}
    + \left(\sqrt{d}+H\right) \min\{d_H(\lambda),T\} +\sqrt{d} \lambda T + \frac{\eta}{2} T \right).
\end{align}

Finally, similar to \eqref{eq:c_bound}, letting $\widehat{f}_{t,h}=f_{t-1}$, $\widehat{\vpi}_{t,h}=\vpi_t$ for each $n \in [N]$ and replace $\eta$ by $2\eta$ in Lemma~\ref{lm:V_diff_bound}, we can bound (iv) as follows:
\begin{align}\label{eq:d_bound} 
    \text{(iv)} &\leq \frac{\eta}{N}\sum_{h=1}^H\sum_{n=1}^N\sum_{t=1}^{T} \sum_{i=1}^{t-1} \E_{(s,\va)\sim d_h^{\vpi_i}(\rho)}\ell(f_{t,h},s,\va) \notag\\
    &\quad+H\left(\frac{2Hd_H(\lambda)}{\eta}
    + \left(\sqrt{d}+H\right) \min\{d_H(\lambda),T\} +\sqrt{d} \lambda T +\eta T\right).
\end{align}

\paragraph{Step 3: combining the bounds.} Letting $\eta = \frac{2}{\alpha}$ in \eqref{eq:b_bound}, \eqref{eq:c_bound} and \eqref{eq:d_bound}, and adding \eqref{eq:a_bound}, \eqref{eq:b_bound}, \eqref{eq:c_bound} and \eqref{eq:d_bound} together, we have with probability at least $1-\delta$:
\begin{align*}
    \regret(T)&\leq \frac{4HT}{\alpha}\left(\sqrt{2} + \log\left(\frac{(N+1)H}{\delta}\right) + d\log\left(1+2\sqrt{d}|\gS|^2T^2\right)\right)\notag\\
    &\quad + H\left(5\alpha d_H(\lambda)H
    + 3\left(\sqrt{d}+H\right) \min\{d_H(\lambda),T\} +3\sqrt{d} \lambda T +\frac{3}{\alpha}T\right).
\end{align*}

By setting
\begin{align}
    \lambda =\sqrt{\frac{d}{T}},\quad \alpha = \sqrt{\frac{\log\left(\frac{(N+1)H}{\delta}\right) + d\log\left(1+2\sqrt{d}|\gS|^2T^2\right)}{Hd\log\left(1+\frac{H^2T^{3/2}}{\sqrt{d}}\right)}T},
\end{align}
in the above expression, we have with probability at least $1-\delta$:
\begin{align}
    \regret(T)&\leq 4(1+\sqrt{2})\sqrt{\frac{d\log\left(1+\frac{H^2T^{3/2}}{\sqrt{d}}\right)}{\log\left(\frac{(N+1)H}{\delta}\right) + d\log\left(1+2\sqrt{d}|\gS|^2T^2\right)}}\cdot \sqrt{HT}\notag\\
    &\quad 14d\sqrt{H^3T}\cdot\sqrt{\left(\frac{1}{d}\log\left(\frac{(N+1)H}{\delta}\right) + \log\left(1+\sqrt{d}|\gS|^2T^2\right)\right)\log\left(1+\frac{H^2T^{3/2}}{\sqrt{d}}\right)}\notag\\
    &\quad + 6H\left(\sqrt{d}+H\right) d\log\left(1+\frac{H^2T^{3/2}}{\sqrt{d}}\right) +3dH\sqrt{T},
\end{align}
which gives the desired result after simplifying the expression.


\subsubsection{Proof of Lemma~\ref{lm:bound_X_Y_mg}}\label{app:proof_bound_X_Y_mg}

Same as in \eqref{eq:covering}, for the parameter spaces $\Theta_h$, $h\in[H]$, by Assumption~\ref{asmp:function_class} and Lemma~\ref{lm:covering} we have
\begin{align}\label{eq:covering_mg}
    \forall h\in[H]:\quad\log \gN(\Theta_h,\epsilon,\norm{\cdot}_2)\leq d\log\left(1+\frac{2\sqrt{d}}{\epsilon}\right)
\end{align}
for any $\epsilon>0$. Thus for any $\epsilon>0$, there exists an $\epsilon$-net $\Theta_{h,\epsilon}$ of $\Theta_h$ ($\Theta_{h,\epsilon}\subset\Theta_h$) such that $\log|\Theta_{h,\epsilon}|\leq d\log\left(1+\frac{2\sqrt{d}}{\epsilon}\right)$, $\forall h\in[H]$.
Define
\begin{align*}
    \gF_{h,\epsilon} \coloneqq \left\{f_h\in\gF_h: f_h(s,\va,s_{h+1})=\phi_h(s,\va,s_{h+1})^\top\theta_h,\theta_h\in\Theta_{h,\epsilon}\right\}.
\end{align*}

For any $f\in\gF$, there exists $\theta_h\in\Theta_h$ such that $f_h(s,\va,s_{h+1})=\phi_h(s,\va,s_{h+1})^\top\theta_h$. In addition, there exists $\theta_{h,\epsilon}\in\Theta_{h,\epsilon}$ such that $\norm{\theta_h-\theta_{h,\epsilon}}_2\leq \epsilon$. We let $f_\epsilon(s,\va,s_{h+1})=\phi_h(s,\va,s_{h+1})^\top\theta_{h,\epsilon}$. Then $f_\epsilon\in\gF_{h,\epsilon}$, and we have
\begin{align}\label{eq:diff_eps}
    |\PP_{f,h}(s_{h+1}|s,\va)-\PP_{f_\epsilon,h}(s_{h+1}|s,\va)|=|\phi_h(s,\va,s_{h+1})^\top(\theta_h-\theta_{h,\epsilon})|\leq \epsilon,
\end{align}
from which we deduce
\begin{align}\label{eq:diff_net_approx}
    \forall t\in[T],h\in[H]:\quad -X_{t,h}^f\leq -\log\left(\frac{\PP_h(s_{t,h+1}|s_{t,h},\va_{t,h})}{\PP_{f_\epsilon,h}(s_{t,h+1}|s_{t,h},\va_{t,h})+\epsilon}\right)\coloneqq -X_{t,h}^{f_\epsilon}(\epsilon).
\end{align}

Let $\gF_t\coloneqq \sigma(\gD_t)$ be the $\sigma$-algebra generated by the dataset $\gD_t$. 
By Lemma~\ref{lm:martingale_exp} we have 
with probability at least $1-\frac{\delta}{N+1}$:
\begin{align}\label{eq:exp_concentration_net}
   \forall t\in[T],h\in[H],f_{h,\epsilon}\in\gF_{h,\epsilon}: \quad -\frac{1}{2}\sum_{i=1}^{t-1} X^{f_\epsilon}_{i,h}(\epsilon) &\leq \sum_{i=1}^{t-1} \log\E\left[\exp\left(-\frac{1}{2}X^{f_\epsilon}_{i,h}(\epsilon)\right)\bigg|\gF_{i-1}\right] \notag\\
   &\quad+ \log\left(\frac{(N+1)H}{\delta}\right) + d\log\left(1+\frac{2\sqrt{d}}{\epsilon}\right).
\end{align}

Then we have for all $t\in[T]$, $h\in[H]$ and $f\in\gF$:
\begin{align}\label{eq:concentration_exp_mg_1}
    -\frac{1}{2}\sum_{i=1}^{t-1}\sum_{h=1}^H X_{i,h}^f& \leq  -\frac{1}{2}\sum_{i=1}^{t-1}\sum_{h=1}^H X^{f_\epsilon}_{i,h}(\epsilon)\notag\\
    & \leq  \sum_{i=1}^t \log\E\left[\exp\left(-\frac{1}{2}X^{f_\epsilon}_{i,h}(\epsilon)\right)\bigg|\gF_{i-1}\right] + H\log\left(\frac{(N+1)H}{\delta}\right) + dH\log\left(1+\frac{2\sqrt{d}}{\epsilon}\right),
\end{align}
where the first line follows \eqref{eq:diff_net_approx}, and the second line follows from \eqref{eq:exp_concentration_net}. The first term in the last line of \eqref{eq:concentration_exp_mg_1} can be further bounded as follows:
\begin{align}\label{eq:concentration_exp_mg_part}
    &\sum_{i=1}^t \log\E\left[\exp\left(-\frac{1}{2}X^{f_\epsilon}_{i,h}(\epsilon)\right)\bigg|\gF_{i-1}\right]\notag\\
    &=\sum_{i=1}^{t-1}\sum_{h=1}^H \log\E\left[\sqrt{\frac{\PP_{f_\epsilon,h}(s_{i,h+1}|s_{i,h},\va_{i,h})+\epsilon}{\PP_h(s_{i,h+1}|s_{i,h},\va_{i,h})}}\bigg|\gF_{s-1}\right]\notag\\
    &=\sum_{i=1}^{t-1}\sum_{h=1}^H \log\E_{(s_{i,h},\va_{i,h})\sim d_h^{\vpi_i}(\rho),\atop s_{i,h+1}\sim\PP_h(\cdot|s_{i,h},\va_{i,h})}\left[\sqrt{\frac{\PP_{f_\epsilon,h}(s_{i,h+1}|s_{i,h},\va_{i,h})+\epsilon}{\PP_h(s_{i,h+1}|s_{i,h},\va_{i,h})}}\right]\notag\\
    &=\sum_{i=1}^{t-1}\sum_{h=1}^H \log\E_{(s_{i,h},\va_{i,h})\sim d_h^{\vpi_i}(\rho)}\left[\int_\gS\sqrt{\left(\PP_{f_\epsilon,h}(s_{i,h+1}|s_{i,h},\va_{i,h})+\epsilon\right)\PP_h(s_{i,h+1}|s_{i,h},\va_{i,h})} ds_{i,h+1}\right]\notag\\
    & \leq  \sum_{i=1}^{t-1}\sum_{h=1}^H \log\E_{(s_{i,h},\va_{i,h})\sim d_h^{\vpi_i}(\rho)}\left[\int_\gS\sqrt{\left(\PP_{f,h}(s_{i,h+1}|s_{i,h},\va_{i,h})+2\epsilon\right)\PP_h(s_{i,h+1}|s_{i,h},\va_{i,h})} ds_{i,h+1}\right],
\end{align}
where the last inequality uses \eqref{eq:diff_eps}. Furthermore, we have
\begin{align}\label{eq:construct_Hellinger}
    &\E_{(s_{i,h},\va_{i,h})\sim d_h^{\vpi_i}(\rho)}\left[\int_\gS\sqrt{\left(\PP_{f,h}(s_{i,h+1}|s_{i,h},\va_{i,h})+2\epsilon\right)\PP_h(s_{i,h+1}|s_{i,h},\va_{i,h})} ds_{i,h+1}\right]\notag\\
    &\leq \E_{(s_{i,h},\va_{i,h})\sim d_h^{\vpi_i}(\rho)}\left[\int_\gS\sqrt{\PP_{f,h}(s_{i,h+1}|s_{i,h},\va_{i,h})\PP_h(s_{i,h+1}|s_{i,h},\va_{i,h})} ds_{i,h+1}\right] \notag\\
    &\quad + \E_{(s_{i,h},\va_{i,h})\sim d_h^{\vpi_i}(\rho)}\left[\int_\gS\sqrt{2\epsilon\PP_h(s_{i,h+1}|s_{i,h},\va_{i,h})} ds_{i,h+1}\right]\notag\\
    &\leq 1-\frac{1}{2}\E_{(s_{i,h},\va_{i,h})\sim d_h^{\vpi_i}(\rho)}\left[\int_\gS\left(\sqrt{\PP_{f,h}(s_{i,h+1}|s_{i,h},\va_{i,h})}-\sqrt{\PP_h(s_{i,h+1}|s_{i,h},\va_{i,h})} \right)^2 ds_{i,h+1}\right] + \sqrt{2\epsilon}|\gS|\notag\\
    &= 1-\E_{(s_{i,h},\va_{i,h})\sim d_h^{\vpi_i}(\rho)}\left[\hellinger{\PP_{f,h}(\cdot|s_{i,h},\va_{i,h})}{\PP_h(\cdot|s_{i,h},\va_{i,h})}\right] + \sqrt{2\epsilon}|\gS|,
\end{align}
where in the first inequality we use the fact that $\sqrt{a+b}\leq \sqrt{a}+\sqrt{b}$ for any $a,b\geq 0$, and the last line uses the definition of the Hellinger distance in \eqref{eq:Hellinger}.

Plugging \eqref{eq:construct_Hellinger} into \eqref{eq:concentration_exp_mg_part}, we have
\begin{align*}
    & \sum_{i=1}^t \log\E\left[\exp\left(-\frac{1}{2}X^{f_\epsilon}_{i,h}(\epsilon)\right)\bigg|\gF_{i-1}\right]\notag\\
    &\leq \sum_{i=1}^{t-1}\sum_{h=1}^H \log \left(1-\E_{(s_{i,h},\va_{i,h})\sim d_h^{\vpi_i}(\rho)}\left[\hellinger{\PP_{f,h}(\cdot|s_{i,h},\va_{i,h})}{\PP_h(\cdot|s_{i,h},\va_{i,h})}\right] + \sqrt{2\epsilon}|\gS|\right)\notag\\
    &\leq -\sum_{i=1}^{t-1}\sum_{h=1}^H \E_{(s_{i,h},\va_{i,h})\sim d_h^{\vpi_i}(\rho)}\left[\hellinger{\PP_{f,h}(\cdot|s_{i,h},\va_{i,h})}{\PP_h(\cdot|s_{i,h},\va_{i,h})}\right] + \sqrt{2\epsilon}|\gS|\notag\\
    & =-\sum_{i=1}^{t-1}\sum_{h=1}^H \E_{(s_{i,h},\va_{i,h})\sim d_h^{\vpi_i}(\rho)}\left[\ell(f_h,s_{i,h},\va_{i,h})\right]+ TH\sqrt{2\epsilon}|\gS|,
\end{align*}
where the second inequality follows from $\log(x)\leq x-1$ for any $x>0$, and the last line follows the definition \eqref{eq:Hellinger_loss}.

Plugging the above inequality into \eqref{eq:concentration_exp_mg_1}, we have with probability at least $1-\frac{\delta}{N+1}$:
\begin{align}\label{eq:sum_X_mg_proof}
    \forall t\in[T],f\in\gF:\quad -\sum_{i=1}^{t-1}\sum_{h=1}^H X_{i,h}^f &\leq -2\sum_{i=1}^{t-1}\sum_{h=1}^H \E_{(s_{i,h},\va_{i,h})\sim d_h^{\vpi_i}(\rho)}\left[\ell(f_h,s_{i,h},\va_{i,h})\right] \notag\\
    &\quad+ 2TH\sqrt{2\epsilon}|\gS| + 2H\log\left(\frac{(N+1)H}{\delta}\right) + 2dH\log\left(1+\frac{2\sqrt{d}}{\epsilon}\right).
\end{align}

Then analogous to \eqref{eq:sum_X_mg_proof}, we can bound $-\sum_{i=1}^{t-1}\sum_{h=1}^H Y_{i,h,n}^f$ for all $n\in[N]$ with probability at least $1-\frac{N\delta}{N+1}$ as follows:
\begin{align}\label{eq:sum_Y_mg_proof}
    \forall t\in[T],f\in\gF, n\in [N] :\quad & -\sum_{i=1}^{t-1}\sum_{h=1}^H Y_{i,h,n}^f  \leq -2\sum_{i=1}^{t-1}\sum_{h=1}^H \E_{(s_{i,h}^n,\va_{i,h}^n)\sim d_h^{\widetilde{\vpi}_{i,n}}(\rho)}\left[\ell(f_h,s_{i,h}^n,\va_{i,h}^n)\right] \notag\\
    &\quad+ 2TH\sqrt{2\epsilon}|\gS| + 2H\log\left(\frac{(N+1)H}{\delta}\right) + 2dH\log\left(1+\frac{2\sqrt{d}}{\epsilon}\right).
\end{align}

Letting $\epsilon=\frac{1}{T^2|\gS|^2}$ in \eqref{eq:sum_X_mg_proof} and \eqref{eq:sum_Y_mg_proof}, we obtain \eqref{eq:sum_X_mg} and \eqref{eq:sum_Y_mg} in Lemma~\ref{lm:bound_X_Y_mg}.


\subsubsection{Proof Lemma~\ref{lm:V_diff_bound}}\label{app:proof_V_diff_bound}

To prove Lemma~\ref{lm:V_diff_bound}, we first express the value difference sum $\sum_{t=1}^T \left|V_{\widehat{f}_{t},n}^{\widehat{\vpi}_{t}}(\rho)-V_{n}^{\widehat{\vpi}_{t}}(\rho)\right|$ on the left hand side of \eqref{eq:V_diff_bound} 
as sum of the expectation of the model estimation errors $\gE_n^{\widehat{\vpi}_{t}}(\widehat{f}_{t,h},s_h,\va_h)$.

\paragraph{Step 1: reformulating the value difference sum.}
    For any $f\in\gF$ and $\vpi=(\pi^1,\cdots,\pi^N):\gS\times[H]\rightarrow\Delta(\gA)$, we have (recall we defined the state-action visitation distribution  $d_h^\vpi(\rho)$ in \eqref{eq:state_action_visitation_finite}) for $n\in[N]$:
    \begin{align}\label{eq:V_f_decompose}
     V_{f,n}^{\vpi}(\rho) &= \E_{\forall h\in[H]:(s_h,\va_h)\sim d_h^\vpi(\rho),\atop s_{h+1}\sim\PP_h(\cdot|s_h,\va_h)}\left[\sum_{h=1}^H\left(V_{f,h,n}^{\vpi}(s_h)-V_{f,h+1,n}^{\vpi}(s_{h+1})\right)\right]\notag\\
        &=\E_{\forall h\in[H]:(s_h,\va_h)\sim d_h^\vpi(\rho),\atop s_{h+1}\sim\PP_h(\cdot|s_h,\va_h)}\left[\sum_{h=1}^H\left(Q_{f,h,n}^{\vpi}(s_h,\va_h)-\beta\log\frac{\pi^n(a_h^n|s_h^n)}{\piref^n(a_h^n|s_h^n)} - V_{f,h+1,n}^\vpi(s_{h+1})\right)\right],
    \end{align}
    where in the second line we use the fact that
    $$V_{f,h,n}^\vpi (s)=\E_{\va\sim\vpi(\cdot|s)}\left[Q_{f,h,n}^\vpi(s,\va)-\beta\log\frac{\pi^n(a^n|s^n)}{\piref^n(a^n|s^n)}\right].$$
    By the definition of $V_{n}^{\vpi}$ we have
    \begin{align}\label{eq:V_by_def}
        \forall n\in[N]:\quad V_n^{\vpi}(\rho) = \E_{\forall h\in[H]:(s_h,\va_h)\sim d_h^\vpi(\rho),\atop s_{h+1}\sim\PP_h(\cdot|s_h,\va_h)}\sum_{h=1}^H\left[r_h^n(s_h,\va_h)-\beta\log\frac{\pi^n(a_h^n|s_h^n)}{\piref^n(a_h^n|s_h^n)}\right].
    \end{align}
    To simplify the notation, we define
    \begin{align}\label{eq:Ppushforward}
        \forall g\in\gF,h\in[H]:\quad\PP_{g,h} V_{f,h+1,n}^{\vpi}(s_h,\va_h)\coloneqq \E_{s_{h+1}\sim\PP_{g,h}(\cdot|s_h,\va_h)}\left[V_{f,h+1,n}^{\vpi}(s_{h+1})\right].
    \end{align}
    Combining \eqref{eq:V_f_decompose} and \eqref{eq:V_by_def}, we have
    \begin{align}
        V_{f,n}^{\vpi}(\rho) - V_{n}^{\vpi}(\rho) &= \E_{\forall h\in[H]:(s_h,\va_h)\sim d_h^\vpi(\rho),\atop s_{h+1}\sim\PP_h(\cdot|s_h,\va_h)}\left[\sum_{h=1}^H \left( Q_{f,n}^{\vpi}(s_h,\va_h) - r_h^n(s_h,\va_h) -  V_{f,h+1,n}^\vpi(s_{h+1})\right)\right]\notag\\
    &= \sum_{h=1}^H\E_{(s_h,\va_h)\sim d_h^\vpi(\rho)}[\underbrace{\PP_{f,h} V_{f,h+1,n}^\vpi(s_h,\va_h) - \PP_h V_{f,h+1,n}^\vpi(s_h,\va_h)}_{ =: \gE_n^\vpi(f_h,s_h,\va_h)}].
    \end{align}

Therefore, we can express the value difference sum $\sum_{t=1}^T \left|V_{\widehat{f}_{t},n}^{\widehat{\vpi}_{t}}(\rho)-V_{n}^{\widehat{\vpi}_{t}}(\rho)\right|$ as sum of the expectation of the model estimation errors $\gE_n^{\widehat{\vpi}_{t}}(\widehat{f}_{t,h},s_h,\va_h)$:
\begin{align}\label{eq:V_to_res}
    \sum_{t=1}^T \left|V_{\widehat{f}_{t},n}^{\widehat{\vpi}_{t}}(\rho)-V_{n}^{\widehat{\vpi}_{t}}(\rho)\right|= \sum_{t=1}^T\sum_{h=1}^H \left| \E_{(s,\va)\sim d_h^{\widehat{\vpi}_{t}}(\rho)}\left[\gE_n^{\widehat{\vpi}_{t}}(\widehat{f}_{t,h},s,\va)\right]\right|.
\end{align}

Thus we only need to bound the right-hand side of \eqref{eq:V_to_res}.

\paragraph{Step 2: bounding the sum of model estimation errors.}

    By Assumption~\ref{asmp:function_class}, there exist $\theta_{f,h}$ and $\theta_h^\star$ in $\Theta_h$ such that $f_h(s_{h+1}|s_h,\va_h)=\phi_h(s_h,\va_h,s_{h+1})^\top\theta_{f,h}$ and $\PP_h(s_{h+1}|s_h,\va_h)=\phi_h(s_h,\va_h,s_{h+1})^\top\theta_h^\star$ for all $h\in[H]$.
    Thus we have
    \begin{align}
        \E_{(s_h,\va_h)\sim d_h^\vpi(\rho)}\left[\gE_n^\vpi(f_h,s_h,\va_h)\right] = (\theta_{f,h}-\theta_h^\star)^\top\underbrace{\E_{(s_h,\va_h)\sim d_h^\vpi(\rho)}\left[\int_{\gS}\phi_h(s_h,\va_h,s_{h+1})V_{f,h+1,n}^\vpi(s_{h+1})ds_{h+1}\right]}_{=: x_{h,n}(f,\vpi)}.
    \end{align}
    
    We let $x_{h,n}^i(f,\vpi)$ denote the $i$-th component of $x_{h,n}(f,\vpi)$, i.e., $$x_{h,n}^i(f,\vpi)=\E_{(s_h,\va_h)\sim d_h^\vpi(\rho)}\left[\int_{\gS}\phi_h^i(s_h,\va_h,s_{h+1})V_{f,h+1,n}^\vpi(s_{h+1})ds_{h+1}\right].$$
    Then we have
    \begin{align}\label{eq:x_{h,n}_i_bound}
        \forall i\in[d]:\quad |x_{h,n}^i(f,\vpi)|\leq H
    \end{align}
    (recall that by the definition of the linear mixture model (c.f. Assumption~\ref{asmp:function_class}), $\phi_h^i(s,\va,\cdot)\in\Delta(\gS)$ for all $i\in[d]$ and $(s,\va)\in\gS\times\gA$), which gives
    \begin{align}\label{eq:x_{h,n}_bound}
        \norm{x_{h,n}(f,\vpi)}_2\leq H\sqrt{d}.
    \end{align}

    For each $t\in[T]$, we define $\Lambda_{t,h}\in\R^{d\times d}$ as
\begin{align}\label{eq:Lambda_mg}
    \Lambda_{t,h}\coloneqq \lambda I_d + \sum_{i=1}^{t-1} x_{h,n}(\widehat{f}_i,\widehat{\vpi}_i)x_{h,n}(\widehat{f}_i,\widehat{\vpi}_i)^\top.
\end{align} 

We can decompose the sum of model estimation errors as follows:
\begin{align}\label{eq:dec_mg}
\sum_{t=1}^T \left| \E_{(s,\va)\sim d_h^{\widehat{\vpi}_{t}}(\rho)}\left[\gE_n^{\widehat{\vpi}_{t}}(\widehat{f}_{t,h},s,\va)\right]\right| 
    &= \underbrace{\sum_{t=1}^T \left| \langle x_{h,n}(\widehat{f}_{t},\widehat{\vpi}_{t}),\widehat{\theta}_{t,h}-\theta_h^\star\rangle\right|\mathbbm{1}\left\{\norm{x_{h,n}(\widehat{f}_{t},\widehat{\vpi}_{t})}_{\Lambda_{t,h}^{-1}}\leq 1\right\}}_{(a)}\notag\\
    & \quad+ \underbrace{\sum_{t=1}^T \left| \langle x_{h,n}(\widehat{f}_{t},\widehat{\vpi}_{t}),\widehat{\theta}_{t,h}-\theta_h^\star\rangle\right|\mathbbm{1}\left\{\norm{x_{h,n}(\widehat{f}_{t},\widehat{\vpi}_{t})}_{\Lambda_{t,h}^{-1}}> 1\right\}}_{(b)}.
\end{align}

Below we bound (a) and (b) respectively.

\paragraph{Step 1: bounding term (a).}
By the Cauchy-Schwarz inequality, we have
\begin{align}\label{eq:bound_a_1}
    (a)&\leq \sum_{t=1}^T \norm{\widehat{\theta}_{t,h}-\theta_h^\star}_{\Lambda_{t,h}}\norm{x_{h,n}(\widehat{f}_{t},\widehat{\vpi}_{t})}_{\Lambda_{t,h}^{-1}}\mathbbm{1}\left\{\norm{x_{h,n}(\widehat{f}_{t},\widehat{\vpi}_{t})}_{\Lambda_{t,h}^{-1}}\leq 1\right\}\notag\\
    &\leq \sum_{t=1}^T \norm{\widehat{\theta}_{t,h}-\theta_h^\star}_{\Lambda_{t,h}}\min\left\{ \norm{x_{h,n}(\widehat{f}_{t},\widehat{\vpi}_{t})}_{\Lambda_{t,h}^{-1}},1\right\},
\end{align}
where the last inequality uses the fact that
$$\norm{x_{h,n}(\widehat{f}_{t},\widehat{\vpi}_{t})}_{\Lambda_{t,h}^{-1}}\mathbbm{1}\left\{\norm{x_{h,n}(\widehat{f}_{t},\widehat{\vpi}_{t})}_{\Lambda_{t,h}^{-1}}\leq 1\right\}\leq \min\left\{ \norm{x_{h,n}(\widehat{f}_{t},\widehat{\vpi}_{t})}_{\Lambda_{t,h}^{-1}},1\right\}.$$

By the definition of $\Lambda_{t,h}$ (c.f. \eqref{eq:Lambda_mg}) and Assumption~\ref{asmp:function_class} we have
\begin{align}\label{eq:|w|_bound_mg}
    \norm{\widehat{\theta}_{t,h}-\theta_h^\star}_{\Lambda_{t,h}}\leq 2\sqrt{\lambda d}+\left(\sum_{i=1}^{t-1} |\langle \widehat{\theta}_{t,h}-\theta_h^\star,x_{h,n}(\widehat{f}_i,\widehat{\vpi}_i)\rangle|^2\right)^{1/2},
\end{align}
which gives
\begin{align}\label{eq:intermediate1}
    &\quad\sum_{t=1}^T \norm{\widehat{\theta}_{t,h}-\theta_h^\star}_{\Lambda_{t,h}}\min\left\{ \norm{x_{h,n}(\widehat{f}_{t},\widehat{\vpi}_{t})}_{\Lambda_{t,h}^{-1}},1\right\}\notag\\
    &\leq \sum_{t=1}^T\left(2\sqrt{\lambda d} +\left(\sum_{i=1}^{t-1} |\langle \widehat{\theta}_{t,h}-\theta_h^\star,x_{h,n}(\widehat{f}_i,\widehat{\vpi}_i)\rangle|^2\right)^{1/2}\right)\cdot\min\left\{ \norm{x_{h,n}(\widehat{f}_{t},\widehat{\vpi}_{t})}_{\Lambda_{t,h}^{-1}},1\right\}\notag\\
    &\leq \left(\sum_{t=1}^T 4\lambda d\right)^{1/2}
    \left(\sum_{t=1}^T \min\left\{ \norm{x_{h,n}(\widehat{f}_{t},\widehat{\vpi}_{t})}_{\Lambda_{t,h}^{-1}},1\right\}\right)^{1/2}\notag\\
    &\quad + \left(\sum_{t=1}^T \sum_{i=1}^{t-1} |\langle \widehat{\theta}_{t,h}-\theta_h^\star,x_{h,n}(\widehat{f}_i,\widehat{\vpi}_i)\rangle|^2\right)^{1/2}\left(\sum_{t=1}^T \min\left\{ \norm{x_{h,n}(\widehat{f}_{t},\widehat{\vpi}_{t})}_{\Lambda_{t,h}^{-1}},1\right\}\right)^{1/2},
\end{align}
where the first inequality uses \eqref{eq:|w|_bound} and the second inequality uses the Cauchy-Schwarz inequality and the fact that 
$$\min\left\{ \norm{x_{h,n}(\widehat{f}_{t},\widehat{\vpi}_{t})}_{\Lambda_{t,h}^{-1}},1\right\}^2\leq \min\left\{ \norm{x_{h,n}(\widehat{f}_{t},\widehat{\vpi}_{t})}_{\Lambda_{t,h}^{-1}},1\right\}. $$

By Lemma~\ref{lm:potential}, Lemma~\ref{lm:information_gain} and \eqref{eq:x_{h,n}_bound}, we have
\begin{align}\label{eq:|x|_bound_mg}
    \sum_{i=1}^t \min\left\{ \norm{x_{h,n}(\widehat{f}_i,\widehat{\vpi}_i)}_{\Lambda_{i,h}^{-1}},1\right\}\leq 2d\log\left(1+\frac{H^2T}{\lambda}\right)\coloneqq d_H(\lambda)
\end{align}
holds for any $\lambda>0$ and $t\in[T]$.
By \eqref{eq:|x|_bound_mg}, \eqref{eq:intermediate1} and \eqref{eq:bound_a_1}, we have
\begin{align}\label{eq:intermediate2}
    (a)\leq 2\sqrt{\lambda dT \min\{d_H(\lambda),T\}} + \left(d_H(\lambda)\sum_{t=1}^T \sum_{i=1}^{t-1} |\langle \widehat{\theta}_{t,h}-\theta_h^\star,x_{h,n}(\widehat{f}_i,\widehat{\vpi}_i)\rangle|^2\right)^{1/2}.
\end{align}

To continue, we have
\begin{align}\label{eq:bound_residual}
    |\langle \widehat{\theta}_{t,h}-\theta_h^\star,x_{h,n}(\widehat{f}_i,\widehat{\vpi}_i)\rangle|^2 &= \left|\E_{(s,\va)\sim d_h^{\widehat{\vpi}_i}(\rho)}\left[\int_\gS \left(\PP_{\widehat{f}_{t,h}}(s_{h+1}|s,\va)-\PP_h(s_{h+1}|s,\va) \right)V_{\widehat{f}_i,h+1,n}^{\widehat{\vpi}_i}(s_{h+1})ds_{h+1}\right]\right|^2\notag\\
    &\leq \E_{(s,\va)\sim d_h^{\widehat{\vpi}_i}(\rho)} \left[\left(\int_\gS \left(\PP_{\widehat{f}_{t,h}}(s_{h+1}|s,\va)-\PP_h(s_{h+1}|s,\va) \right)V_{\widehat{f}_i,h+1,n}^{\widehat{\vpi}_i}(s_{h+1})ds_{h+1}\right)^2\right]\notag\\
    &\leq 4\norm{V_{\widehat{f}_i,h+1,n}^{\widehat{\vpi}_i}(\cdot)}_\infty\E_{(s,\va)\sim d_h^{\widehat{\vpi}_{t}}(\rho)}D_{\mathsf{TV}}^2\left(\PP_{\widehat{f}_{t,h}}(\cdot|s,\va)\big\|\PP_h(\cdot|s,\va)\right)\notag\\
    &\leq 8H\E_{(s,\va)\sim d_h^{\widehat{\vpi}_i}(\rho)}\hellinger{\PP_{\widehat{f}_{t,h}}(\cdot|s,\va)}{\PP_h(\cdot|s,\va)}\notag\\
    &=8H \E_{(s,\va)\sim d_h^{\widehat{\vpi}_i}(\rho)}\ell(\widehat{f}_{t,h},s,\va),
\end{align}
where the second line uses the Cauchy-Schwarz inequality, the third line follows from H\"older's inequality, and $D_\mathsf{TV}$ denote the TV distance:
\begin{align}\label{eq:TV}
    D_{\mathsf{TV}}(P\|Q)\coloneqq \frac{1}{2}\int_\gX |P(x)-Q(x)|dx.
\end{align}
The fourth line uses the following inequality:
$$D_{\mathsf{TV}}^2(P\|Q)\leq2\hellinger{P}{Q},$$
and the fact that $\norm{V_{\widehat{f}_{t},h+1,n}^{\widehat{\vpi}_{t,h}}(\cdot)}_\infty\leq H$ (recall we assume $r(s,\va)\in[0,1]$). The last line uses \eqref{eq:Hellinger_loss}.

Plugging \eqref{eq:bound_residual} into \eqref{eq:intermediate2}, we have
\begin{align}\label{eqLbound_a_mg}
    (a)\leq 2\sqrt{d}\cdot\sqrt{\lambda T \min\{d_H(\lambda),T\}} + \left(8Hd_H(\lambda)\sum_{t=1}^T \sum_{i=1}^{t-1} \E_{(s,\va)\sim d_h^{\widehat{\vpi}_i}(\rho)}\ell(\widehat{f}_{t,h},s,\va)\right)^{1/2}.
\end{align}

\paragraph{Step 2: bounding term (b).} Now we bound (b) in \eqref{eq:dec_mg}. Note that
$$
\mathbbm{1}\left\{\norm{x_{h,n}(\widehat{f}_{t},\widehat{\vpi}_{t})}_{\Lambda_{t,h}^{-1}}> 1\right\}\leq \min\left\{ \norm{x_{h,n}(\widehat{f}_{t},\widehat{\vpi}_{t})}_{\Lambda_{t,h}^{-1}},1\right\},
$$
which gives
\begin{align}\label{eq:bound_b_pre}
    (b)\leq \sum_{t=1}^T \left| \langle x_{h,n}(\widehat{f}_{t},\widehat{\vpi}_{t}),\widehat{\theta}_{t,h}-\theta_h^\star\rangle\right|\min\left\{ \norm{x_{h,n}(\widehat{f}_{t},\widehat{\vpi}_{t})}_{\Lambda_{t,h}^{-1}},1\right\}.
\end{align}

We also have
\begin{align}\label{eq:H_bound}
    \left| \langle x_{h,n}(\widehat{f}_{t},\widehat{\vpi}_{t}),\widehat{\theta}_{t,h}-\theta_h^\star\rangle\right|=\left|\E_{(s,\va)\sim d_h^{\widehat{\vpi}_{t}}(\rho)}\left[\PP_{\widehat{f}_{t,h}} V_{\widehat{f}_{t},h+1,n}^{\widehat{\vpi}_{t,h}}(s,\va)-\PP_h V_{\widehat{f}_{t},h+1,n}^{\widehat{\vpi}_{t,h}}(s,\va)\right]\right|\leq H.
\end{align}

Combining the \eqref{eq:H_bound},\eqref{eq:|x|_bound_mg} with \eqref{eq:bound_b_pre}, we have
\begin{align}\label{eq:bound_b_mg}
    (b)\leq H\min\{T, d_H(\lambda)\}.
\end{align}

\paragraph{Step 3: combining everything together.}
Plugging \eqref{eqLbound_a_mg} and \eqref{eq:bound_b_mg} into \eqref{eq:dec_mg}, we have
\begin{align*}
    &\sum_{t=1}^T \left| \E_{(s,\va)\sim d_h^{\widehat{\vpi}_{t}}(\rho)}\left[\gE_n^{\widehat{\vpi}_{t}}(\widehat{f}_{t,h},s,\va)\right]\right|\notag\\
    &\leq 2\sqrt{d}\cdot\sqrt{\lambda T \min\{d_H(\lambda),T\}} + \left(8Hd_H(\lambda)\sum_{t=1}^T \sum_{i=1}^{t-1} \E_{(s,\va)\sim d_h^{\widehat{\vpi}_i}(\rho)}\ell(\widehat{f}_{t,h},s,\va)\right)^{1/2}+H \min\{T, d_H(\lambda)\}\notag\\
    &\leq \left(\frac{8Hd_H(\lambda)}{\eta}\cdot\eta\sum_{t=1}^T \sum_{i=1}^{t-1}  \E_{(s,\va)\sim d_h^{\widehat{\vpi}_i}(\rho)}\ell(\widehat{f}_{t,h},s,\va)\right)^{1/2} + \left(\sqrt{d}+H\right) \min\{d_H(\lambda),T\} +\sqrt{d} \lambda T\notag\\
    &\leq \frac{4d_H(\lambda)H}{\eta} +\frac{\eta}{2}\sum_{t=1}^T \sum_{i=1}^{t-1} \E_{(s,\va)\sim d_h^{\widehat{\vpi}_i}(\rho)}\ell(\widehat{f}_{t,h},s,\va) + \left(\sqrt{d}+H\right) \min\{d_H(\lambda),T\} +\sqrt{d} \lambda T
\end{align*}
for any $\eta>0$, where the second and third inequalities both use the fact that $\sqrt{ab}\leq \frac{a+b}{2}$ for any $a,b\geq 0$. 

Finally, combining \eqref{eq:V_to_res} with the above inequality, we have \eqref{eq:V_diff_bound}.

\section{Extension to the Infinite-horizon Setting}\label{app:infinite}

In this section, we consider the $N$-player general-sum episodic Markov game with infinite horizon denoted as $\gM_\PP\coloneqq(\gS,\gA,\PP, r,\gamma)$ as a generalization of the finite-horizon case in the main paper, where $\gamma\in[0,1)$ is the discounted factor, and $\PP:\gS\times\gA\rightarrow\Delta(\gS)$ is the homogeneous transition kernel: the probability of transitioning from state $s$ to state $s'$ by the action $\va=(a^1,\cdots,a^n)$ is $\PP(s'|s,\va)$. For the infinite horizon case, the KL-regularized value function is defined as
\begin{align}\label{eq:V_beta_infinite}
    \forall s\in\gS:\quad V_{n}^{\vpi}(s)&\coloneqq \E_{\PP,\vpi}\left[\sum_{h=0}^\infty \gamma^h\left(r^n(s_h,\va_h)-\beta\log\frac{\pi^n(a_h|s_h)}{\piref^n(a_h|s_h)}\right)\bigg|s_0=s\right]\notag\\
    &=\frac{1}{1-\gamma}\E_{(\bar s,\bar\va)\sim d^\vpi(s)}\left[r^n(\bar s,\bar \va)-\beta\log\frac{\pi^n(\bar a^n|\bar s)}{\piref^n(\bar a^n|\bar s)}\right],
\end{align}
where $s_h$ and $\va_h$ are the state and action at timestep $h$, respectively, $d^\vpi(s)\in\Delta(\gS\times\gA)$ is the \textit{discounted state-action visitation distribution} under policy $\vpi$ starting from state $s$:
\begin{align}\label{eq:state_action_visitation_infinite}
    d^\vpi_{\bar s, \bar \va}(s)\coloneqq (1-\gamma)\sum_{h=0}^\infty \gamma^h \PP^{\vpi}(s_h = \bar s, \va_h =  \bar \va|s_0=s).
\end{align}
We assume $\rho\in\Delta(\gS)$ is the initial state distribution, i.e. $s_0\sim\rho$.
We define $d^\vpi(\rho)\coloneqq \E_{s_0\sim\rho}[d^\vpi(s_0)]$ as the discounted state-action visitation distribution under policy $\vpi$ starting from the initial state distribution $\rho$.
The KL-regularized Q-function is defined as
\begin{align}\label{eq:Q_beta_infinite}
    \forall (s,\va)\in\gS\times\gA:&\quad Q_{n}^{\vpi}(s,\va)\coloneqq r^n(s,\va) +  \gamma\E_{s'\sim\PP(\cdot|s,\va)}\left[V_{n}^{\vpi}(s')\right].
\end{align}

We let $\gF$ denote the function class of the estimators of the transition kernel of the Markov game, and we denote $f=\PP_f\in\gF$.
Without otherwise specified, we assume the other notations and settings are the same as the finite-horizon case stated in Section~\ref{sec:markov}.


\subsection{Algorithm development}\label{sec:markov_alg_infinite}

The algorithm for solving the (KL-regularized) Markov game is shown in Algorithm~\ref{alg:markov_game_infinite}, where in \eqref{eq:model_update_mg_infinite} we set the loss function at each iteration $t$ as the negative log-likelihood of the transition kernel estimator $f$:
\begin{align}\label{eq:loss_mg_infinite}
    \gL_t(f)\coloneqq \sum_{(s,\va,s')\in\gD_{t-1}}-\log \PP_f(s'|s,\va).
\end{align}
Except for the loss function, the main change in Algorithm~\ref{alg:markov_game_infinite} is that we need to sample the state-action pair $(s,\va)$ from the discounted state-action visitation distribution $d^\vpi(\rho)$, and sample the next state $s'$ from the transition kernel $\PP(\cdot|s,\va)$, which can be done by calling Algorithm~\ref{alg:sample}. Algorithm~\ref{alg:sample} is adapted from Algorithm~3 in \citet{yuan2023linear}, see also Algorithm~5 in \citet{yang2023federated}. Algorithm~\ref{alg:sample} satisfies $\E[h+1]=\frac{1}{1-\gamma}$, and $\PP(s_h=s,\va_h=\va)=d^\vpi(\rho)$~\citep{yuan2023linear}.

\begin{algorithm}[!thb]
    \caption{Value-incentive Infinite-horizon Markov Game}
    \label{alg:markov_game_infinite}
    \begin{algorithmic}[1]
    \STATE \textbf{Input:} reference policies $\allpiref$, KL coefficient $\beta$, initial state distribution $\rho$, initial transition kernel estimator $f_0\in\gF$, regularization coefficient $\alpha>0$, iteration number $T$.
    \STATE \textbf{Initialization:} dataset $\gD_0^n\coloneqq \emptyset$, $\forall n\in[N]$. $\gD_0 = \cup_{n=1}^N \gD_0^n$.
    \FOR{$t = 1,\cdots,T$}
    \STATE $\vpi_t \leftarrow \gamesolver(\gM_{f_{t-1}})$. \trianglecomment{$\gamesolver(\gM_f)$ returns a CCE or NE of game $\gM_f$.}
    \STATE Model update: Update the estimator $f_t$ by minimizing the following objective:
    \begin{align}\label{eq:model_update_mg_infinite}         
        f_t=\argmin_{f\in\gF}\sum_{(s,\va,s')\in\gD_{t-1}}-\log \PP_f(s'|s,\va) -\alpha \sum_{n=1}^N V_{f,n}^{\star,\vpi_t^{-n}}(\rho).
    \end{align}
    \STATE Compute best-response policies $\{\widetilde{\pi}_t^n\}_{n\in[N]}$:
    \begin{align}\label{eq:policy_update_2_mg_infinite}
        \text{For all }n\in[N]:\quad \widetilde{\pi}_t^n& = \argmax_{\pi^n\in\Delta(\gA_n)} V_{f_t,n}^{\pi^n,\vpi_t^{-n}}(\rho).
    \end{align}
    \STATE Data collection: sample $(s_t,\va_t,s_t')\leftarrow \sample(\vpi_t,\rho)$.
    For all $n\in[N]$, sample 
    $(s_{t}^n,\va_{t}^n,{s_{t}^n}')\leftarrow \sample((\widetilde{\pi}_t^n,\vpi_t^{-n}),\rho)$, and update the dataset $\gD_t^n = \gD_{t-1}^n\cup\{(s_t,\va_t,s_t'),(s_t^n,\va_{t}^n,{s_t^n}')\}$. $\gD_t = \cup_{n=1}^N \gD_t^n$.
    \trianglecomment{$\sample(\vpi,\rho)$ returns $(s,\va)\sim d^\vpi(\rho)$ and $s'\sim \PP(\cdot|s,\va)$, see Algorithm~\ref{alg:sample}.}
    \ENDFOR 
    \end{algorithmic}
\end{algorithm}

\begin{algorithm}[!thb]
    \caption{$\sample$ for $(s,\va)\sim d^\vpi(\rho)$ and $s'\sim \PP(\cdot|s,\va)$}
    \label{alg:sample}
    \begin{algorithmic}[1]
    \STATE \textbf{Input:} policy $\vpi$, initial state distribution $\rho$, player index $n$.
    \STATE \textbf{Initialization:} $s_0\sim \rho$, $\va_0\sim \vpi(\cdot|s_0)$, time step $h=0$, variable $X\sim$ Bernoulli($\gamma$).
    \WHILE{$X = 1$}
        \STATE Sample $s_{h+1}\sim \PP(\cdot|s_h,\va_h)$
        \STATE Sample $\va_{h+1}\sim \vpi(\cdot|s_{h+1})$
        \STATE $h\leftarrow h+1$
        \STATE $X\sim$ Bernoulli($\gamma$)
    \ENDWHILE
    \STATE Sample $s_{h+1}\sim \PP(\cdot|s_h,\va_h)$
    \RETURN $(s_h,\va_h, s_{h+1})$.
    \end{algorithmic}
\end{algorithm}


\subsection{Theoretical guarantee}\label{sec:markov_analysis_infinite}

We first state our assumptions on the function class for Markov game with infinite horizon.

\begin{asmp}[linear mixture model, infinite horizon]\label{asmp:function_class_infinite}
    The function class $\gF$ is
    $$\gF\coloneqq\left\{f|f(s,\va,s')=\phi(s,\va,s')^\top\theta,\forall (s,\va,s')\in\gS\times\gA\times\gS,\theta\in\Theta\right\},$$
    where $\phi:\gS\times\gA\times\gS\rightarrow\R^d$ is the known feature map, $\norm{\phi(s,\va,s')}_2\leq 1$ for all $(s,\va,s')$, and $\Theta\subseteq\mathbb{B}^d_2(\sqrt{d})$. Moreover,  for each $f\in\gF$ and $(s,\va)\in\gS\times\gA$, $f(\cdot|s,\va)\in\Delta(\gS)$.
\end{asmp}
We also assume the function class $\gF$ is expressive enough to describe the true transition kernel of the Markov game.
\begin{asmp}[realizability]\label{asmp:realizable_infinite}
    There exists $f^\star\in\gF$ such that $\PP_{f^\star}=\PP$.
\end{asmp}

Theorem~\ref{thm:regret_MG_infinite} states our main result for the regret of the infinite-horizon online Markov game, whoes proof is deferred to Appendix~\ref{app:proof_thm_regret_MP_infinite}.

\begin{thm}\label{thm:regret_MG_infinite}
    Under Assumption~\ref{asmp:function_class_infinite} and Assumption~\ref{asmp:realizable_infinite}, if setting the regularization coefficient $\alpha$ as
    $$\alpha = \frac{(1-\gamma)^{3/2}}{\gamma}\sqrt{\frac{\log\left(\frac{N}{\delta}\right) + d\log\left(d|\gS|T\right)}{d\log\left(1+\frac{T^{3/2}}{(1-\gamma)^2\sqrt{d}}\right)}T},$$
    then for any $\beta\geq 0$, with any initial state distribution $\rho$, transition kernel estimator $f_0\in\gF$ and reference policy $\allpiref$, the regret of Algorithm~\ref{alg:markov_game} satisfies the following bound with probability at least $1-\delta$ for any $\delta\in(0,1)$:
    \begin{align}\label{eq:regret_MG_bd_infinite}
        \forall T\in\NN_+:\quad\regret(T)\leq \widetilde{\gO}\left( \frac{\gamma d\sqrt{T}}{(1-\gamma)^{3/2}}\cdot \sqrt{\frac{1}{d}\log\left(\frac{N}{\delta}\right) + \log\left(d|\gS|T\right)}\right).
    \end{align}
\end{thm}

Note that
\begin{align}
    \min_{t\in[T]}\gap{\vpi_t}\leq \frac{\regret(T)}{T},
\end{align}
similar to earlier arguments, Theorem~\ref{thm:regret_MG_infinite} also implies an order of $\widetilde{\gO}\left(\frac{\gamma^2 N d^2}{(1-\gamma)^4 \varepsilon^2}\right)$ sample complexity for Algorithm~\ref{alg:markov_game} to find an $\varepsilon$-NE or $\varepsilon$-CCE of $\gM_\PP$.

\subsection{Proof of Theorem~\ref{thm:regret_MG_infinite}}\label{app:proof_thm_regret_MP_infinite}

The proof of Theorem~\ref{thm:regret_MG_infinite} resembles that of Theorem~\ref{thm:regret_MG}. We repeat some of the proof for clarity and completeness.
Here we also have \eqref{eq:regret_dec2_mg}, and will upper bound each term in \eqref{eq:regret_dec2_mg} separately.

\paragraph{Step 1: bounding (i).} 
By Assumption~\ref{asmp:realizable_infinite} we know that there exists $f^\star\in\gF$ such that $f^\star\coloneqq \PP = \PP_{f^\star}$

 \eqref{eq:loss_diff_mg} also holds here, and we define random variables $X_t^f$ and $Y_t^f$ as
\begin{align}\label{eq:X_t_mg_infinite}
    X_t^f\coloneqq \log \left(\frac{\PP(s'_t|s_t,\va_t)}{\PP_f(s'_t|s_t,\va_t)}\right)\quad\text{and}\quad Y_{t,n}^f\coloneqq \log \left(\frac{\PP({s_t^n}'|s_t^n,\va_t^n)}{\PP_f({s_t^n}'|s_t^n,\va_t^n)}\right),\quad \forall n\in[N].
\end{align}
Then by the definition of the loss function~\eqref{eq:loss_mg_infinite}, we have
\begin{align}\label{eq:loss_diff_decompose_mg_infinite}
    \gL_t(f^\star)-\gL_t(f) = -\sum_{i=1}^{t-1}\sum_{n=1}^N \left(X_i^f+Y_{i,n}^f\right).
\end{align}

Same as in the proof of Theorem~\ref{thm:regret_MG}, we define
\begin{align}\label{eq:pi_t_n_infinite}
    \widetilde{\vpi}_{t,n}\coloneqq (\widetilde{\pi}_t^n, \vpi_t^{-n}),\quad \forall n\in[N].
\end{align}
We also define
\begin{align}\label{eq:Hellinger_loss_infinite}
    \ell(f,s,\va)\coloneqq \hellinger{\PP_{f}(\cdot|s,\va)}{\PP(\cdot|s,\va)}.
\end{align}

In the following lemma we provide a concentration result for the random variables $X_{t}^f$ and $Y_{t,n}^f$ in \eqref{eq:loss_diff_decompose_mg_infinite}.
\begin{lm}\label{lm:bound_X_Y_mg_infinite}
    When Assumption~\ref{asmp:function_class_infinite},\ref{asmp:realizable_infinite} hold, for any $\delta \in (0,1)$, with probability at least $1-\delta$, we have
    \begin{align}\label{eq:sum_X_mg_infinite}
        \forall t\in[T],f\in\gF:\quad -\sum_{i=1}^{t-1} X_i^f &\leq -2\sum_{i=1}^{t-1} \E_{(s_i,\va_i)\sim d^{\vpi_i}(\rho)}\left[\ell(f,s_i,\va_i)\right] \notag\\
        &\quad+ 2\sqrt{2} + 2\log\left(\frac{N+1}{\delta}\right) + 2d\log\left(1+2\sqrt{d}T^2|\gS|^2\right),
    \end{align}
    and 
    \begin{align}\label{eq:sum_Y_mg_infinite}
        \forall t\in[T],f\in\gF,n\in[N]:\quad -\sum_{i=1}^{t-1} Y_{i,n}^f &\leq -2\sum_{i=1}^{t-1} \E_{(s_i^n,\va_i^n)\sim d^{\widetilde{\vpi}_{i,n}}(\rho)}\left[\ell(f,s_i^n,\va_i^n)\right] \notag\\
        &\quad+ 2\sqrt{2} + 2\log\left(\frac{N+1}{\delta}\right) + 2d\log\left(1+2\sqrt{d}T^2|\gS|^2\right).
    \end{align}
\end{lm}
The proof of Lemma~\ref{lm:bound_X_Y_mg_infinite} is provided in Appendix~\ref{app:proof_bound_X_Y_mg_infinite}.

By \eqref{eq:loss_diff_mg}, \eqref{eq:loss_diff_decompose_mg_infinite} and Lemma~\ref{lm:bound_X_Y_mg_infinite}, we have with probability at least $1-\delta$:
\begin{align}\label{eq:a_bound'}
    \text{(i)}&\leq -\frac{2}{N\alpha}\sum_{n=1}^N\left\{\sum_{t=1}^T\sum_{i=1}^{t-1} 
    \E_{(s_i,\va_i)\sim d^{\vpi_i}(\rho)}\left[\ell(f_t,s_i,\va_i)\right]+\sum_{t=1}^T\sum_{i=1}^{t-1} 
    \E_{(s_i^n,\va_i^n)\sim d^{\widetilde{\vpi}_{i,n}}(\rho)}\left[\ell(f_t,s_i^n,\va_i^n)\right]\right\}\notag\\
    &\quad\,\, + \frac{4T}{\alpha}\left(\sqrt{2} + \log\left(\frac{N+1}{\delta}\right) + d\log\left(1+\sqrt{d}|\gS|^2T^2\right)\right).
\end{align}

\paragraph{Step 2: bounding (ii), (iii) and (iv).} To bound (ii), (iii) and (iv), we introduce the following lemma.
\begin{lm}\label{lm:V_diff_bound_infinite}
    Under Assumption~\ref{asmp:function_class_infinite} and Assumption~\ref{asmp:realizable_infinite}, for any $n\in[N]$, $\beta\geq 0$, $\{\widehat{\vpi}_t: \gS\rightarrow\Delta(\gA)\}_{t\in[T]}$ and $\{\widehat{f}_t\}_{t\in[T]}\subset \gF$,  we have
    \begin{align}\label{eq:V_diff_bound_infinite}
       \sum_{t=1}^T \left|V_{\widehat{f}_t,n}^{\widehat{\vpi}_t}(\rho)-V_{n}^{\widehat{\vpi}_t}(\rho)\right|&\leq \frac{\gamma}{1-\gamma}\bigg(\frac{\eta}{2}\sum_{t=1}^T \sum_{i=1}^{t-1} \E_{(s,\va)\sim d^{\widehat{\vpi}_i}(\rho)}\ell(\widehat{f}_t,s,\va) \notag\\
       &\quad+\frac{4d_\gamma(\lambda)}{(1-\gamma)\eta}
       + \left(\sqrt{d}+\frac{1}{1-\gamma}\right) \min\{d_\gamma(\lambda),T\} +\sqrt{d} \lambda T\bigg)
    \end{align}
    for any $\eta>0$ and $\lambda>0$, where $d_\gamma(\lambda)$ is defined as
    $$d_\gamma(\lambda)\coloneqq 2d\log\left(1+\frac{T}{d\lambda}\right).$$
\end{lm}
The proof of Lemma~\ref{lm:V_diff_bound_infinite} is provided in Appendix~\ref{app:proof_V_diff_bound_infinite}.

Now we are ready to bound (ii), (iii) and (iv). To bound (ii), letting $\widehat{f}_t=f_t$ and
$\widehat{\vpi}_t=\widetilde{\vpi}_{t,n}$ for each $n$ in Lemma~\ref{lm:V_diff_bound_infinite} (recall we define $\widetilde{\vpi}_{t,n}\coloneqq (\widetilde{\pi}_t^n,\vpi_t^{-n})$ in \eqref{eq:pi_t_n}), we have for any $\eta>0$:
\begin{align}\label{eq:b_bound'} 
    \text{(ii)}&\leq \frac{\gamma}{1-\gamma}\cdot\frac{\eta}{2N}\sum_{n=1}^N\sum_{t=1}^T \sum_{i=1}^{t-1} \E_{(s,\va)\sim d^{\widetilde{\vpi}_{i,n}}(\rho)}\ell(f_t,s,\va) \notag\\
    &\quad+\frac{\gamma}{1-\gamma}\left(\frac{4d_\gamma(\lambda)}{(1-\gamma)\eta}
    + \left(\sqrt{d}+\frac{1}{1-\gamma}\right) \min\{d_\gamma(\lambda),T\} +\sqrt{d} \lambda T\right).
\end{align}
Letting $\widehat{f}_t=f_{t-1}$ and
$\widehat{\vpi}_t=\widetilde{\vpi}_{t,n}$ for each $n$ in Lemma~\ref{lm:V_diff_bound_infinite}, we can bound (iii) as follows:
\begin{align}\label{eq:c_bound'_pre}
    \text{(iii)}&\leq \frac{\gamma}{1-\gamma}\cdot\frac{\eta}{2N}\sum_{n=1}^N\sum_{t=1}^T \sum_{i=1}^{t-1} \E_{(s,\va)\sim d^{\widetilde{\vpi}_{i,n}}(\rho)}\ell(f_{t-1},s,\va) \notag\\
    &\quad+\frac{\gamma}{1-\gamma}\left(\frac{4d_\gamma(\lambda)}{(1-\gamma)\eta}
    + \left(\sqrt{d}+\frac{1}{1-\gamma}\right) \min\{d_\gamma(\lambda),T\} +\sqrt{d} \lambda T\right).
\end{align}
To continue to bound the first term, note that
\begin{align}
    \sum_{t=1}^T \sum_{i=1}^{t-1} \E_{(s,\va)\sim d^{\widetilde{\vpi}_{i,n}}(\rho)}\ell(f_{t-1},s,\va) &\leq \sum_{t=1}^T \sum_{i=1}^{t-2} \E_{(s,\va)\sim d^{\widetilde{\vpi}_{i,n}}(\rho)}\ell(f_{t-1},s,\va) + T\notag\\
    &= \sum_{t=0}^{T-1} \sum_{i=1}^{t-1} \E_{(s,\va)\sim d^{\widetilde{\vpi}_{i,n}}(\rho)}\ell(f_{t-1},s,\va) + T\notag\\
    &\leq \sum_{t=1}^{T} \sum_{i=1}^{t-1} \E_{(s,\va)\sim d^{\widetilde{\vpi}_{i,n}}(\rho)}\ell(f_{t-1},s,\va) + T,
\end{align}
where the first inequality uses the fact that
\begin{align}
    \ell(f,s,\va)=\hellinger{\PP_{f}(\cdot|s,\va)}{\PP(\cdot|s,\va)}\leq 1,
\end{align}
the second line shifts the index of $t$ by 1, and the last line follows by noticing the first summand is 0 at $t=0$.

Plugging the above relation back to \eqref{eq:c_bound'_pre} leads to
\begin{align}\label{eq:c_bound'} 
    \text{(iii)}
    &\leq \frac{\gamma}{1-\gamma}\cdot\frac{\eta}{2N}\sum_{n=1}^N\sum_{t=1}^{T} \sum_{i=1}^{t-1} \E_{(s,\va)\sim d^{\widetilde{\vpi}_{i,n}}(\rho)}\ell(f_t,s,\va) \notag\\
    &\quad+\frac{\gamma}{1-\gamma}\left(\frac{4d_\gamma(\lambda)}{(1-\gamma)\eta}
    + \left(\sqrt{d}+\frac{1}{1-\gamma}\right) \min\{d_\gamma(\lambda),T\} +\sqrt{d} \lambda T +\frac{\eta}{2}T\right).
\end{align}

Finally, similar to \eqref{eq:c_bound'}, letting $\widehat{f}_t=f_{t-1}$, $\widehat{\vpi}_t=\vpi_t$ for each $n$ and $\eta\leftarrow2\eta$ in Lemma~\ref{lm:V_diff_bound_infinite}, we can bound (iv) as follows:
\begin{align}\label{eq:d_bound'} 
    \text{(iv)} &\leq \frac{\gamma}{1-\gamma}\cdot\frac{\eta}{N}\sum_{n=1}^N\sum_{t=1}^{T} \sum_{i=1}^{t-1} \E_{(s,\va)\sim d^{\vpi_i}(\rho)}\ell(f_t,s,\va) \notag\\
    &\quad+\frac{\gamma}{1-\gamma}\left(\frac{2d_\gamma(\lambda)}{(1-\gamma)\eta}
    + \left(\sqrt{d}+\frac{1}{1-\gamma}\right) \min\{d_\gamma(\lambda),T\} +\sqrt{d} \lambda T +\eta T\right).
\end{align}

\paragraph{Step 3: combining the bounds.} Letting $\eta = \frac{2(1-\gamma)}{\gamma\alpha}$ in \eqref{eq:b_bound'}, \eqref{eq:c_bound'} and \eqref{eq:d_bound'}, we have with probability at least $1-\delta$:
\begin{align*}
    \regret(T)&\leq\frac{4T}{\alpha}\left(\sqrt{2} + \log\left(\frac{N+1}{\delta}\right) + d\log\left(1+\sqrt{d}|\gS|^2T^2\right)\right)\notag\\
    &\quad+\frac{\gamma}{1-\gamma}\left(\frac{5\gamma\alpha d_\gamma(\lambda)}{(1-\gamma)^2}
    + 3\left(\sqrt{d}+\frac{1}{1-\gamma}\right) \min\{d_\gamma(\lambda),T\} +3\sqrt{d} \lambda T +\frac{3(1-\gamma)}{\gamma\alpha} T\right).
\end{align*}
By setting
\begin{align}
    \lambda =\sqrt{\frac{d}{T}},\quad \alpha = \frac{(1-\gamma)^{3/2}}{\gamma}\sqrt{\frac{\log\left(\frac{N+1}{\delta}\right) + d\log\left(1+\sqrt{d}|\gS|^2T^2\right)}{d\log\left(1+\frac{T^{3/2}}{(1-\gamma)^2\sqrt{d}}\right)}T}
\end{align}
in the above expression, we have with probability at least $1-\delta$:
\begin{align}
    \regret(T)&\leq \frac{4(1+\sqrt{2})\gamma}{(1-\gamma)^{3/2}}\sqrt{\frac{d\log\left(1+\frac{T^{3/2}}{(1-\gamma)^2\sqrt{d}}\right)}{\log\left(\frac{N+1}{\delta}\right) + d\log\left(1+\sqrt{d}|\gS|^2T^2\right)}}\cdot \sqrt{T}\notag\\
    &\quad + \frac{\gamma d\sqrt{T}}{(1-\gamma)^{3/2}}\cdot 14\sqrt{\left(\frac{1}{d}\log\left(\frac{N+1}{\delta}\right) + \log\left(1+\sqrt{d}|\gS|^2T^2\right)\right)\log\left(1+\frac{T^{3/2}}{(1-\gamma)^2\sqrt{d}}\right)}\notag\\
    &\quad + \frac{6\gamma}{1-\gamma}\left(\sqrt{d}+\frac{1}{1-\gamma}\right) d\log\left(1+\frac{T^{3/2}}{(1-\gamma)^2\sqrt{d}}\right) +\frac{3\gamma}{1-\gamma}d\sqrt{T},
\end{align}
which gives the desired result.


\subsubsection{Proof of Lemma~\ref{lm:bound_X_Y_mg_infinite}}\label{app:proof_bound_X_Y_mg_infinite}

Same as in \eqref{eq:covering}, for the parameter space $\Theta$, by Assumption~\ref{asmp:function_class_infinite} and Lemma~\ref{lm:covering} we have
\begin{align}\label{eq:covering_mg_infinite}
    \log \gN(\Theta,\epsilon,\norm{\cdot}_2)\leq d\log\left(1+\frac{2\sqrt{d}}{\epsilon}\right)
\end{align}
for any $\epsilon>0$. Thus there exists an $\epsilon$-net $\Theta_\epsilon$ of $\Theta$ ($\Theta_\epsilon\subset\Theta$) such that $\log|\Theta_\epsilon|\leq d\log\left(1+\frac{2\sqrt{d}}{\epsilon}\right)$.
Define
\begin{align*}
    \gF_\epsilon \coloneqq \left\{f\in\gF: f(s,\va,s')=\phi(s,\va,s')^\top\theta,\theta\in\Theta_\epsilon\right\}.
\end{align*}

For any $f\in\gF$, there exists $\theta\in\Theta$ such that $f(s,\va,s')=\phi(s,\va,s')^\top\theta$. And there exists $\theta_\epsilon\in\Theta_\epsilon$ such that $\norm{\theta-\theta_\epsilon}_2\leq \epsilon$. We let $f_\epsilon(s,\va,s')=\phi(s,\va,s')^\top\theta_\epsilon$. Then $f_\epsilon\in\gF_\epsilon$, and we have
\begin{align}\label{eq:diff_eps_infinite}
    |\PP_f(s'|s,\va)-\PP_{f_\epsilon}(s'|s,\va)|=|\phi(s,\va,s')^\top(\theta-\theta_\epsilon)|\leq \epsilon,
\end{align}
from which we deduce
\begin{align}\label{eq:diff_net_approx_infinite}
    \forall t\in[T]:\quad -X_t^f\leq -\log\left(\frac{\PP(s'_t|s_t,\va_t)}{\PP_{f_\epsilon}(s'_t|s_t,\va_t)+\epsilon}\right)\coloneqq -X_t^{f_\epsilon}(\epsilon).
\end{align}

Let $\gF_t\coloneqq \sigma(\gD_t)$ be the $\sigma$-algebra generated by the data $\gD_t$. 
By Lemma~\ref{lm:martingale_exp} we have 
with probability at least $1-\frac{\delta}{N+1}$:
\begin{align}\label{eq:exp_concentration_net_infinite}
   \forall t\in[T],f_\epsilon\in\gF_\epsilon: \quad -\frac{1}{2}\sum_{i=1}^{t-1} X^{f_\epsilon}_i(\epsilon) &\leq \sum_{i=1}^{t-1} \log\E\left[\exp\left(-\frac{1}{2}X^{f_\epsilon}_i(\epsilon)\right)\bigg|\gF_{i-1}\right] \notag\\
   &\quad+ \log\left(\frac{N+1}{\delta}\right) + d\log\left(1+\frac{2\sqrt{d}}{\epsilon}\right).
\end{align}

Thus we have
\begin{align}\label{eq:concentration_exp_mg_infinite_1}
    \forall t\in[T],f\in\gF:\quad -\frac{1}{2}\sum_{i=1}^{t-1} X_i^f &\overset{\eqref{eq:diff_net_approx}}{\leq} -\frac{1}{2}\sum_{i=1}^{t-1} X^{f_\epsilon}_i(\epsilon)\notag\\
    &\overset{\eqref{eq:exp_concentration_net}}{\leq} \sum_{i=1}^t \log\E\left[\exp\left(-\frac{1}{2}X^{f_\epsilon}_i(\epsilon)\right)\bigg|\gF_{i-1}\right] + \log\left(\frac{N+1}{\delta}\right) + d\log\left(1+\frac{2\sqrt{d}}{\epsilon}\right).
\end{align}
We can further bound the first term in \eqref{eq:concentration_exp_mg_infinite_1} as follows:
\begin{align}
    &\quad \sum_{i=1}^t \log\E\left[\exp\left(-\frac{1}{2}X^{f_\epsilon}_i(\epsilon)\right)\bigg|\gF_{i-1}\right]\notag\\
    &= \sum_{i=1}^{t-1} \log\E_{(s_i,\va_i)\sim d^{\vpi_i}(\rho),\atop s_i'\sim\PP(\cdot|s_i,\va_i)}\left[\sqrt{\frac{\PP_{f_\epsilon}(s'_i|s_i,\va_i)+\epsilon}{\PP(s'_i|s_i,\va_i)}}\right]\notag\\
    &= \sum_{i=1}^{t-1} \log\E\left[\sqrt{\frac{\PP_{f_\epsilon}(s'_i|s_i,\va_i)+\epsilon}{\PP(s'_i|s_i,\va_i)}}\bigg|\gF_{s-1}\right]\notag\\
    &= \sum_{i=1}^{t-1} \log\E_{(s_i,\va_i)\sim d^{\vpi_i}(\rho)}\left[\int_\gS\sqrt{\left(\PP_{f_\epsilon}(s'_i|s_i,\va_i)+\epsilon\right)\PP(s'_i|s_i,\va_i)} ds_i'\right]\notag\\
    &\overset{\eqref{eq:diff_eps}}{\leq} \sum_{i=1}^{t-1} \log\E_{(s_i,\va_i)\sim d^{\vpi_i}(\rho)}\left[\int_\gS\sqrt{\left(\PP_{f}(s'_i|s_i,\va_i)+2\epsilon\right)\PP(s'_i|s_i,\va_i)} ds_i'\right].
\end{align}
Moreover, we have
\begin{align}\label{eq:construct_Hellinger_infinite}
    &\quad\E_{(s_i,\va_i)\sim d^{\vpi_i}(\rho)}\left[\int_\gS\sqrt{\left(\PP_{f}(s'_i|s_i,\va_i)+2\epsilon\right)\PP(s'_i|s_i,\va_i)} ds_i'\right]\notag\\
    &\leq \E_{(s_i,\va_i)\sim d^{\vpi_i}(\rho)}\left[\int_\gS\sqrt{\PP_{f}(s'_i|s_i,\va_i)\PP(s'_i|s_i,\va_i)} ds_i'\right] + \E_{(s_i,\va_i)\sim d^{\vpi_i}(\rho)}\left[\int_\gS\sqrt{2\epsilon\PP(s'_i|s_i,\va_i)} ds_i'\right]\notag\\
    &\leq 1-\frac{1}{2}\E_{(s_i,\va_i)\sim d^{\vpi_i}(\rho)}\left[\int_\gS\left(\sqrt{\PP_{f}(s'_i|s_i,\va_i)}-\sqrt{\PP(s'_i|s_i,\va_i)} \right)^2 ds_i'\right] + \sqrt{2\epsilon}|\gS|\notag\\
    &= 1-\E_{(s_i,\va_i)\sim d^{\vpi_i}(\rho)}\left[\hellinger{\PP_{f}(\cdot|s_i,\va_i)}{\PP(\cdot|s_i,\va_i)}\right] + \sqrt{2\epsilon}|\gS|,
\end{align}
where the first inequality we use the fact that $\sqrt{a+b}\leq \sqrt{a}+\sqrt{b}$ for any $a,b\geq 0$.

Then combining \eqref{eq:concentration_exp_mg_infinite_1}, \eqref{eq:construct_Hellinger_infinite} and \eqref{eq:Hellinger_loss_infinite}, we have
\begin{align*}
    \forall t\in[T],f\in\gF:\quad -\frac{1}{2}\sum_{i=1}^{t-1} X_i^f &\leq -\sum_{i=1}^{t-1} \E_{(s_i,\va_i)\sim d^{\vpi_i}(\rho)}\left[\ell(f,s_i,\va_i)\right] \notag\\
    &\quad+ T\sqrt{2\epsilon}|\gS| + \log\left(\frac{N+1}{\delta}\right) + d\log\left(1+\frac{2\sqrt{d}}{\epsilon}\right),
\end{align*}
where we use the fact that $\log(x)\leq x-1$ for any $x>0$. Multiplying both sides by 2, we have with probability at least $1-\frac{\delta}{N+1}$:
\begin{align}
    \forall t\in[T],f\in\gF:\quad -\sum_{i=1}^{t-1} X_i^f &\leq -2\sum_{i=1}^{t-1} \E_{(s_i,\va_i)\sim d^{\vpi_i}(\rho)}\left[\ell(f,s_i,\va_i)\right] \notag\\
    &\quad+ 2T\sqrt{2\epsilon}|\gS| + 2\log\left(\frac{N+1}{\delta}\right) + 2d\log\left(1+\frac{2\sqrt{d}}{\epsilon}\right).
\end{align}

Analogously, we can buond $-\sum_{i=1}^{t-1} Y_{i,n}^f$ for all $n\in[N]$ with probability at least $1-\frac{N\delta}{N+1}$ as follows:
\begin{align}
    \forall t\in[T],f\in\gF,n\in[N]:\quad -\sum_{i=1}^{t-1} Y_{i,n}^f &\leq -2\sum_{i=1}^{t-1} \E_{(s_i^n,\va_i^n)\sim d^{\widetilde{\vpi}_{i,n}}(\rho)}\left[\ell(f,s_i^n,\va_i^n)\right] \notag\\
    &\quad+ 2T\sqrt{2\epsilon}|\gS| + 2\log\left(\frac{N+1}{\delta}\right) + 2d\log\left(1+\frac{2\sqrt{d}}{\epsilon}\right).
\end{align}
By letting $\epsilon=\frac{1}{T^2|\gS|^2}$ in the above two inequalities, we have the desired result.

\subsubsection{Proof of Lemma~\ref{lm:V_diff_bound_infinite}}\label{app:proof_V_diff_bound_infinite}
Similar as in the proof of Lemma~\ref{lm:V_diff_bound} in Appendix~\ref{app:proof_V_diff_bound}, we first reformulate the value difference sequence $\sum_{t=1}^T \left|V_{\widehat{f}_{t},n}^{\widehat{\vpi}_{t}}(\rho)-V_{n}^{\widehat{\vpi}_{t}}(\rho)\right|$.

\paragraph{Step 1: reformulation of the value difference sequence.}
For any $f\in\gF$ and $\vpi=(\pi^1,\cdots,\pi^N):\gS\rightarrow\Delta(\gA)$, we have
\begin{align}\label{eq:V_f_decompose_infinite}
    \forall n\in[N]:\quad V_{f,n}^{\vpi}(\rho) &= \E_{s_0\sim\rho,\va_h\sim\pi(\cdot|s_h),\atop s_{h+1}\sim\PP(\cdot|s_h,\va_h)}\left[\sum_{h=0}^\infty \gamma^h V_{f,n}^{\vpi}(s_h)-\gamma^{h+1} V_{f,n}^{\vpi}(s_{h+1})\right]\notag\\
    &=\E_{s_0\sim\rho,\va_h\sim\pi(\cdot|s_h),\atop s_{h+1}\sim\PP(\cdot|s_h,\va_h)}\left[\sum_{h=0}^\infty \gamma^h\left( Q_{f,n}^{\vpi}(s_h,\va_h)-\beta\log\frac{\pi^n(a_h^n|s_h^n)}{\piref^n(a_h^n|s_h^n)} -\gamma V_{f,n}^\vpi(s_{h+1})\right)\right],
\end{align}
where in the second line we use the fact that
$$V_{f,n}^\vpi (s)=\E_{\va\sim\vpi(\cdot|s)}\left[Q_{f,n}^\vpi(s,\va)-\beta\log\frac{\pi^n(a^n|s^n)}{\piref^n(a^n|s^n)}\right].$$
And by the definition of $V_{n}^{\vpi}$ we have
\begin{align}\label{eq:V_by_def_infinite}
    \forall n\in[N]:\quad V_{n}^{\vpi}(\rho) = \E_{s_0\sim\rho,\va_h\sim\pi(\cdot|s_h),\atop s_{h+1}\sim \PP(\cdot|s_h,\va_h)}\left[\sum_{h=0}^\infty \gamma^h \left(r^n(s_h,\va_h)-\beta\log\frac{\pi^n(a_h^n|s_h^n)}{\piref^n(a_h^n|s_h^n)}\right)\right].
\end{align}
To simplify the notation, we define
\begin{align}\label{eq:Ppushforward_infinite}
    \forall g\in\gF:\quad\PP_g V_{f,n}^{\vpi}(s,\va)\coloneqq \E_{s'\sim\PP_g(\cdot|s,\va)}\left[V_{f,n}^{\vpi}(s')\right].
\end{align}
Combining \eqref{eq:V_f_decompose_infinite} and \eqref{eq:V_by_def}, we have
\begin{align}\label{eq:V_to_res_infinite}
    V_{f,n}^{\vpi}(\rho) - V_{n}^{\vpi}(\rho) &= \E_{s_0\sim\rho,\va_h\sim\pi(\cdot|s_h),\atop s_{h+1}\sim\PP(\cdot|s_h,\va_h)}\left[\sum_{h=0}^\infty \gamma^h\left( Q_{f,n}^{\vpi}(s_h,\va_h) - r^n(s_h,\va_h) - \gamma V_{f,n}^\vpi(s_{h+1})\right)\right]\notag\\
&= \frac{1}{1-\gamma}\E_{(s,\va)\sim d^\vpi(\rho)}\left[Q_{f,n}^{\vpi}(s,\va) - r^n(s,\va) - \gamma \PP V_{f,n}^\vpi(s,\va)\right]\notag\\
&= \frac{\gamma}{1-\gamma}\E_{(s,\va)\sim d^\vpi(\rho)}\big[\underbrace{\PP_f V_{f,n}^\vpi(s,\va)-\PP V_{f,n}^\vpi(s,\va)}_{\coloneqq \gE_n^\vpi(f,s,\va)}\big],
\end{align}
where the last relation follows from \eqref{eq:Q_beta_infinite}, and we define
\begin{align}\label{eq:gE_infinite}
    \gE_n^\vpi(f,s,\va)\coloneqq \PP_f V_{f,n}^\vpi(s,\va)-\PP V_{f,n}^\vpi(s,\va).
\end{align}
Thus we have
\begin{align}\label{eq:V_diff_decompose_infinite}
    \sum_{t=1}^T \left|V_{\widehat{f}_t,n}^{\widehat{\vpi}_t}(\rho)-V_{n}^{\widehat{\vpi}_t}(\rho)\right| = \frac{\gamma}{1-\gamma}\sum_{t=1}^T \left|\E_{(s,\va)\sim d^{\widehat{\vpi}_t}(\rho)}\left[\gE_n^{\widehat{\vpi}_t}(\widehat{f}_t,s,\va)\right]\right|.
\end{align}

Therefore, to bound $\sum_{t=1}^T \left|V_{\widehat{f}_t,n}^{\widehat{\vpi}_t}(\rho)-V_{n}^{\widehat{\vpi}_t}(\rho)\right|$, it suffices to bound the sum of model estimation errors $\sum_{t=1}^T\E_{(s,\va)\sim d^{\widehat{\vpi}_t}(\rho)}\left[\gE_n^{\widehat{\vpi}_t}(\widehat{f}_t,s,\va)\right]$.

\paragraph{Step 2: bounding the sum of model estimation errors.} 
By Assumption~\ref{asmp:function_class}, there exist $\theta_f$ and $\theta^\star$ in $\Theta$ such that $f(s'|s,\va)=\phi(s,\va,s')^\top\theta_f$ and $\PP(s'|s,\va)=\phi(s,\va,s')^\top\theta^\star$.
Thus we have
\begin{align}
    \E_{(s,\va)\sim d^\vpi(\rho)}\left[\gE_n^\vpi(f,s,\va)\right] = (\theta_f-\theta^\star)^\top\underbrace{\E_{(s,\va)\sim d^\vpi(\rho)}\left[\int_{\gS}\phi(s,\va,s')V_{f,n}^\vpi(s')ds'\right]}_{\coloneqq x_n(f,\vpi)}.
\end{align}

We let $x_n^i(f,\vpi)$ denote the $i$-th component of $x_n(f,\vpi)$, i.e., $$x_n^i(f,\vpi)=\E_{(s,\va)\sim d^\vpi(\rho)}\left[\int_{\gS}\phi^i(s,\va,s')V_{f,n}^\vpi(s')ds'\right].$$
Then we have
\begin{align}\label{eq:x_n_i_bound_infinite}
    \forall i\in[d]:\quad |x_n^i(f,\vpi)|\leq \frac{1}{1-\gamma}
\end{align}
(recall that by the definition of linear mixture model (c.f. Assumption~\ref{asmp:function_class_infinite}), $\phi^i(s,\va,\cdot)\in\Delta(\gS)$ for each $i\in[d]$), which gives
\begin{align}\label{eq:x_n_bound_infinite}
    \norm{x_n(f,\vpi)}_2\leq \frac{1}{1-\gamma}\sqrt{d}.
\end{align}

For each $t\in[T]$, we define $\Lambda_t\in\R^{d\times d}$ as
\begin{align}\label{eq:Lambda_mg_infinite}
\Lambda_t\coloneqq \lambda I_d + \sum_{i=1}^{t-1} x_n(\widehat{f}_i,\widehat{\vpi}_i)x_n(\widehat{f}_i,\widehat{\vpi}_i)^\top.
\end{align}

We write $\widehat{\theta}_t$ as the parameter of $\widehat{f}_t$.
Then we have the following decomposition:
\begin{align}\label{eq:bound_sum_gE_infinite}
    \sum_{t=1}^T \left| \E_{(s,\va)\sim d^{\widehat{\vpi}_t}(\rho)}\left[\gE_n^{\widehat{\vpi}_t}(\widehat{f}_t,s,\va)\right]\right|&= \underbrace{\sum_{t=1}^T \left| \langle x_n(\widehat{f}_t,\widehat{\vpi}_t),\widehat{\theta}_t-\theta^\star\rangle\right|\mathbbm{1}\left\{\norm{x_n(\widehat{f}_t,\widehat{\vpi}_t)}_{\Lambda_{t}^{-1}}\leq 1\right\}}_{(a)}\notag\\
& \quad + \underbrace{\sum_{t=1}^T \left| \langle x_n(\widehat{f}_t,\widehat{\vpi}_t),\widehat{\theta}_t-\theta^\star\rangle\right|\mathbbm{1}\left\{\norm{x_n(\widehat{f}_t,\widehat{\vpi}_t)}_{\Lambda_{t}^{-1}}> 1\right\}}_{(b)}.
\end{align}
Below we bound (a) and (b) separately.

\paragraph{Bounding (a).}

By the Cauchy-Schwarz inequality, we have
\begin{align}\label{eq:bound_sum_gE_infinite_2}
    (a) &\leq \sum_{t=1}^T \norm{\widehat{\theta}_t-\theta^\star}_{\Lambda_t}\norm{x_n(\widehat{f}_t,\widehat{\vpi}_t)}_{\Lambda_{t}^{-1}}\mathbbm{1}\left\{\norm{x_n(\widehat{f}_t,\widehat{\vpi}_t)}_{\Lambda_{t}^{-1}}\leq 1\right\}\notag\\
    &\leq \sum_{t=1}^T \norm{\widehat{\theta}_t-\theta^\star}_{\Lambda_t}\min\left\{ \norm{x_n(\widehat{f}_t,\widehat{\vpi}_t)}_{\Lambda_{t}^{-1}},1\right\},
\end{align}
where the last inequality follows from the fact that
$$\norm{x_n(\widehat{f}_t,\widehat{\vpi}_t)}_{\Lambda_{t}^{-1}}\mathbbm{1}\left\{\norm{x_n(\widehat{f}_t,\widehat{\vpi}_t)}_{\Lambda_{t}^{-1}}\leq 1\right\}\leq \min\left\{ \norm{x_n(\widehat{f}_t,\widehat{\vpi}_t)}_{\Lambda_{t}^{-1}},1\right\}.$$

By Lemma~\ref{lm:potential}, Lemma~\ref{lm:information_gain} and \eqref{eq:x_n_bound_infinite}, we have
\begin{align}\label{eq:|x|_bound_mg_infinite}
\sum_{i=1}^t \min\left\{ \norm{x_n(\widehat{f}_i,\widehat{\vpi}_i)}_{\Lambda_{i}^{-1}},1\right\}\leq 2d\log\left(1+\frac{T}{(1-\gamma)^2\lambda}\right)\coloneqq d_\gamma(\lambda).
\end{align}
holds for any $\lambda>0$ and $t\in[T]$.

Further, by the definition of $\Lambda_t$ (c.f. \eqref{eq:Lambda_mg_infinite}) and Assumption~\ref{asmp:function_class_infinite} we have
\begin{align}\label{eq:|w|_bound_mg_infinite}
\norm{\widehat{\theta}_t-\theta^\star}_{\Lambda_t}\leq 2\sqrt{\lambda d}+\left(\sum_{i=1}^{t-1} |\langle \widehat{\theta}_t-\theta^\star,x_n(\widehat{f}_i,\widehat{\vpi}_i)\rangle|^2\right)^{1/2},
\end{align}
which gives
\begin{align}\label{eq:intermediate1_infinite}
&\quad\sum_{t=1}^T \norm{\widehat{\theta}_t-\theta^\star}_{\Lambda_t}\min\left\{ \norm{x_n(\widehat{f}_t,\widehat{\vpi}_t)}_{\Lambda_{t}^{-1}},1\right\}\notag\\
&\leq \sum_{t=1}^T\left(2\sqrt{\lambda d} +\left(\sum_{i=1}^{t-1} |\langle \widehat{\theta}_t-\theta^\star,x_n(\widehat{f}_i,\widehat{\vpi}_i)\rangle|^2\right)^{1/2}\right)\cdot\min\left\{ \norm{x_n(\widehat{f}_t,\widehat{\vpi}_t)}_{\Lambda_{t}^{-1}},1\right\}\notag\\
&\leq \left(\sum_{t=1}^T 4\lambda d\right)^{1/2}
\left(\sum_{t=1}^T \min\left\{ \norm{x_n(\widehat{f}_t,\widehat{\vpi}_t)}_{\Lambda_{t}^{-1}},1\right\}\right)^{1/2}\notag\\
&\quad + \left(\sum_{t=1}^T \sum_{i=1}^{t-1} |\langle \widehat{\theta}_t-\theta^\star,x_n(\widehat{f}_i,\widehat{\vpi}_i)\rangle|^2\right)^{1/2}\left(\sum_{t=1}^T \min\left\{ \norm{x_n(\widehat{f}_t,\widehat{\vpi}_t)}_{\Lambda_{t}^{-1}},1\right\}\right)^{1/2}\notag\\
&\leq 2\sqrt{\lambda dT \min\{d_\gamma(\lambda),T\}} + \left(d_\gamma(\lambda)\sum_{t=1}^T \sum_{i=1}^{t-1} |\langle \widehat{\theta}_t-\theta^\star,x_n(\widehat{f}_i,\widehat{\vpi}_i)\rangle|^2\right)^{1/2},
\end{align}
where the first inequality uses \eqref{eq:|w|_bound_mg_infinite} and the second inequality uses the Cauchy-Schwarz inequality and the fact that 
$$\min\left\{ \norm{x_n(\widehat{f}_t,\widehat{\vpi}_t)}_{\Lambda_{t}^{-1}},1\right\}^2\leq \min\left\{ \norm{x_n(\widehat{f}_t,\widehat{\vpi}_t)}_{\Lambda_{t}^{-1}},1\right\},$$
and the last inequality uses \eqref{eq:|x|_bound_mg_infinite}.

Furthermore, we have
\begin{align}\label{eq:bound_residual_infinite}
|\langle \widehat{\theta}_t-\theta^\star,x_n(\widehat{f}_i,\widehat{\vpi}_i)\rangle|^2 &= \left|\E_{(s,\va)\sim d^{\widehat{\vpi}_i}(\rho)}\left[\int_\gS \left(\PP_{\widehat{f}_t}(s'|s,\va)-\PP(s'|s,\va) \right)V_{\widehat{f}_i,n}^{\widehat{\vpi}_i}(s')ds'\right]\right|^2\notag\\
&\leq \E_{(s,\va)\sim d^{\widehat{\vpi}_i}(\rho)} \left[\left(\int_\gS \left(\PP_{\widehat{f}_t}(s'|s,\va)-\PP(s'|s,\va) \right)V_{\widehat{f}_i,n}^{\widehat{\vpi}_i}(s')ds'\right)^2\right]\notag\\
&\leq 4\norm{V_{\widehat{f}_i,n}^{\widehat{\vpi}_i}(\cdot)}_\infty\E_{(s,\va)\sim d^{\widehat{\vpi}_t}(\rho)}D_{\textnormal{TV}}^2\left(\PP_{\widehat{f}_t}(\cdot|s,\va)\big\|\PP(\cdot|s,\va)\right)\notag\\
&\leq \frac{8}{1-\gamma}\E_{(s,\va)\sim d^{\widehat{\vpi}_i}(\rho)}\hellinger{\PP_{\widehat{f}_t}(\cdot|s,\va)}{\PP(\cdot|s,\va)}\notag\\
&=\frac{8}{1-\gamma} \E_{(s,\va)\sim d^{\widehat{\vpi}_i}(\rho)}\ell(\widehat{f}_t,s,\va),
\end{align}
where the second line uses the Cauchy-Schwarz inequality, the third line follows from H\"oder's inequality,
the fourth line uses the inequality $D_{\textnormal{TV}}^2(P\|Q)\leq2\hellinger{P}{Q}$
and the fact that $\norm{V_{\widehat{f}_t,n}^{\widehat{\vpi}_t}(\cdot)}_\infty\leq \frac{1}{1-\gamma}$. The last line uses \eqref{eq:Hellinger_loss_infinite}.

Plugging \eqref{eq:bound_residual_infinite} into \eqref{eq:intermediate1_infinite}, we have
\begin{align}\label{eq:bound_a_infinite}
    (a)\leq 2\sqrt{d}\cdot\sqrt{\lambda T \min\{d_\gamma(\lambda),T\}} + \left(\frac{8d_\gamma(\lambda)}{1-\gamma}\sum_{t=1}^T \sum_{i=1}^{t-1} \E_{(s,\va)\sim d^{\widehat{\vpi}_i}(\rho)}\ell(\widehat{f}_t,s,\va)\right)^{1/2}.
\end{align}

\paragraph{Bounding (b).} 
\begin{align}\label{eq:bound_b_infinite_pre}
    (b) &= \sum_{t=1}^T \left| \langle x_n(\widehat{f}_t,\widehat{\vpi}_t),\widehat{\theta}_t-\theta^\star\rangle\right|\mathbbm{1}\left\{\norm{x_n(\widehat{f}_t,\widehat{\vpi}_t)}_{\Lambda_{t}^{-1}}> 1\right\}\notag\\
    &\leq \sum_{t=1}^T \left| \langle x_n(\widehat{f}_t,\widehat{\vpi}_t),\widehat{\theta}_t-\theta^\star\rangle\right|\min\left\{ \norm{x_n(\widehat{f}_t,\widehat{\vpi}_t)}_{\Lambda_{t}^{-1}},1\right\},
\end{align}
where the inequality follows from the fact that
$$
\mathbbm{1}\left\{\norm{x_n(\widehat{f}_t,\widehat{\vpi}_t)}_{\Lambda_{t}^{-1}}> 1\right\}\leq \min\left\{ \norm{x_n(\widehat{f}_t,\widehat{\vpi}_t)}_{\Lambda_{t}^{-1}},1\right\}.
$$

Note that 
$$\norm{x_n(\widehat{f}_t,\widehat{\vpi}_t)}_{\Lambda_{t}^{-1}}\mathbbm{1}\left\{\norm{x_n(\widehat{f}_t,\widehat{\vpi}_t)}_{\Lambda_{t}^{-1}}\leq 1\right\}\leq \min\left\{ \norm{x(\widehat{\mu}_t,\widehat{\nu}_t)}_{\Lambda_{t}^{-1}},1\right\}$$

Thus by \eqref{eq:|x|_bound_mg_infinite} and \eqref{eq:bound_b_infinite_pre}, we have
\begin{align}\label{eq:bound_b_infinite}
    (b)\leq \frac{1}{1-\gamma} \min\{T, d_\gamma(\lambda)\}.
\end{align}

Plugging \eqref{eq:bound_a_infinite}, \eqref{eq:bound_b_infinite} into \eqref{eq:bound_sum_gE_infinite}, we have
\begin{align}
&\quad\sum_{t=1}^T \left| \E_{(s,\va)\sim d^{\widehat{\vpi}_t}(\rho)}\left[\gE_n^{\widehat{\vpi}_t}(\widehat{f}_t,s,\va)\right]\right|\notag\\
&\leq 2\sqrt{d}\cdot\sqrt{\lambda T \min\{d_\gamma(\lambda),T\}} + \left(\frac{8d_\gamma(\lambda)}{1-\gamma}\sum_{t=1}^T \sum_{i=1}^{t-1} \E_{(s,\va)\sim d^{\widehat{\vpi}_i}(\rho)}\ell(\widehat{f}_t,s,\va)\right)^{1/2}+\frac{1}{1-\gamma} \min\{T, d_\gamma(\lambda)\}\notag\\
&\leq \left(\frac{8d_\gamma(\lambda)}{(1-\gamma)\eta}\cdot\eta\sum_{t=1}^T \sum_{i=1}^{t-1}  \E_{(s,\va)\sim d^{\widehat{\vpi}_i}(\rho)}\ell(\widehat{f}_t,s,\va)\right)^{1/2} + \left(\sqrt{d}+\frac{1}{1-\gamma}\right) \min\{d_\gamma(\lambda),T\} +\sqrt{d} \lambda T\notag\\
&\leq \frac{4d_\gamma(\lambda)}{(1-\gamma)\eta} +\frac{\eta}{2}\sum_{t=1}^T \sum_{i=1}^{t-1} \E_{(s,\va)\sim d^{\widehat{\vpi}_i}(\rho)}\ell(\widehat{f}_t,s,\va) + \left(\sqrt{d}+\frac{1}{1-\gamma}\right) \min\{d_\gamma(\lambda),T\} +\sqrt{d} \lambda T
\end{align}
for any $\eta>0$, where the second and third inequalities both use the fact that $\sqrt{ab}\leq \frac{a+b}{2}$ for any $a,b\geq 0$. 

Finally, combining \eqref{eq:V_diff_decompose_infinite} with the above inequality, we have \eqref{eq:V_diff_bound_infinite}.

\end{document}